\documentclass[runningheads]{llncs}
\usepackage{times}
\usepackage{graphicx}
\usepackage{amsmath,amssymb} 
\usepackage{color}
\usepackage{epstopdf}
\usepackage{array}
\usepackage{multirow}
\usepackage{microtype}
\usepackage{algpseudocode}
\usepackage{algorithm}
\usepackage{bm}
\usepackage{framed}
\usepackage{subfig}
\usepackage[font=small,skip=5pt]{caption}
\usepackage[width=122mm,left=12mm,paperwidth=146mm,height=193mm,top=12mm,paperheight=217mm]{geometry}

\newcommand{\realset}{\mathbb{R}}

\newcommand{\vecq}{\mathbf{q}}
\newcommand{\vecx}{\mathbf{x}}

\newcommand{\vecv}{\mathbf{v}}

\newcommand{\vecb}{\mathbf{b}}

\newcommand{\bA}{\mathbf{A}}
\newcommand{\bB}{\mathbf{B}}
\newcommand{\bM}{\mathbf{M}}

\newcommand{\bX}{\mathbf{X}}
\newcommand{\bQ}{\mathbf{Q}}
\newcommand{\bC}{\mathbf{C}}
\newcommand{\bL}{\mathbf{L}}
\newcommand{\bY}{\mathbf{Y}}
\newcommand{\bI}{\mathbf{I}}

\newcommand{\bZ}{\mathbf{Z}}

\newcommand{\setE}{\mathcal{E}}
\newcommand{\setB}{\mathcal{B}}
\newcommand{\setA}{\mathcal{A}}

\DeclareMathOperator*{\argmin}{arg\,min}
\DeclareMathOperator*{\trace}{tr}
\newcommand{\repeatthanks}{\textsuperscript{\thefootnote}}

\begin{document}
\pagestyle{headings}
\mainmatter

\title{Biconvex Relaxation for Semidefinite Programming in Computer Vision} 

\titlerunning{Biconvex Relaxation for Semidefinite Programming in Computer Vision}

\authorrunning{S. Shah, A. Yadav, C. Castillo, D. Jacobs, C. Studer and T. Goldstein}

\author{Sohil Shah\thanks{The first two authors contributed equally to this work.}$^{,1}$, Abhay Kumar Yadav\repeatthanks$^{,1}$, Carlos D. Castillo$^1$, David W. Jacobs$^1$, Christoph Studer$^2$, \and Tom Goldstein$^1$}
\institute{$^1$ University of Maryland, College Park, MD USA; $^2$ Cornell University, Ithaca, NY USA. \\
\email{sohilas@umd.edu,\{jaiabhay,djacobs,tomg\}@cs.umd.edu, carlos@umiacs.umd.edu, studer@cornell.edu}}

\maketitle

\begin{abstract}
Semidefinite programming (SDP) is an indispensable tool in computer vision, but general-purpose solvers for SDPs are often too slow and memory intensive for large-scale problems. Our framework, referred to as biconvex relaxation (BCR), transforms an SDP consisting of PSD constraint matrices into a specific biconvex optimization problem, which can then be approximately solved in the original, low-dimensional variable space at low complexity. The resulting problem is solved using an efficient alternating minimization (AM) procedure. Since AM has the potential to get stuck in local minima, we propose a general initialization scheme that enables BCR to start close to a global optimum---this is key for BCR to quickly converge to optimal or near-optimal solutions. We showcase the efficacy of our approach on three applications in computer vision, namely segmentation, co-segmentation, and manifold metric learning. BCR achieves solution quality comparable to state-of-the-art SDP methods with speedups between $4\times$ and $35\times$.


\end{abstract}
\section{Introduction}
Optimization problems involving either integer-valued vectors or low-rank matrices are ubiquitous in computer vision.
Graph-cut methods for image segmentation, for example, involve optimization 
problems where integer-valued variables represent region labels~\mbox{\cite{shi2000normalized,keuchel2003binary,torr2003solving,goemans1995improved}}. Problems in multi-camera structure from motion \cite{arie2012global}, manifold embedding~\cite{weinberger2006unsupervised}, and matrix
completion \cite{mitra2010large} all rely on optimization problems involving matrices with low rank constraints. Since these constraints are non-convex, the design of efficient algorithms that find globally optimal solutions is a difficult task.

For a wide range of applications \cite{luo2010semidefinite,lasserre2001explicit,boyd1997semidefinite,ecker2008semidefinite,shirdhonkar2005non,weinberger2006unsupervised}, non-convex constraints
can be handled by {\em semidefinite relaxation} (SDR)~\cite{luo2010semidefinite}.  In this approach, a non-convex optimization problem involving a vector of unknowns is ``lifted'' to a higher dimensional
convex problem that involves a positive semidefinite (PSD) matrix, which then enables one to solve a SDP~\cite{vandenberghe1996semidefinite}. 
While SDR delivers state-of-the-art performance in a wide range of applications \cite{goemans1995improved,luo2010semidefinite,heiler2005semidefinite,torr2003solving,weinberger2006unsupervised,mitra2010large}, the approach significantly increases the dimensionality of the original optimization problem (i.e., replacing a vector with a matrix), which typically results in exorbitant computational costs and memory requirements. Nevertheless, SDR leads to SDPs whose global optimal solution can be found using robust numerical methods. 

A growing number of computer-vision applications involve high-resolution images (or videos) that require SDPs with a large number of variables. General-purpose (interior point) solvers for SDPs  do not scale well to such problem sizes; the worst-case complexity is $O(N^{6.5}\log(1/\varepsilon))$ for an  $N\times N$ problem with $\varepsilon$ objective error~\cite{shen2011scalable}. In imaging applications, $N$ is often proportional to the number of pixels, which is potentially large.
 
  %
 %
 %
The prohibitive complexity and memory requirements of solving SDPs exactly with a large number of variables has spawned interest
in fast, non-convex solvers that avoid lifting.
For example, recent progress in phase retrieval by Netrapalli \emph{et al.}~\cite{netrapalli2013phase} and  Cand\`es \emph{et al.}~\cite{candes2015phase} has shown that non-convex
optimization methods provably achieve solution quality comparable to exact SDR-based methods with significantly lower complexity. These methods operate on the original dimensions of the (un-lifted) problem, which enables their use on high-dimensional problems.
Another prominent example is max-norm regularization by Lee \emph{et al.}~\cite{lee2010practical}, which was proposed for solving high-dimensional matrix-completion problems and to approximately perform max-cut clustering. This method was shown to outperform exact SDR-based methods in terms of computational complexity, while delivering acceptable solution quality. 
While both of these examples outperform classical SDP-based methods, they are limited to very specific problem types,
and cannot handle more complex SDPs that typically appear in computer vision. 

\subsection{Contributions}

We introduce a novel framework for approximately solving SDPs with positive semi-definite constraint matrices in a computationally efficient manner and with small memory footprint.
Our proposed {\em bi-convex} relaxation (BCR), transforms an SDP into a biconvex optimization problem, which can then be solved in the original, low-dimensional variable space at low complexity. 
%
%
The resulting biconvex problem is solved using a computationally-efficient AM procedure.
Since AM is prone to get stuck in local minima, we propose an initialization scheme that enables BCR to start close to the global optimum of the original SDP---this initialization is key for our algorithm to quickly converge to an optimal or near-optimal solution. 
We showcase the effectiveness of the BCR framework by comparing to highly-specialized SDP solvers for a selected set
of problems in computer vision involving image segmentation, co-segmentation, and metric learning on manifolds.
Our results demonstrate that BCR enables high-quality results 
while achieving speedups ranging from $4\times$ to
$35\times$ over state-of-the-art competitor methods \cite{Wang_2013,biasedncut,Journee,huang2015log,harandi2014manifold} for the studied applications.


\section{Background and Relevant Prior Art}

We now briefly review semidefinite programs (SDPs) and discuss prior work on fast, approximate solvers for SDPs in computer vision and related applications.

\subsection{Semidefinite Programs (SDPs)}

SDPs find use in a large and growing number of fields, including computer vision, machine learning, signal and image processing, statistics, communications, and control~\cite{vandenberghe1996semidefinite}. SDPs can be written in the following general form:
\begin{equation} \label{eq:master}
{\small
  \begin{aligned}
     \underset{\bY \in \mathcal S^+_{N\times N}}{\text{minimize}}
     & \quad \langle \bC, \bY  \rangle \\
     \text{subject to}   & \quad \langle \bA_i, \bY \rangle = b_i, \quad \forall i \in \setE,\\
         & \quad \langle \bA_j, \bY \rangle \leq b_j, \quad\! \forall j \in \setB,
  \end{aligned}}
\end{equation}
where $S^+_{N\times N}$ represents the set of  $N\times N$ symmetric positive semidefinite matrices, and $\langle \bC, \bY \rangle=\trace(\bC^T \bY)$ is the matrix inner product.
The sets $\setE$ and $\setB$ contain the indices associated with the equality and inequality constraints, respectively; ~$\bA_i$ and~$\bA_j$ are symmetric matrices of appropriate dimensions. 

The key advantages of SDPs are that (i) they enable the transformation of certain non-convex constraints into convex constraints via semidefinite relaxation (SDR)~\cite{luo2010semidefinite} and (ii) the resulting problems often come with strong theoretical guarantees. 
%

In computer vision, a large number of problems can be cast as SDPs of the general form \eqref{eq:master}.
For example, \cite{weinberger2006unsupervised}
formulates image manifold learning as an SDP, \cite{shirdhonkar2005non} uses an SDP to enforce a non-negative lighting constraint when recovering scene lighting and
object albedos, \cite{bai2004graph} uses an SDP for graph matching,
\cite{arie2012global} proposes an SDP that recovers the orientation of multiple cameras
from point correspondences and essential matrices, and \cite{mitra2010large} uses
low-rank SDPs to solve matrix-completion problems that arise in
structure-from-motion and photometric stereo.

\subsection{SDR for Binary-Valued Quadratic Problems}
Semidefinite relaxation is commonly used to solve binary-valued labeling problems.
For such problems, a set of variables take on binary values while minimizing a quadratic cost function that
depends on the assignment of pairs of variables.  
Such labeling problems typically arise from Markov random fields (MRFs) for which many solution methods exist \cite{wang2013markov}.  Spectral methods, e.g.,
\cite{shi2000normalized}, are often used to solve such binary-valued quadratic
problems (BQPs)---the references \cite{keuchel2003binary,torr2003solving} used SDR inspired by the work of
\cite{goemans1995improved} that provides a generalized SDR for the max-cut problem. 
BQP problems have wide applicability to computer vision problems, such 
as segmentation and perceptual organization~\cite{keuchel2003binary,Wang_2013,Joulin_2010}, semantic segmentation~\cite{wang2015efficient}, matching~\cite{torr2003solving,schellewald2005probabilistic}, surface reconstruction
including photometric stereo and shape from defocus~\cite{ecker2008semidefinite},
and image restoration~\cite{olsson2007solving}.

BQPs can be solved by lifting the binary-valued label vector $\vecb\in\{\pm1\}^N$ to an $N^2$-dimensional matrix space by forming the PSD matrix $\bB = \vecb\vecb^T$, whose non-convex rank-1 constraint is relaxed to PSD matrices $\bB\in S^+_{N\times N}$ with an all-ones diagonal~\cite{luo2010semidefinite}. 
The goal is then to solve a SDP for $\bB$ in the hope that the resulting matrix has rank 1;
if $\bB$ has higher rank, an approximate solution must be extracted which can either be obtained from the leading eigenvector or via randomization methods~\cite{luo2010semidefinite,lang2005fixing}. 

\subsection{Specialized Solvers for SDPs}
General-purpose solvers for SDPs, such as SeDuMi~\cite{Sturm98usingsedumi} or SDPT3~\cite{Toh98sdpt3}, rely on interior point methods with high computational complexity and memory requirements. Hence, their use is restricted to low-dimensional problems. For problems in computer vision, where the number of variables can become comparable to the number of pixels in an image, more efficient algorithms are necessary. 
A handful of special-purpose algorithms have been proposed to solve specific problem types arising in computer vision.   These algorithms fit into two classes: (i) convex algorithms that solve the original SDP by exploiting problem structure and (ii) non-convex methods that avoid lifting.

For certain problems, one can exactly solve SDPs with much lower complexity than interior point schemes, especially for  BQP problems in computer vision. Ecker \emph{et al.}~\cite{ecker2008semidefinite} deployed a number of heuristics to speed up
the Goemans-Williamson SDR~\cite{goemans1995improved} for surface reconstruction. 
Olsson \emph{et al.}~\cite{olsson2007solving} proposed a spectral subgradient
method to solve BQP problems that include a linear term, but are unable to handle inequality constraints. 
A particularly popular approach is the SDCut algorithms of Wang \emph{et al.}~\cite{Wang_2013}. This method solves BQP for some types of segmentation problems using dual gradient descent.  SDCut
leads to a similar relaxation as for BQP problems, but enables significantly
lower complexity for graph cutting and its variants.  To the best of our knowledge, the method by Wang \emph{et al.}~\cite{Wang_2013} yields state-of-the-art performance---nevertheless, our proposed method is at least an order of magnitude faster, as shown in Section  \ref{applications}.

Another algorithm class contains non-convex approximation methods that avoid lifting altogether. Since these methods work with low-dimensional unknowns, they  are potentially more efficient than lifted methods.  Simple examples include the Wiberg method \cite{okatani2007wiberg}  for low-rank matrix approximation, which uses Newton-type iterations to minimize a non-convex objective.  A number of methods have been proposed for SDPs where the objective function is simply the trace-norm of $\bY$ (i.e., problem \eqref{eq:master} with \mbox{$\bC=\bI$}) and without inequality constraints.  Approaches include replacing the trace norm with the max-norm~\cite{lee2010practical}, or using the so-called Wirtinger flow to solve phase-retrieval problems~\cite{candes2015phase}.  One of the earliest approaches for non-convex methods are due to Burer and Montiero~\cite{burer2003nonlinear}, who propose an augmented Lagrangian method.  While this method is able to handle arbitrary objective functions, it does not naturally support inequality constraints (without introducing auxiliary slack variables). Furthermore, this approach uses convex methods for which convergence is not well understood and is sensitive to the initialization value. 

While most of the above-mentioned methods provide best-in-class performance at
low computational complexity, they are limited to very specific problems and cannot be
generalized to other, more general SDPs.

%
%
%
%
%
%
%

\section{Biconvex Relaxation (BCR) Framework}
We now present the proposed \emph{biconvex relaxation (BCR)} framework. We then propose an alternating minimization procedure and a suitable initialization method. 

\subsection{Biconvex Relaxation}
%
Rather than solving the general SDP \eqref{eq:master} directly, we
exploit the following key fact: any matrix $\bY$ is symmetric positive
semidefinite if and only if it has an expansion of the form $\bY =\bX\bX^T.$
By substituting the factorization $\bY =\bX\bX^T$  into \eqref{eq:master}, we
are able to remove the semidefinite constraint and arrive at the following
problem:  
%
%
\begin{equation}
  \label{no_q}
  {
  \begin{aligned}
     \underset{\bX \in \realset^{N \times r}}{\text{minimize}}
     & \quad \trace ( \bX^T \bC \bX ) \\
     \text{subject to} & \quad \trace(\bX^T \bA_i  \bX ) = b_i, \quad \forall i \in \setE,\\
     &\quad \trace(\bX^T \bA_j  \bX ) \leq b_j, \quad\! \forall j \in \setB,
  \end{aligned}}
\end{equation}
where $r=\text{rank}(\bY)$.\footnote{Straightforward extensions of our approach allow us to handle constraints of the form $\trace(\bX^T \bA_k  \bX ) \geq b_k, \forall k \in \setA$, as well as complex-valued matrices and vectors.}
Note that any symmetric semi-definite matrix $\bA$ has a (possibly
complex-valued) square root $\bL$ of the form $\bA=\bL^T\bL.$  Furthermore, we
have $\trace(\bX^T \bA  \bX ) =\trace(\bX^T \bL^T\bL  \bX )= \|\bL  \bX\|^2_F$,
where $\|\cdot\|_F$ is the Frobenius (matrix) norm. This formulation enables us
to rewrite \eqref{no_q} as follows:
\begin{equation}
  \label{nonconvex}
  {
  \begin{aligned}
     \underset{\bX \in \realset^{N \times r}}{\text{minimize}}
     & \quad \trace ( \bX^T \bC \bX ) \\
     \text{subject to} &  \quad  \bQ_i =  \bL_i  \bX, \quad \|\bQ_i\|^2_F = b_i, \quad \forall i \in \setE,\\
     &\quad  \bQ_j =  \bL_j  \bX, \quad  \! \|\bQ_j\|^2_F \leq b_j, \quad\! \forall j \in \setB.
  \end{aligned}}
\end{equation}
If the matrices $\{\bA_i\}$, $\{\bA_j\}$, and $\bC$ are themselves PSDs, then the objective function in \eqref{nonconvex} is convex and quadratic, and the inequality constraints in \eqref{nonconvex} are convex---non-convexity of the problem is only caused by the equality constraints. The core idea of BCR explained next is to relax these equality constraints. Here, we assume that the factors of these matrices are easily obtained from the underlying problem structure. For some applications, where these factors are not readily available this could be a computational burden (worst case $\mathcal{O}(N^3)$) rather than an asset.  
 
 \fussy
 
In the formulation \eqref{nonconvex}, we have lost convexity.  Nevertheless, whenever $r<N,$ we achieved a (potentially large) dimensionality reduction compared to the original SDP~\eqref{eq:master}.  We now relax \eqref{nonconvex} in a form that is biconvex, i.e., convex with respect to a group of variables when the remaining variables are held constant.  
By relaxing the convex problem in biconvex form, we retain many advantages of the convex formulation while maintaining low dimensionality and speed.  In particular, we propose to approximate \eqref{nonconvex} with the following \emph{biconvex relaxation (BCR)}:
\begin{equation}
  \label{biconvex}
  {
  \begin{aligned}
     \underset{\bX , \bQ_i, i \in \{ \mathcal B \cup \mathcal E\}}{\text{minimize}} \!\!\!\!\!
     & \quad \trace ( \bX^T \bC \bX ) + \frac{\alpha}{2}\!\!\sum_{i \in \{\setE \cup \setB\} }\!\! \!\| \bQ_i - \bL_i \bX\|_F^2
 - \frac{\beta}{2} \sum_{j \in \mathcal E} \|\mathbf{Q}_j\|_F^2  \\[-0.1cm]
     \text{subject to} & \quad \|\bQ_i\|^2_F \leq b_i, \quad \forall i \in \{ \mathcal B \cup \mathcal E\},
  \end{aligned}}
\end{equation}
where $\alpha>\beta>0$ are relaxation parameters (discussed in detail below).
In this BCR formulation, we relaxed the equality constraints $\|\bQ_i\|^2_F = b_i$, $\forall i \in \setE,$ to inequality constraints $\|\bQ_i\|^2_F \leq b_i$, $\forall i \in \setE$, and added negative quadratic penalty functions $-\frac{\beta}{2}\|\bQ_i\|$, $\forall i \in \setE,$ to the objective function. These quadratic penalties attempt to force the inequality constraints in $\setE$ to be satisfied exactly.
We also replaced the constraints $\bQ_i =  \bL_i \bX$ and $\bQ_j =  \bL_j \bX$ by quadratic penalty functions in the objective function. 

The relaxation parameters are chosen by freezing the ratio $\alpha/\beta$ to 2, and following a simple, principled way of setting $\beta$. Unless stated otherwise, we set $\beta$ to match the curvature of the penalty term with the curvature of the objective i.e., $\beta = \|\mathbf{C}\|_2$, so that the resulting bi-convex problem is well-conditioned.


Our BCR formulation \eqref{biconvex} has some important properties.  First, if $\bC\in  \mathcal S^+_{N\times N}$ then the problem is biconvex, i.e., convex with respect to $\bX$ when the $\{\bQ_i\}$ are held constant, and vice versa. Furthermore, consider the case of solving a constraint feasibility problem (i.e., problem~\eqref{eq:master} with $\bC=\boldsymbol 0$).  When $\bY = \bX\bX^T$ is a solution to \eqref{eq:master} with $\bC=\boldsymbol 0$, the problem \eqref{biconvex} assumes objective value $-\frac{\beta}{2}\sum_j b_j,$ which is the global minimizer of the BCR formulation \eqref{biconvex}.  Likewise, it is easy to see that any global minimizer of \eqref{biconvex} with objective value $-\frac{\beta}{2}\sum_j b_j$ must be a solution to the original problem \eqref{eq:master}.   

\subsection{Alternating Minimization (AM) Algorithm}
One of the key benefits of biconvexity is that~\eqref{biconvex} can be globally minimized with respect to $\bQ$ or~$\bX.$  Hence, it is natural to compute  approximate solutions to \eqref{biconvex} via alternating minimization. Note the convergence of AM for biconvex problems is well understood~\cite{duchisinger,douglasgunn}. 
%
The two stages of the proposed method for BCR are detailed next.


\textbf{Stage 1: Minimize with respect to $\{\bQ_i\}$.} The BCR objective in~\eqref{biconvex} is quadratic in~$\{\bQ_i\}$ with no dependence between matrices. Consequently, the optimal value of $\bQ_i$ can be found by minimizing the quadratic objective, and then reprojecting back into a unit Frobenius-norm ball of radius $\sqrt {b_i}.$  The minimizer of the quadratic objective is given by $\frac{\alpha}{\alpha-\beta_i}\bL_i\bX,$ where $\beta_i=0$ if $i\in \mathcal B$ and  $\beta_i=\beta$ if $i\in \mathcal E.$  The projection onto the unit ball then leads to the following \emph{expansion--reprojection} update:
\begin{equation}
  {
  \begin{aligned}
  \bQ_i \gets  \frac{ \bL_i\bX}{ \|\bL_i\bX\|_F}\min \Big\{\sqrt{b_i},\,\frac{\alpha}{\alpha-\beta_i}\|\bL_i\bX\|_F \Big \}.
\end{aligned}
}
\end{equation}
Intuitively, this expansion--reprojection update causes the matrix $\bQ_i$ to expand if $i\in \mathcal E$,
thus encouraging it to satisfy the relaxed constraints in \eqref{biconvex} with equality.

\textbf{Stage 2: Minimize with respect to $\bX$.} This stage solves the least-squares problem:
\begin{equation}
  {
  \begin{aligned}
   \bX \gets \argmin_{\bX \in\realset^{N \times r}} \, \trace ( \bX^T \bC \bX ) +  \frac{\alpha}{2} \!\!\!\sum_{i \in \{\mathcal E \cup \mathcal B\}}\!\!\!  \|\bQ_i\!-\!\bL_i\bX \|_F^2.
\end{aligned}
}
\end{equation}
The optimality conditions for this problem are linear equations, and the solution is
\begin{equation}
  {
  \begin{aligned}
\bX \gets  \left( \! \bC+ \alpha \sum_{i\in \{\mathcal E \cup \mathcal B\}} \bL_i^T \bL_i \right)^{\!\!-1} \!\left( \sum_{i \in \{ \mathcal E \cup \mathcal B\}} \bL_i^T \bQ_i  \right)\!,
\end{aligned}
}
\end{equation}
where the matrix inverse (one-time computation) may be replaced by a pseudo-inverse if necessary. Alternatively, one may perform a simple gradient-descent step with a suitable step size, which avoids the inversion of a  potentially large-dimensional matrix. 

The resulting AM algorithm for the proposed BCR~\eqref{biconvex} is summarized in Algorithm~\ref{algorithm}. 

\setlength{\textfloatsep}{10pt}
\begin{algorithm}[t]
\caption{AM for Biconvex Relaxation} 
\begin{algorithmic}[1]
\State\textbf{inputs}: $\bC,$ $\{\bL_i\}$, $b_i,$  $\alpha$, and $\beta$, \textbf{output}: $\bX$
\State Compute an initializer for $\bX$ as in Section~\ref{sec:leadingvector}
\State Precompute  $\bM=\big(  \bC+ \alpha \sum_{i\in \mathcal \{E \cup \mathcal B\}} \bL_i^T \bL_i \big)^{-1}$
\While{\text{not converged}}
\State $\bQ_i \gets  \frac{ \bL_i\bX}{ \|\bL_i\bX\|_F}\min \Big\{\sqrt{b_i},\,\frac{\alpha}{\alpha-\beta_i}\|\bL_i\bX\|_F \Big \}$
\State $\bX \gets  \bM \left( \sum_{i \in \{ \mathcal E \cup \mathcal B\} } \bL_i^T \bQ_i  \right),$
\EndWhile
\end{algorithmic}

\label{algorithm}
\end{algorithm}

\subsection{Initialization}
\label{sec:leadingvector}
The problem \eqref{biconvex} is biconvex and hence, a global minimizer can be found with respect to either $\{\bQ_i\}$ or $\bX,$ although a global minimizer of the joint problem is not guaranteed. 
We hope to find a global minimizer at low complexity using the AM method, but
in practice AM may get trapped in local minima, especially if the variables have been initialized poorly. 
We now propose a principled method for computing an initializer for~$\bX$ that is often close to the global optimum of the BCR problem---our initializer is key for the success of the proposed AM procedure and enables fast convergence. 
   
The papers \cite{netrapalli2013phase,candes2015phase} have considered optimization problems that arise in phase retrieval where $\mathcal B = \varnothing$ (i.e., there are only equality constraints),  $\bC=\bI$ being the identity, and $\bY$ being rank one. For such problems, the objective of \eqref{eq:master} reduces to $\trace(\bY).$  By setting $\bY=\vecx\vecx^T$, we obtain the following formulation:
   \begin{equation}
  \label{restricted}
  {
  \begin{aligned}
     \underset{\vecx \in \realset^{N}}{\text{minimize}}
     \,\, \| \vecx \|_2^2 
     \quad \text{subject to}\enspace\vecq_i =  \bL_i  \vecx, \quad \|\vecq_i\|^2_2 = b_i, \quad \forall i \in \setE.
  \end{aligned}
  }
\end{equation}
Netrapali \emph{et al.}~\cite{netrapalli2013phase} proposed an iterative algorithm for solving \eqref{restricted}, which has been initialized by the following strategy.  Define 
   \begin{equation}
  \label{eq:initializersarecool}
  {
  \begin{aligned} 
 \bZ = \frac{1}{|\setE|} \sum_{i\in \setE}  b_i\bL_i^T\bL_i.
  \end{aligned}
  }
\end{equation}
  Let $\vecv$ be the leading eigenvector of $\bZ$ and $\lambda$ the leading eigenvalue.  Then $\vecx=\lambda \vecv$ is an accurate approximation to the true solution of \eqref{restricted}.  In fact, if the matrices $\bL_i$ are sampled from a random normal distribution, then it was shown in \cite{netrapalli2013phase,candes2015phase} that $\mathbb{E}\| \vecx^\star -  \lambda \vecx \|^2_2\to 0$ (in expectation) as $|\setE| \to \infty,$ where $\vecx^\star$ is the true solution to \eqref{restricted}.
%

      

We are interested in a good initializer for the general problem in \eqref{nonconvex} where $\bX$ can be rank one or higher. We focus on problems with equality constraints only---note that one can use slack variables to convert a problem with inequality constraints into the same form~\cite{vandenberghe1996semidefinite}. 
Given that $\mathbf{C}$ is a symmetric positive definite matrix, it can be decomposed into $\mathbf{C} = \mathbf{U}^T\mathbf{U}$. By the change of variables $\widetilde{\mathbf{X}} = \mathbf{U}\mathbf{X}$, we can rewrite \eqref{eq:master} as follows: 
\begin{equation}
\label{eq:newform}
{
  \begin{aligned}
\underset{\bX \in \realset^{N \times r}}{\text{minimize}}  \,\,\| \widetilde{\mathbf{X}}\|^2_F \quad 
\text{subject to}\enspace\langle \widetilde{\mathbf{A}}_i, \widetilde{\mathbf{X}}\widetilde{\mathbf{X}}^T \rangle = b_i, \quad \forall i \in \setE,
\end{aligned}
}
\end{equation}
where $\widetilde{\mathbf{A}}_i = \mathbf{U}^{-T}\mathbf{A}_i\mathbf{U}^{-1}$, and we omitted the inequality constraints. 
%
%
To initialize the proposed AM procedure in Algorithm~\ref{algorithm}, we make the change of variables $\widetilde{\mathbf{X}} = \mathbf{U}\mathbf{X}$ to transform the BCR formulation into the form of \eqref{eq:newform}.  
Analogously to the initialization procedure in \cite{netrapalli2013phase} for phase retrieval, we then compute an initializer $\widetilde{\mathbf{X}}_0$ using the leading~$r$ eigenvectors of $\bZ$ scaled by the leading eigenvalue $\lambda$.  Finally, we calculate the initializer for the original problem by reversing the change of variables as $\bX_0 = \mathbf{U}^{-1}\widetilde{\mathbf{X}}_0 $. 
For most problems the initialization time is a small fraction of the total runtime.
%


%
%

\subsection{Advantages of Biconvex Relaxation}
The proposed framework has numerous advantages over other non-convex methods.  First and foremost, BCR can be applied to general SDPs.  Specialized methods, such as Wirtinger flow \cite{candes2015phase} for phase retrieval and the Wiberg method~\cite{okatani2007wiberg} for low-rank approximation are computationally efficient, but restricted to specific problem types.  Similarly, the max-norm method \cite{lee2010practical} is limited to solving trace-norm-regularized SDPs.  The method of Burer and Montiero \cite{burer2003nonlinear} is less specialized, but does not naturally support inequality constraints.
Furthermore, since BCR problems are biconvex, one can use numerical solvers with guaranteed convergence. Convergence is guaranteed not only for the proposed AM least-squares method in Algorithm \ref{algorithm} (for which the objective decreases monotonically), but also for a broad range of gradient-descent schemes suitable to find solutions to biconvex problems \cite{xu2013block}.  In contrast,  the method in~\cite{burer2003nonlinear} uses augmented Lagrangian methods with non-linear constraints for which convergence is not guaranteed.

\section{Benchmark Problems}
\label{applications}

We now evaluate our solver using both synthetic and real-world data.  We begin with a brief comparison showing that biconvex solvers outperform both interior-point methods for general SDPs and also state-of-the-art low-rank solvers.  Of course, specialized solvers for specific problem forms achieve superior performance to classical interior point schemes.  For this reason, we evaluate our proposed method on three important computer vision applications, i.e., segmentation, co-segmentation, and manifold metric learning, using public datasets, and we compare our results to state-of-the-art methods.  These applications are ideal because (i) they involve large scale SDPs and (ii) customized solvers are available that exploit problem structure to solve these problems efficiently.  Hence, we can compare our BCR framework to powerful and optimized solvers.

\subsection{General-Form Problems}
We briefly demonstrate that BCR performs well on general SDPs by comparing to the widely used SDP
solver, SDPT3~\cite{Toh98sdpt3} and the state-of-the-art, low-rank SDP solver CGDSP~\cite{NIPS2015_5830}. Note that SDPT3 uses an interior point approach to solve the convex problem in \eqref{eq:master} whereas the CGDSP solver uses gradient-descent to solve a non-convex formulation. 
 For fairness, we initialize both algorithms using the proposed initializer and the gradient descent step in CGDSP was implemented using various acceleration techniques~\cite{GoldsteinStuderBaraniuk:2014}. Since CGDSP cannot handle inequality constraints we restrict our comparison to  equality constraints only.

\begin{figure}[!tp]
\centering
\subfloat[Average solver runtime]{
\includegraphics[width=0.49\linewidth, height=0.14\textheight]{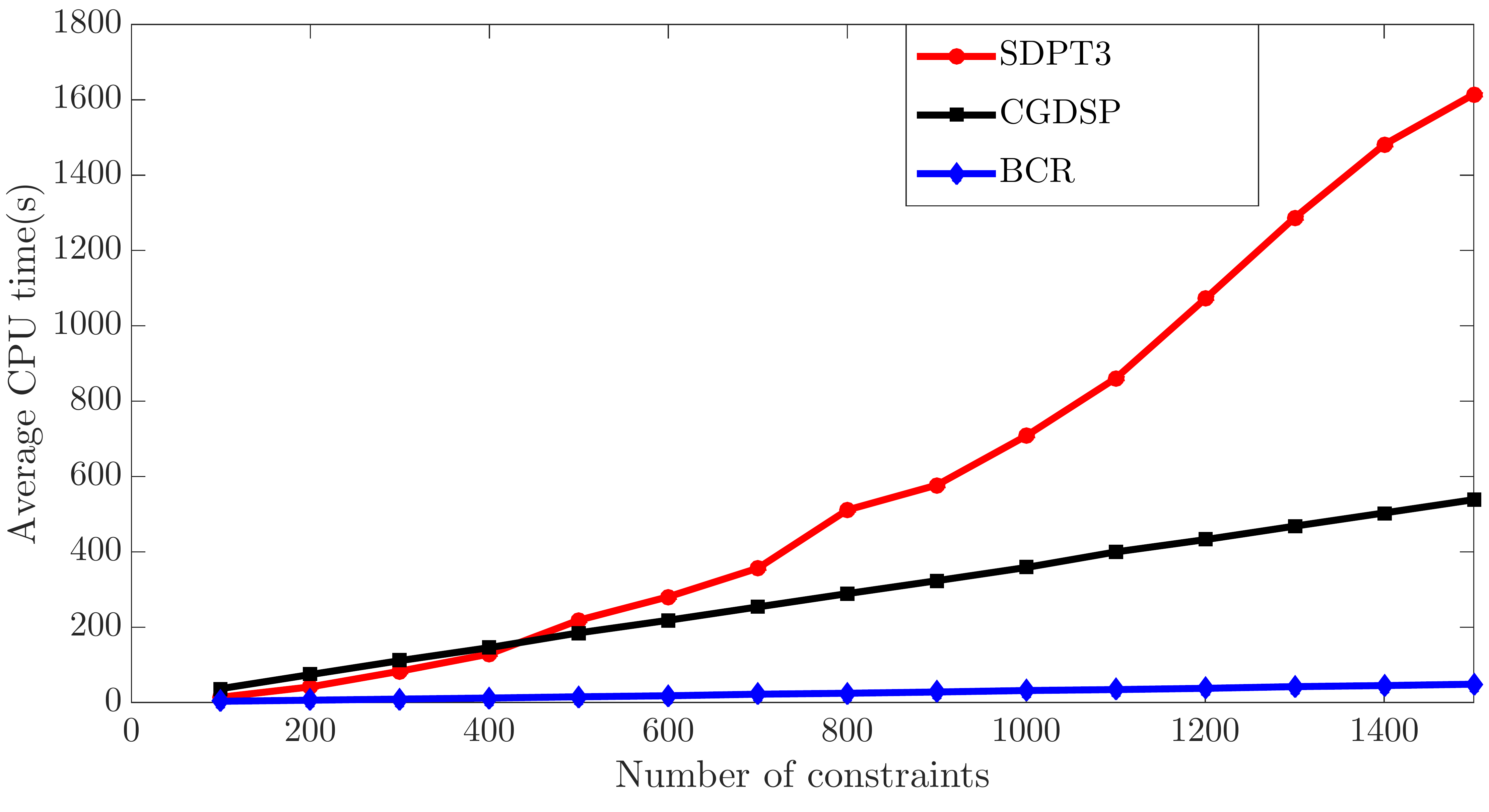}
\label{syn:fig1_a}
}
\subfloat[Average relative error]{
 \includegraphics[width=0.49\linewidth, height=0.14\textheight]{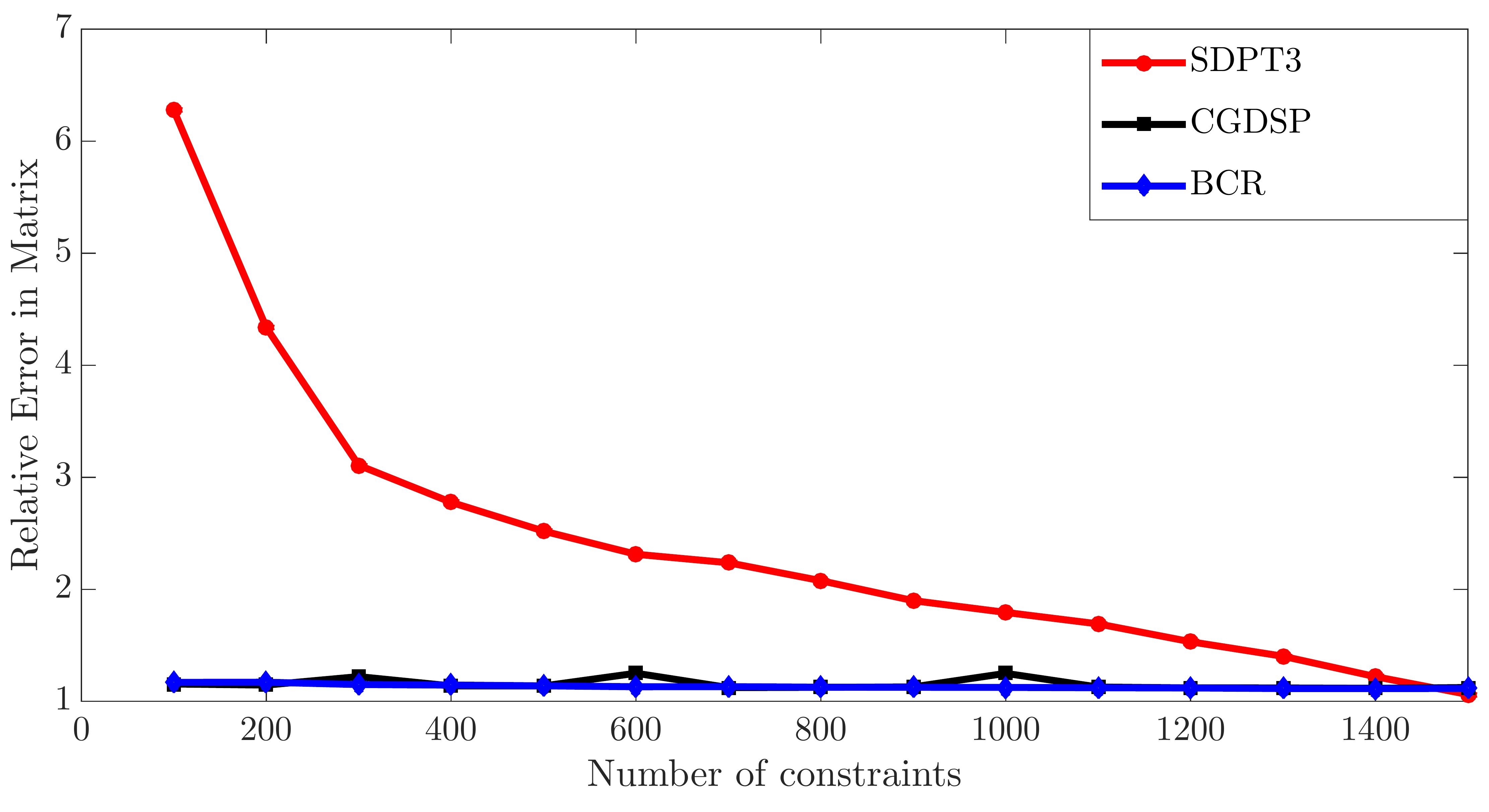}
 \label{syn:fig1_b}
 }
 \caption{\small Results on synthetic data for varying number of linear constraints.}
 \label{syn:fig1}
\end{figure}

{\bf Experiments:}
We randomly generate a $256\times 256$ rank-3 data matrix of the form  $ \mathbf{Y}_\text{true} = \vecx_1\vecx_1^T + \vecx_2\vecx_2^T +\vecx_3 \vecx_3^T,$  where $\{\vecx_i\}$ are standard normal vectors.
We generate a standard normal matrix $\bL$ and compute $\mathbf{C}=\bL^T\bL$. Gaussian matrices $\mathbf{A}_i\in \mathbb{R}^{250\times250}$  form equality constraints. We report the relative error in the recovered solution $ \mathbf{Y}_{\text{rec}}$ measured as
$\| \mathbf{Y}_{\text{rec}}- \mathbf{Y}_{\text{true}}\|/ \|\mathbf{Y}_{\text{true}}\|$. Average runtimes for varying
numbers of constraints are shown in Figure \ref{syn:fig1_a}, while Figure~\ref{syn:fig1_b}
plots the average relative error. Figure \ref{syn:fig1_a} shows that our method has the best runtime of all the schemes.   Figure \ref{syn:fig1_b} shows convex interior point methods do not recover the correct solution for small numbers of constraints.  With few constraints, the full lifted SDP is under-determined, allowing the objective to go to zero.  In contrast, the proposed BCR approach is able to enforce an additional rank-3 constraint, which is advantageous when the number of constraints is low. 

\subsection{Image Segmentation} 
Consider an image of $N$ pixels. Segmentation of foreground and background objects can be accomplished using graph-based approaches, where graph edges encode the similarities between pixel pairs. Such approaches include normalized cut~\cite{shi2000normalized} and ratio cut \cite{ratiocut}. The graph cut problem can be formulated as an NP-hard integer program \cite{goemans1995improved}
\begin{equation}
\small
{
\begin{aligned}
\label{imsegeq1}
\underset{\mathbf{x}\in \{-1,1\}^N}{\text{minimize}} \,\, \mathbf{x}^T \mathbf{L} \mathbf{x},
\end{aligned}
}
\end{equation}
where $\mathbf{L}$ encodes edge weights and $\mathbf{x}$ contains binary region labels, one for each pixel.    
This problem can be ``lifted'' to the equivalent higher dimensional problem 
\begin{equation}
{\small
  \begin{aligned}
\underset{\mathbf{X}\in S^+_{N\times N}}{\text{minimize}} \,\,\trace(\mathbf{L}^T \mathbf{X}) \quad
\text{subject to} \enspace 
\textrm{diag}(\mathbf{X}) = \mathbf{1}, \enspace
\textrm{rank}(\mathbf{X}) = 1.
\end{aligned}  \label{imsegeq3}
}
\end{equation}
%
After dropping the non-convex rank constraint, \eqref{imsegeq3} becomes an SDP that is solvable using convex optimization ~\cite{heiler2005semidefinite,keuchel2003binary,schellewald2005probabilistic}. The SDP approach is computationally intractable if solved using off-the-shelf SDP solvers (such as SDPT3 \cite{Toh98sdpt3} or other interior point methods).  Furthermore, exact solutions cannot be recovered when the solution to the SDP has rank greater than 1.  In contrast, BCR is computational efficient for large problems and can easily incorporate rank constraints, leading to efficient spectral clustering. 

BCR is also capable of incorporating annotated foreground and background pixel priors~\cite{yu2004segmentation} using linear equality and inequality constraints. We consider the SDP based segmentation presented in~\cite{yu2004segmentation}, which contains three grouping constraints on the pixels: $(\mathbf{t}_f^T \mathbf{P} \mathbf{x})^2 \geq \kappa \|\mathbf{t}_f^T \mathbf{P} \mathbf{x}\|_1^2$, $(\mathbf{t}_b^T \mathbf{P} \mathbf{x})^2 \geq \kappa \|\mathbf{t}_b^T \mathbf{P} \mathbf{x}\|_1^2$ and $( ( \mathbf{t}_f - \mathbf{t}_b)^T \mathbf{P} \mathbf{x})^2 \geq \kappa \|(\mathbf{t}_f - \mathbf{t}_b)^T \mathbf{P} \mathbf{x}\|_1^2$, where $\kappa \in [0,1]$. $\mathbf{P} = \mathbf{D}^{-1} \mathbf{W}$ is the normalized pairwise affinity matrix and $\mathbf{t}_f$ and $\mathbf{t}_b$ are indicator variables denoting the foreground and background pixels. These constraints enforce that the segmentation respects the pre-labeled pixels given by the user, and also pushes high similarity pixels to have the same label. 
The affinity matrix $\mathbf{W}$ is given by
\begin{equation}
 {W}_{i,j} = \begin{cases} \exp\! \left(-\frac{\|\mathbf{f}_i-\mathbf{f}_j\|_2^2}{\gamma_f^2} - \frac{d(i,j)^2}{\gamma_d^2} \right)\!, & \!\!\!\mbox{if } d(i,j) < r \\ 0,  \quad \mbox{otherwise,} \end{cases} \label{imsegeq4}
\end{equation}
where $\mathbf{f}_i$ is the color histogram of the $i$th super-pixel and $d(i,j)$ is the spatial distance between~$i$ and~$j$. 
Considering these constraints and letting $\mathbf{X}=\mathbf{Y}\mathbf{Y}^T$,~\eqref{imsegeq3} can be written in the form of \eqref{no_q}  as follows:
\begin{equation}
\begin{aligned}
& \underset{\mathbf{Y} \in \mathbb{R}^{N\times r}}{\text{minimize}}
& & \trace(\mathbf{Y}^T \mathbf{L}\mathbf{Y}) \\
& \text{subject to}
& & \trace(\mathbf{Y}^T\mathbf{A}_i\mathbf{Y}) = 1, \quad \forall i=1,\ldots,N \\
& & &  \trace(\mathbf{Y}^T\mathbf{B}_2\mathbf{Y}) \geq \kappa \|\mathbf{t}_f^T \mathbf{P} \mathbf{x}\|_1^2, \,\,
\trace(\mathbf{Y}^T\mathbf{B}_3\mathbf{Y}) \geq \kappa\|\mathbf{t}_b^T \mathbf{P} \mathbf{x}\|_1^2\\
& & &  \trace(\mathbf{Y}^T\mathbf{B}_4\mathbf{Y}) \geq \kappa\|(\mathbf{t}_f - \mathbf{t}_b)^T \mathbf{P} \mathbf{x}\|_1^2, \,\, \trace(\mathbf{Y}^T\mathbf{B}_1\mathbf{Y}) = 0. \\
\end{aligned}
\label{imsegeq5}
\end{equation}
Here, $r$ is the rank of the desired solution, $\mathbf{B}_1 = \mathbf{1}\mathbf{1}^T$, $\mathbf{B}_2 = \mathbf{P}\mathbf{t}_f\mathbf{t}_f^T\mathbf{P}$, $\mathbf{B}_3 = \mathbf{P}\mathbf{t}_b\mathbf{t}_b^T\mathbf{P}$, $\mathbf{B}_4 = \mathbf{P}(\mathbf{t}_f - \mathbf{t}_b)(\mathbf{t}_f - \mathbf{t}_b)^T\mathbf{P}$, $\mathbf{A}_i = \mathbf{e}_i\mathbf{e}_i^T$, $\mathbf{e}_i \in \mathbb{R}^{n}$ is an elementary vector with a 1 at the $i$th position. 
After solving \eqref{imsegeq5} using BCR \eqref{biconvex}, the final binary solution is extracted from the score vector using the swept random hyperplanes method \cite{lang2005fixing}.

We compare the performance of BCR with the highly customized BQP solver SDCut~\cite{Wang_2013} and biased normalized cut (BNCut) \cite{biasedncut}. BNCut is an extension of the Normalized cut algorithm \cite{shi2000normalized} whereas SDCut is currently the most efficient and accurate SDR solver but limited only to solving BQP problems. 
Also, BNCut can support only one quadratic grouping constraint per problem.

{\bf Experiments:}
We consider the Berkeley image segmentation dataset~\cite{MartinFTM01}. Each image is segmented into super-pixels using the VL-Feat \cite{vedaldi2012vlfeat} toolbox. For SDCut and BNCut, we use the publicly available code with hyper-parameters set to the values suggested in \cite{Wang_2013}. For BCR, we set $\beta = \lambda / \sqrt{|\mathcal{B} \cup \setE |}$, where $\lambda$ controls the coarseness of the segmentation by mediating the tradeoff between the objective and constraints, and would typically be chosen from $[1,10]$ via cross validation.   For simplicity, we just set $\lambda = 5$ in all experiments reported here. 

\begin{figure*}[!tp]
   \centering
   \begin{tabular}{c c c c c c }
     \raisebox{0.5cm}{Original} &\includegraphics[width=0.16\linewidth]{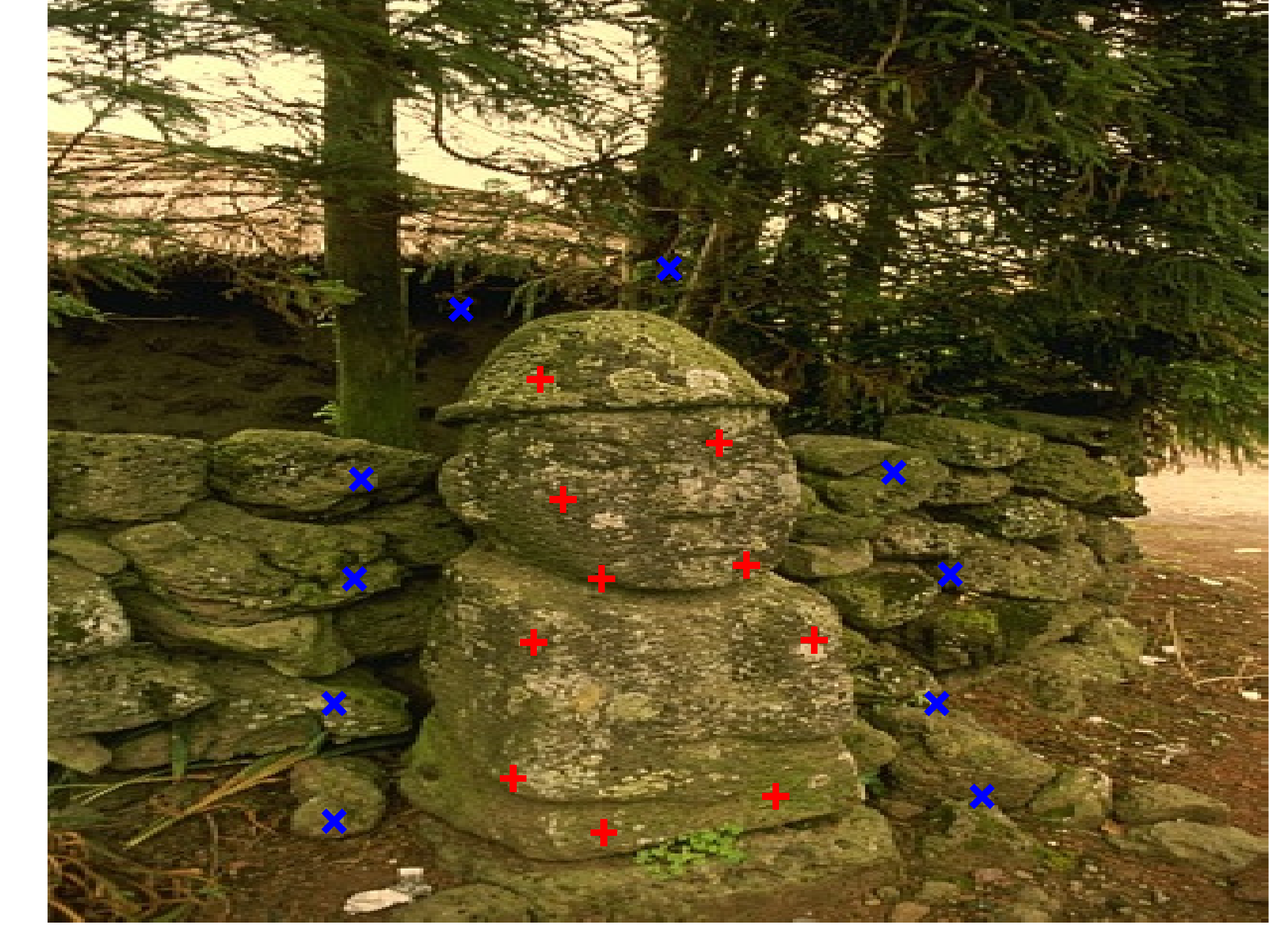}&
     \includegraphics[width=0.16\linewidth]{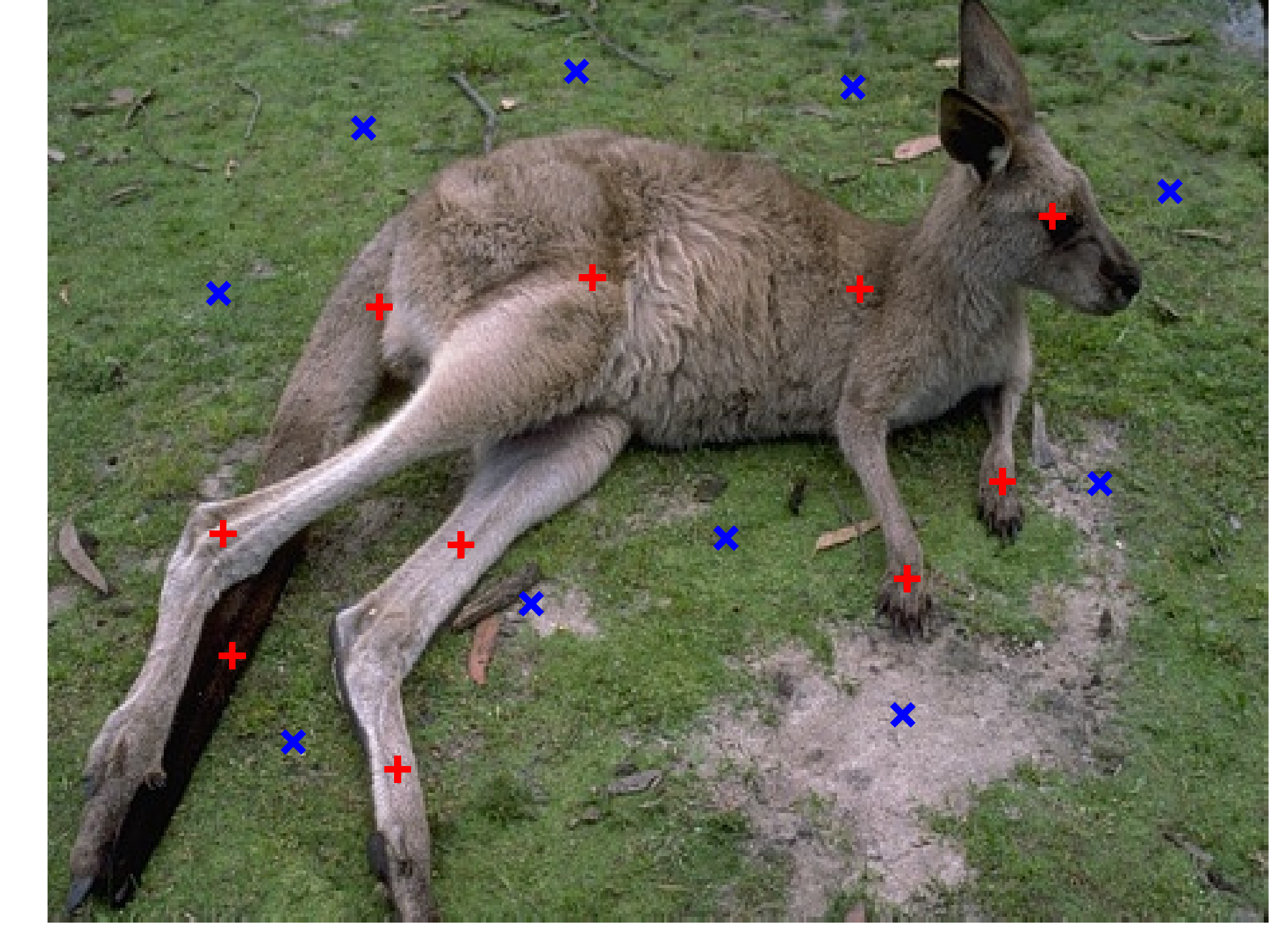}&
     \includegraphics[width=0.16\linewidth]{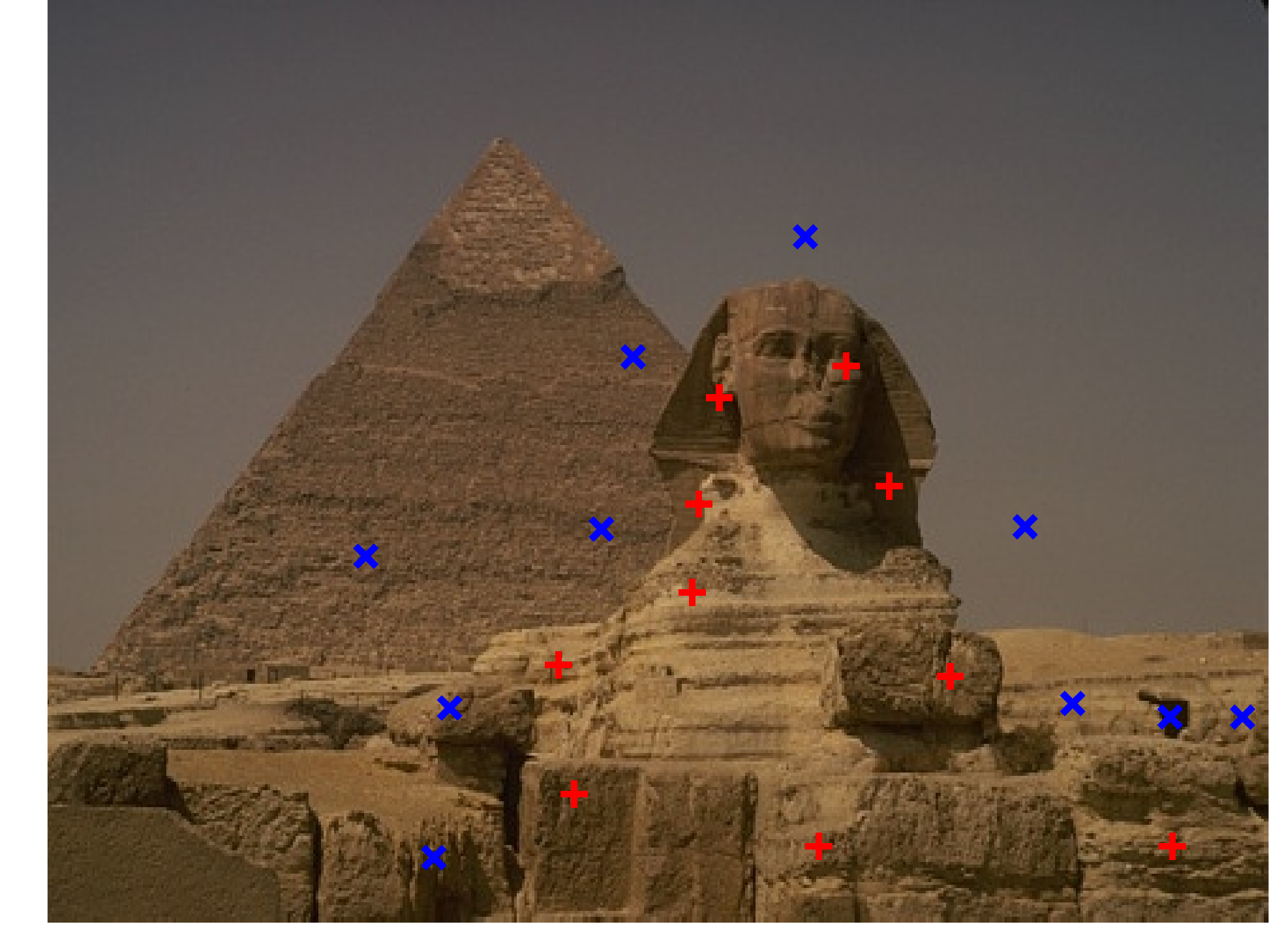}&
     \includegraphics[width=0.16\linewidth]{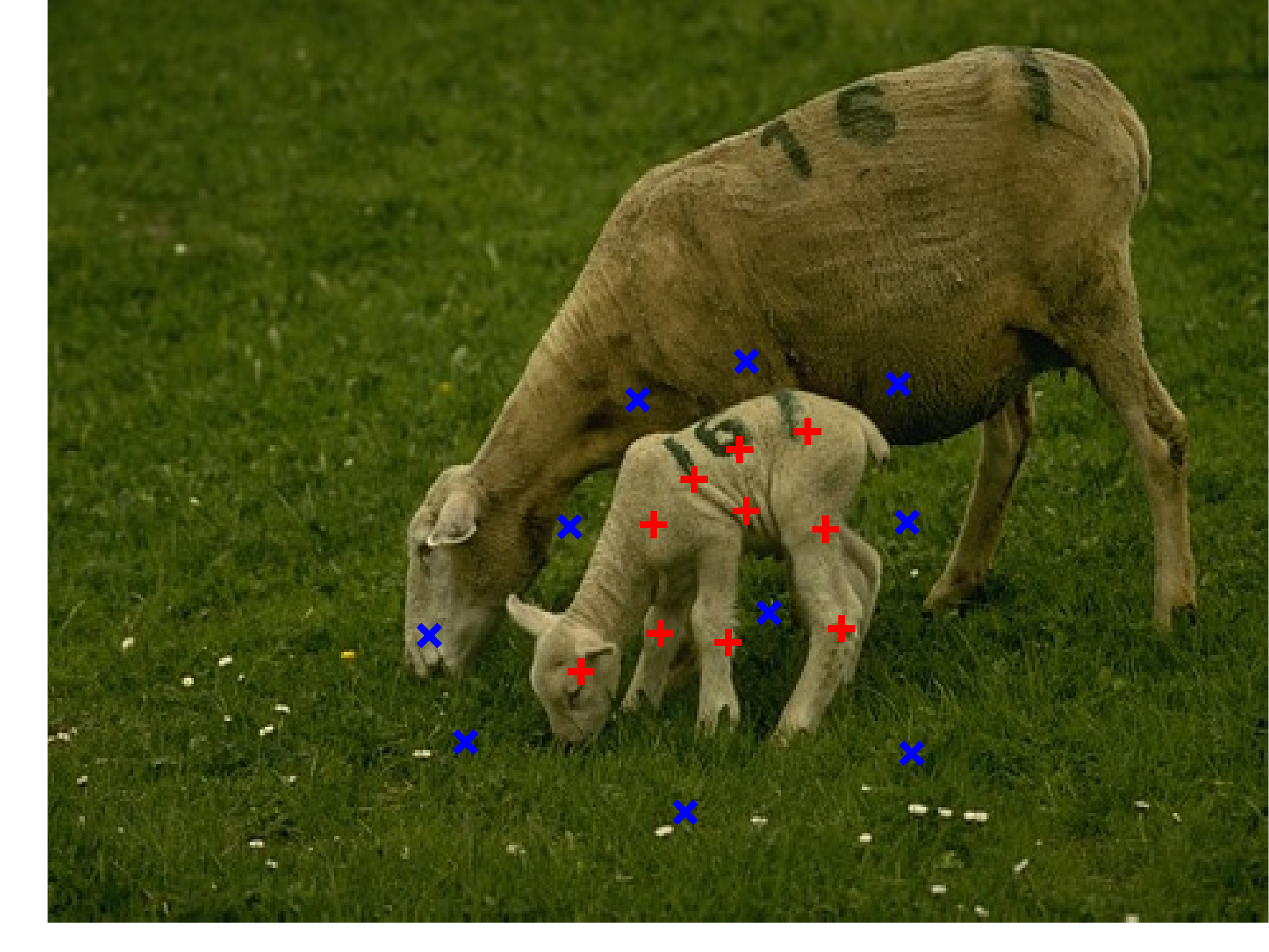}&
     \includegraphics[width=0.16\linewidth]{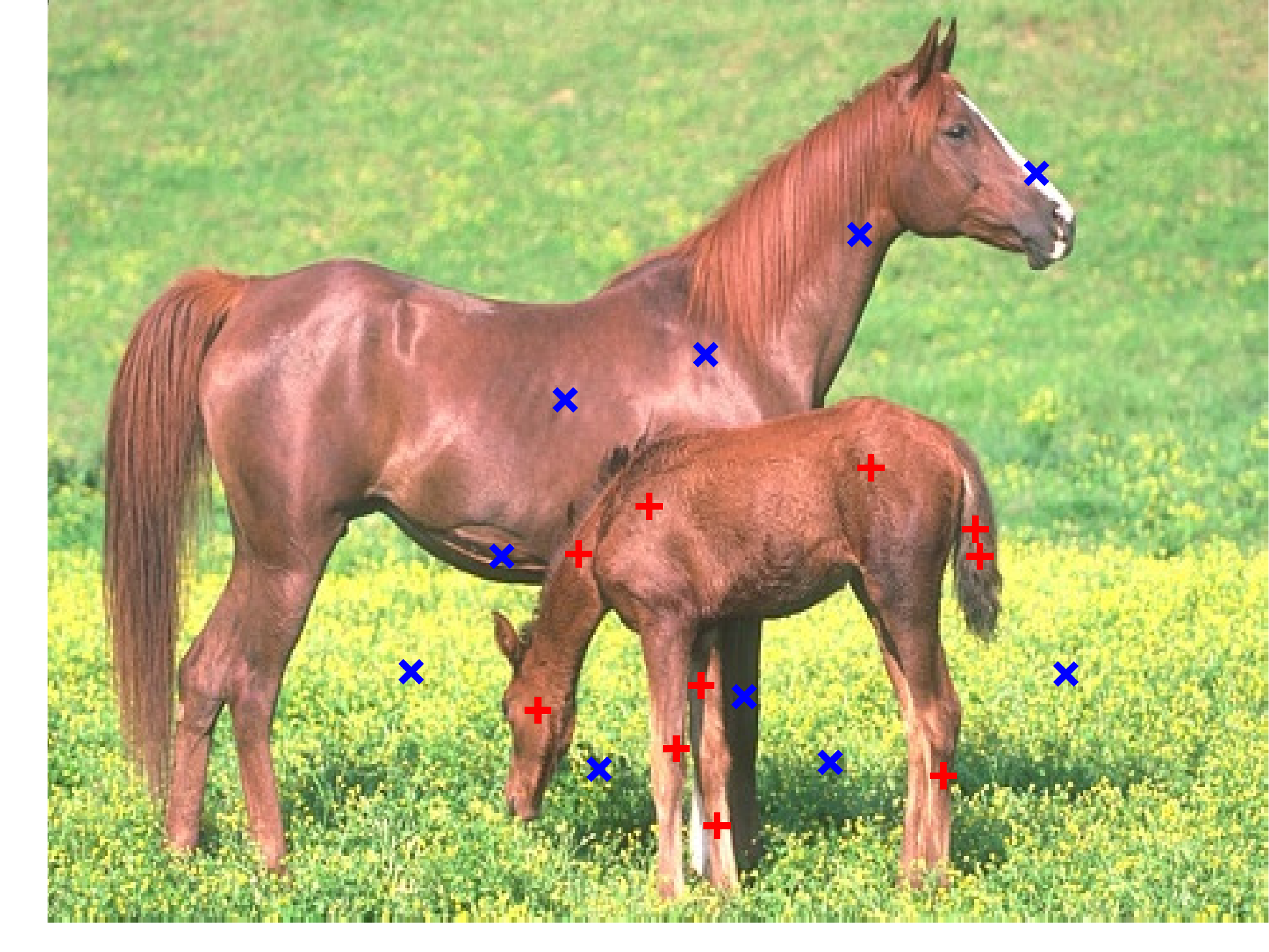}\\[-1ex]
     \raisebox{0.5cm}{BNCut} & \includegraphics[width=0.16\linewidth]{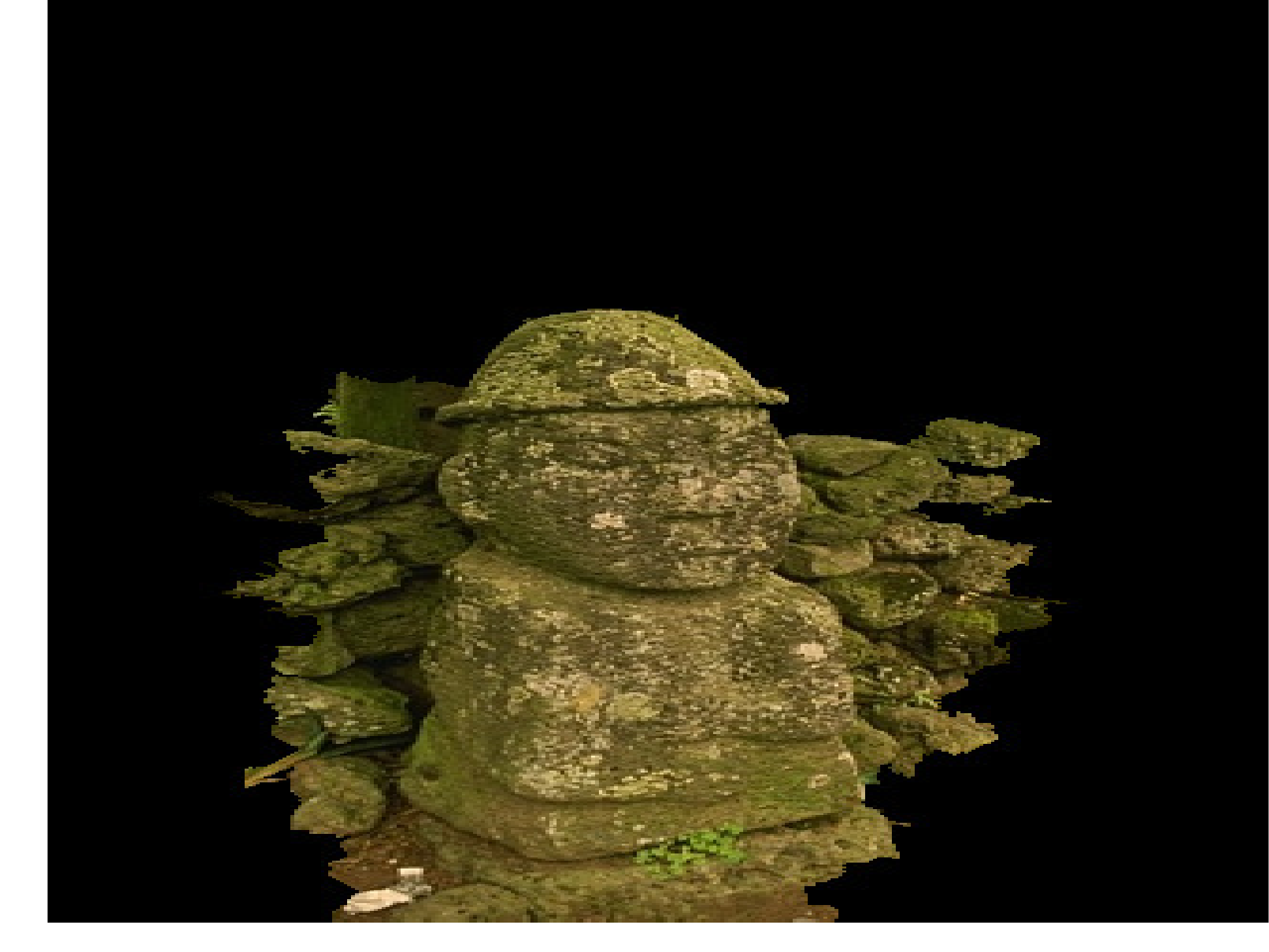}&
     \includegraphics[width=0.16\linewidth]{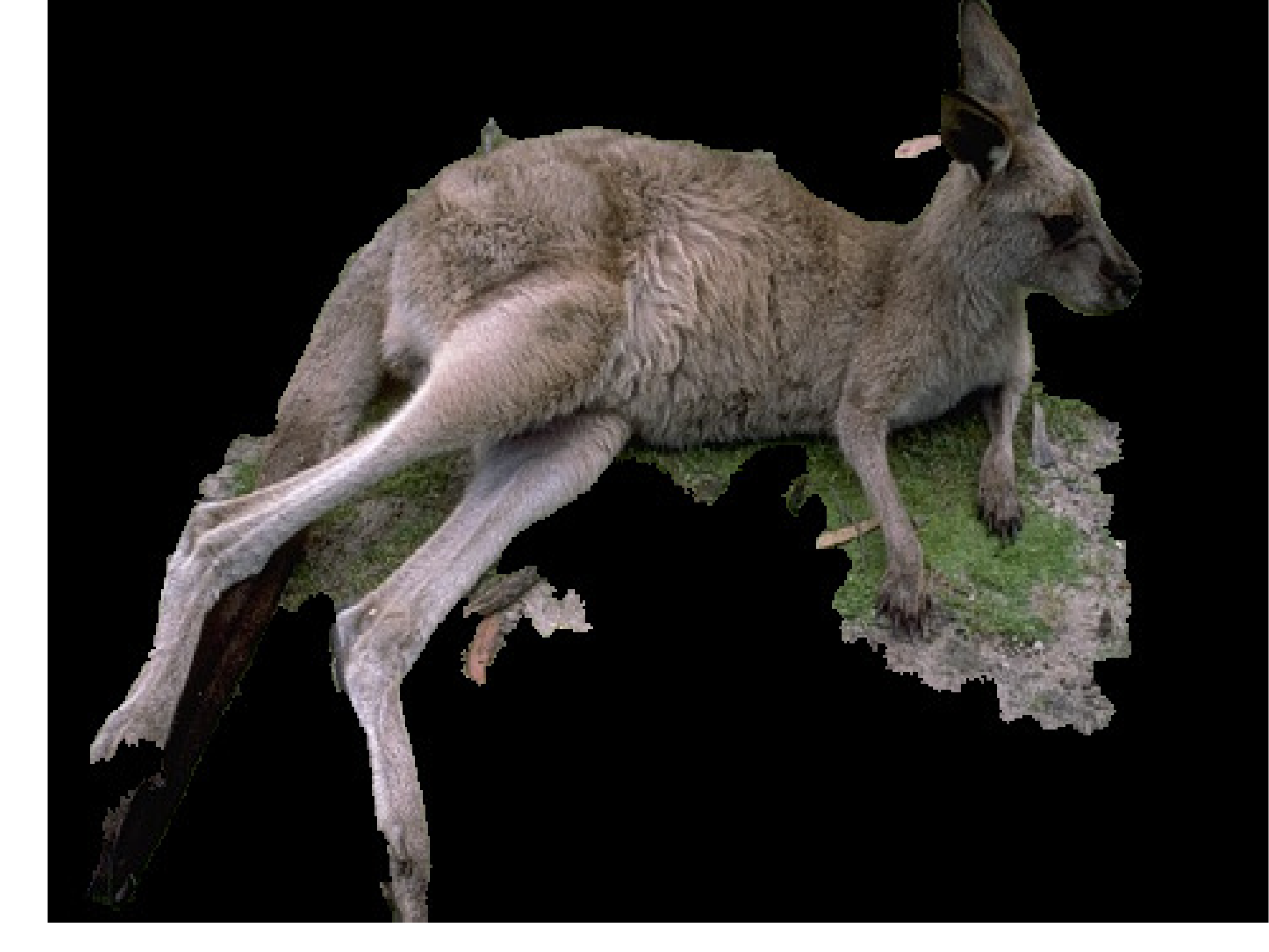}&
     \includegraphics[width=0.16\linewidth]{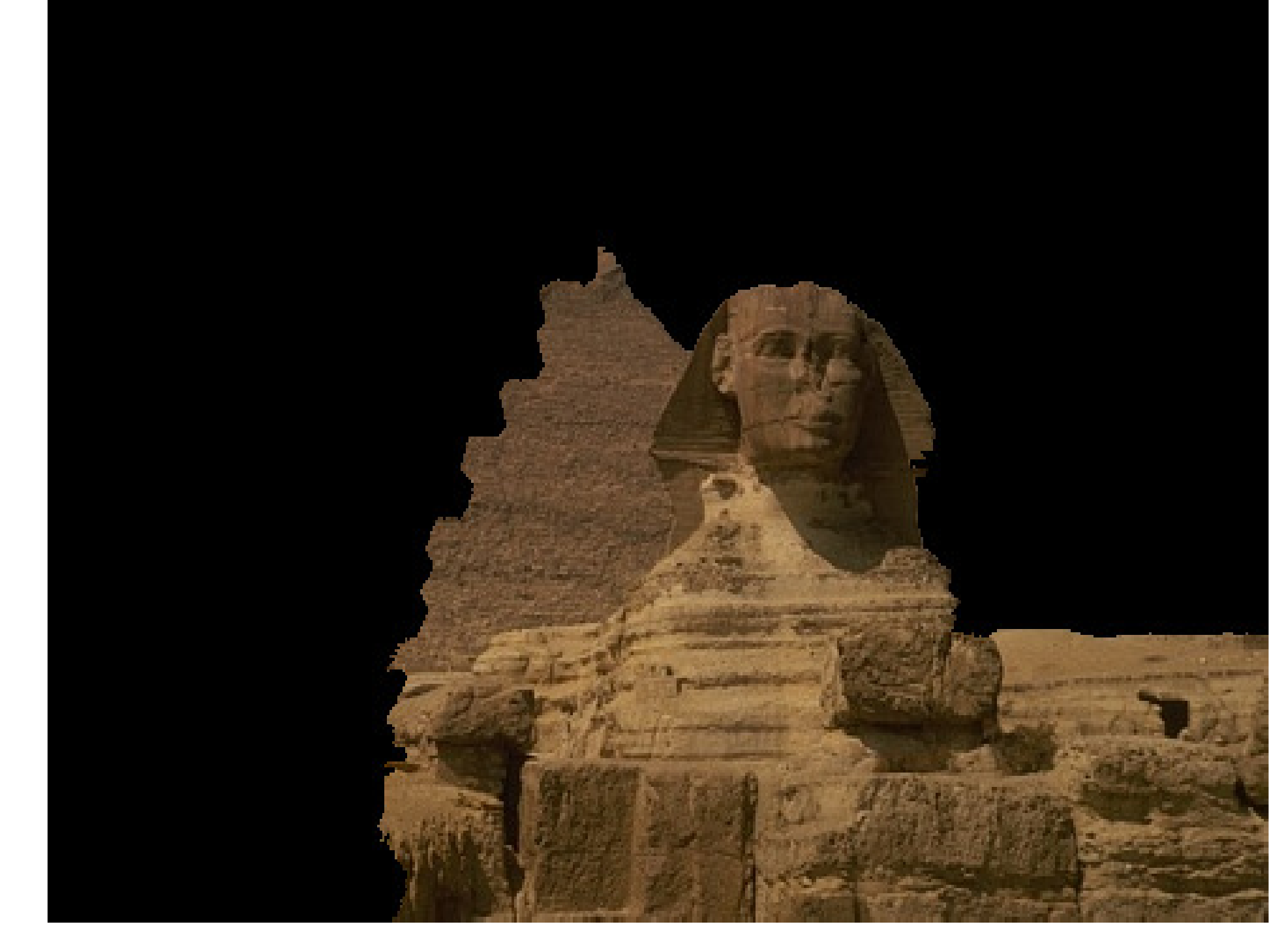}&
     \includegraphics[width=0.16\linewidth]{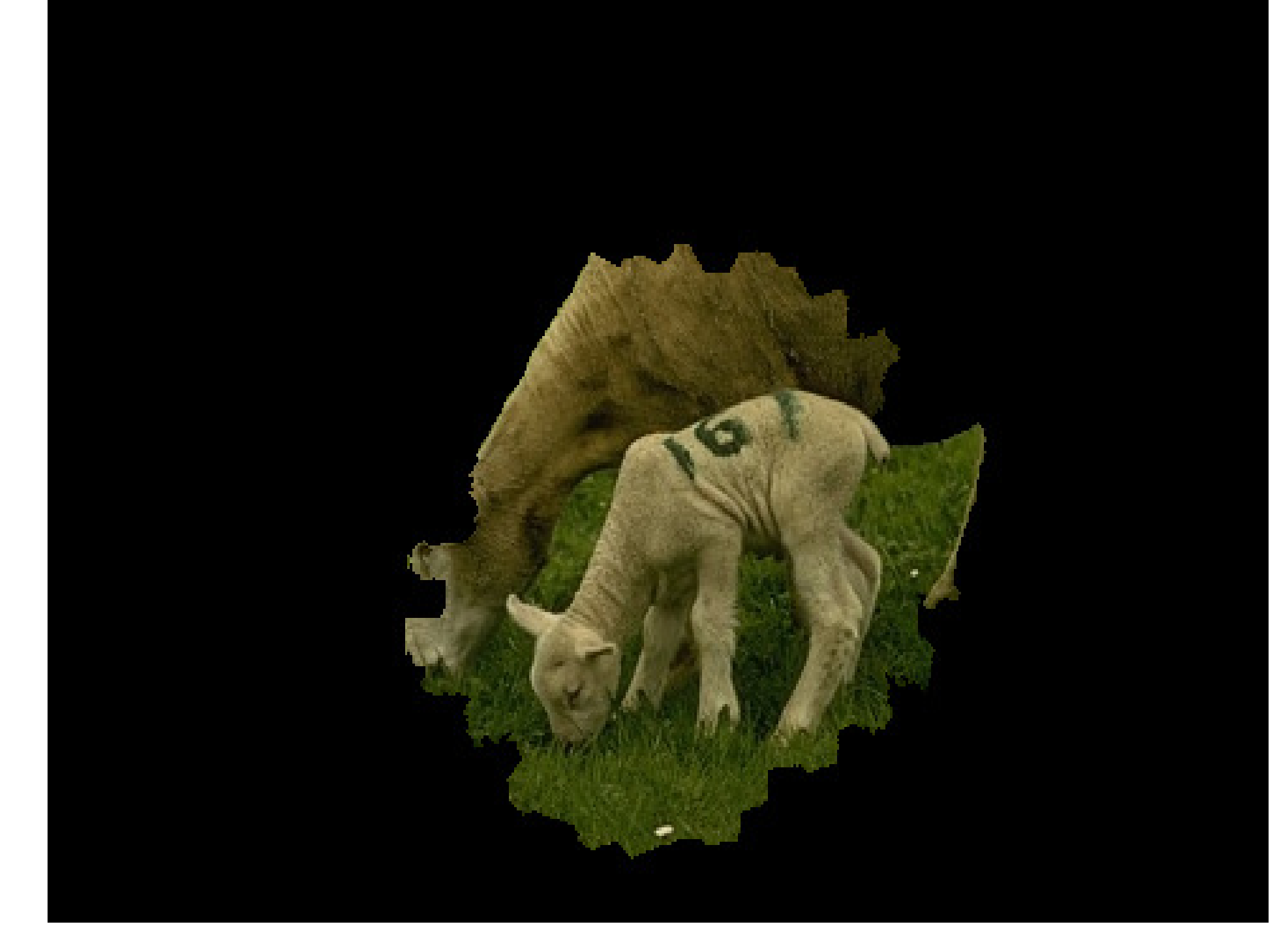}&
     \includegraphics[width=0.16\linewidth]{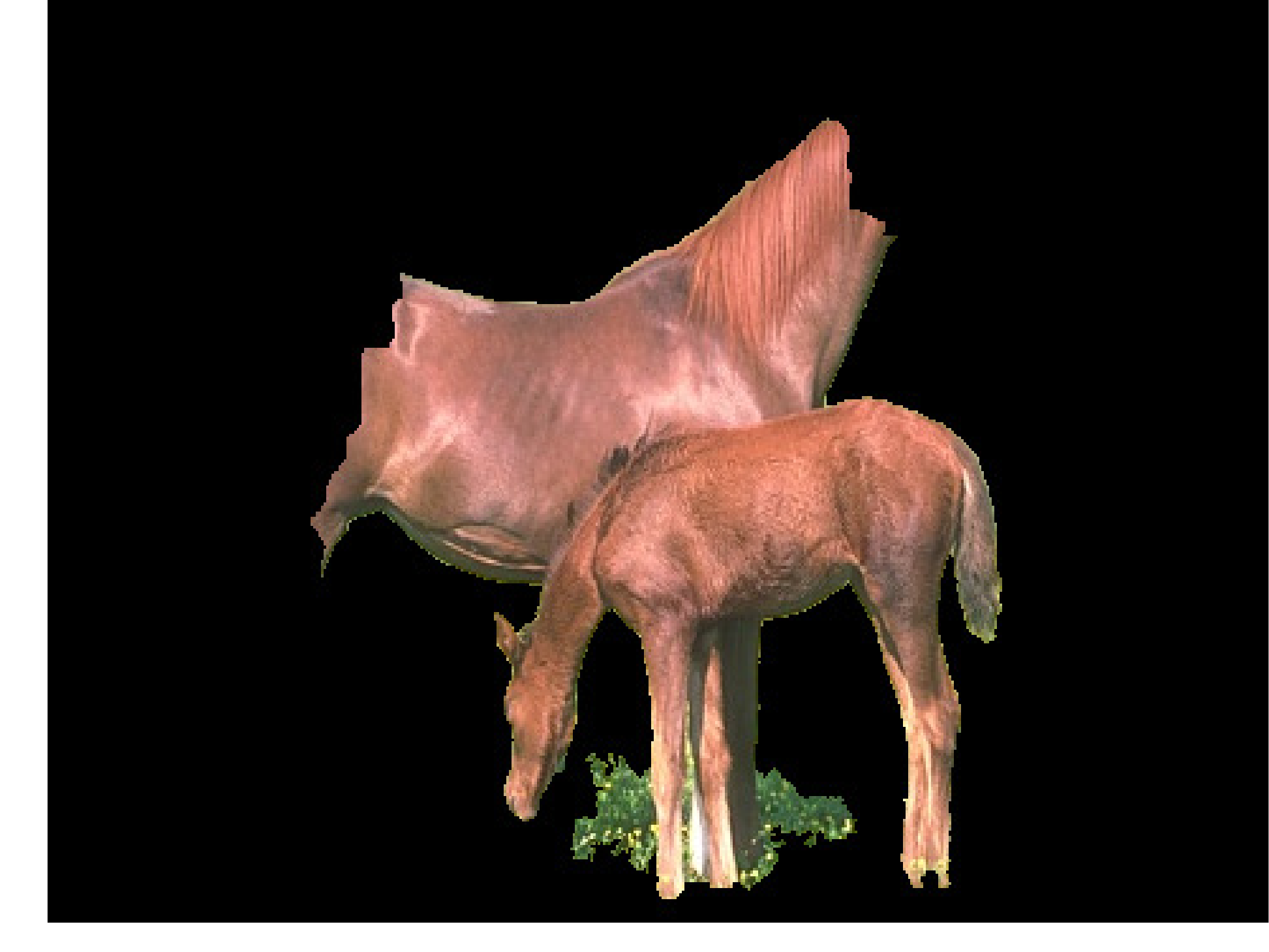}\\[-1ex]
     \raisebox{0.5cm}{SDCut} & \includegraphics[width=0.16\linewidth]{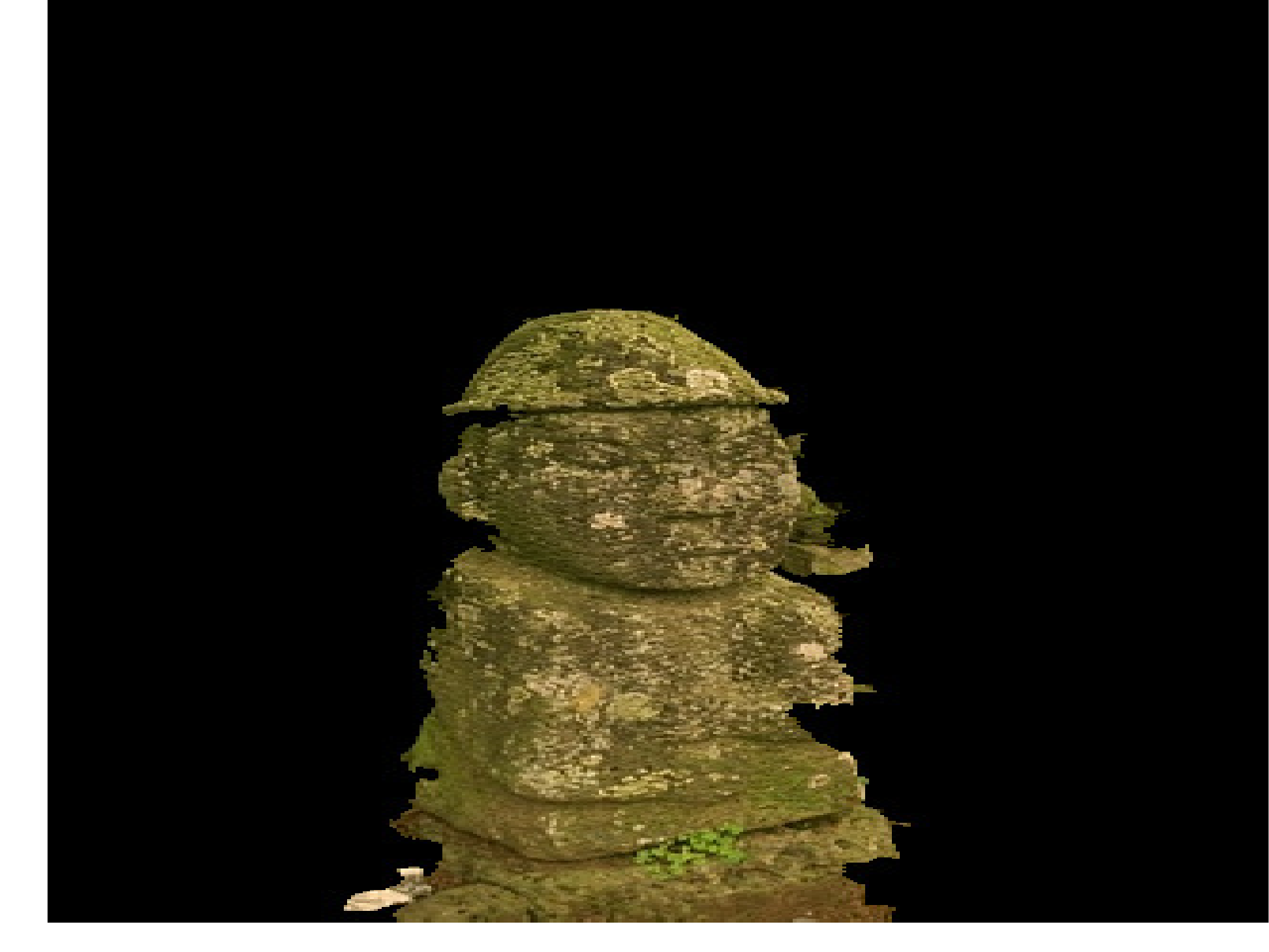}&
     \includegraphics[width=0.16\linewidth]{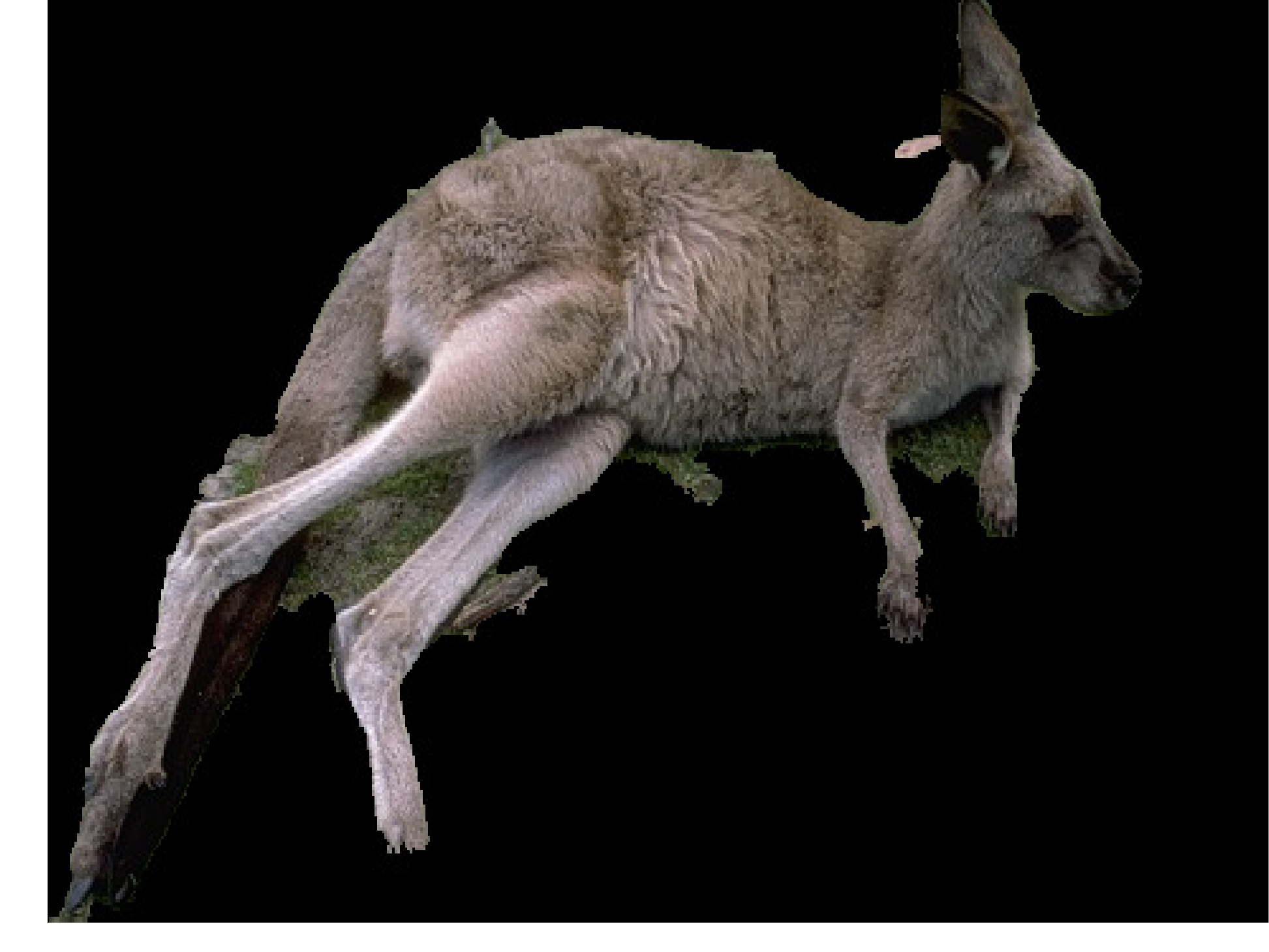}&
     \includegraphics[width=0.16\linewidth]{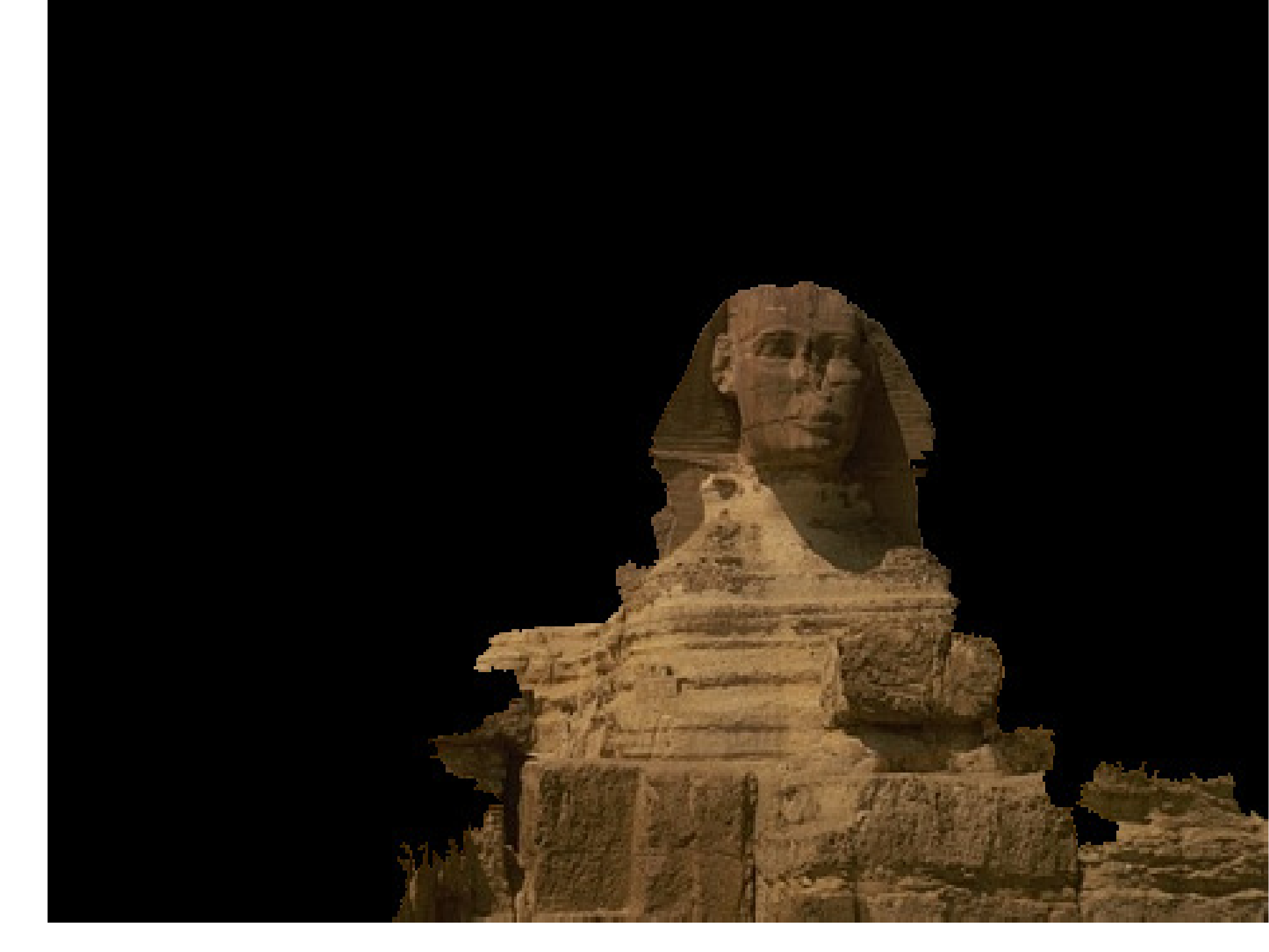}&
     \includegraphics[width=0.16\linewidth]{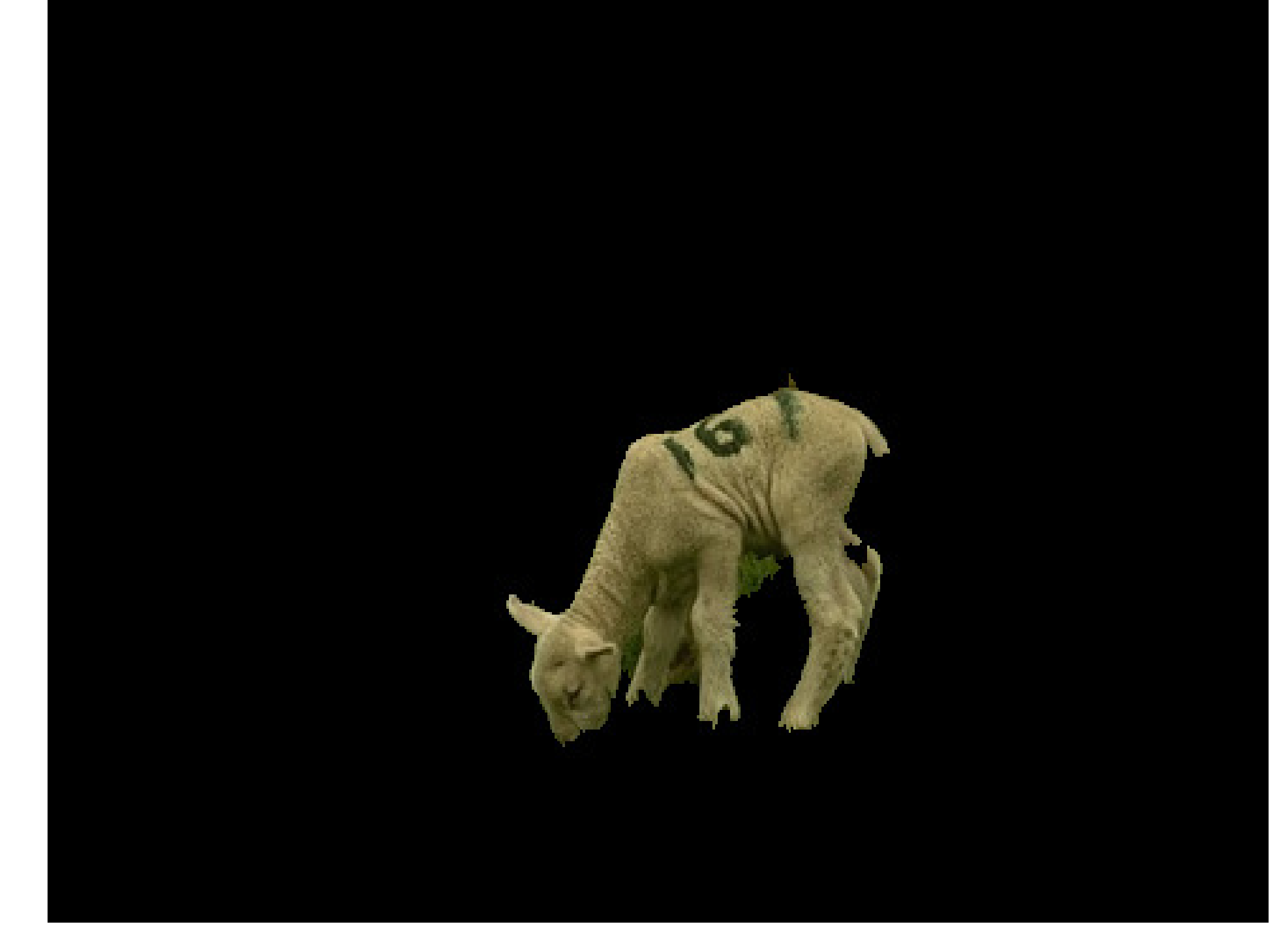}&
     \includegraphics[width=0.16\linewidth]{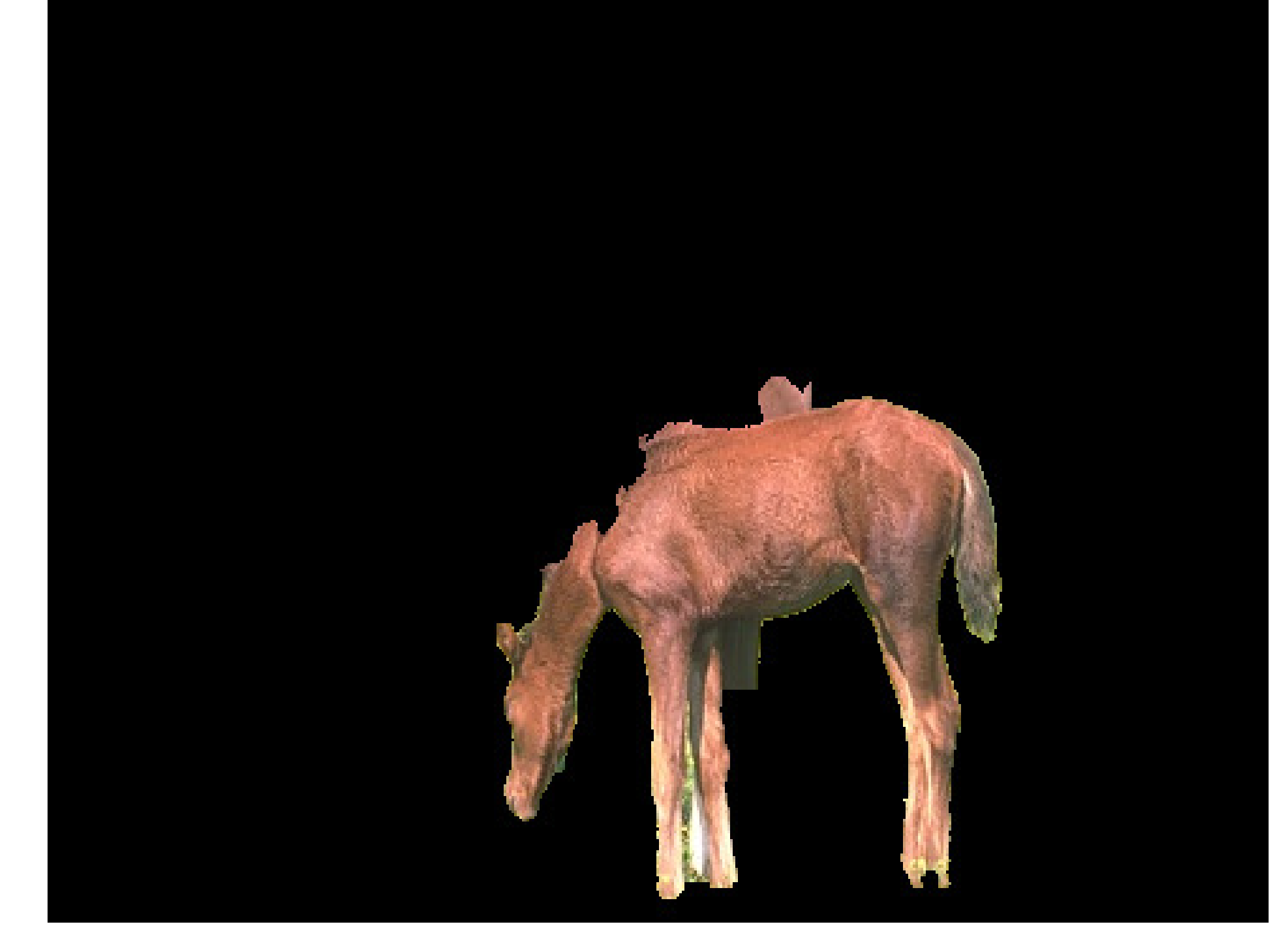}\\[-1ex]
     \raisebox{0.5cm}{BCR} & \includegraphics[width=0.16\linewidth]{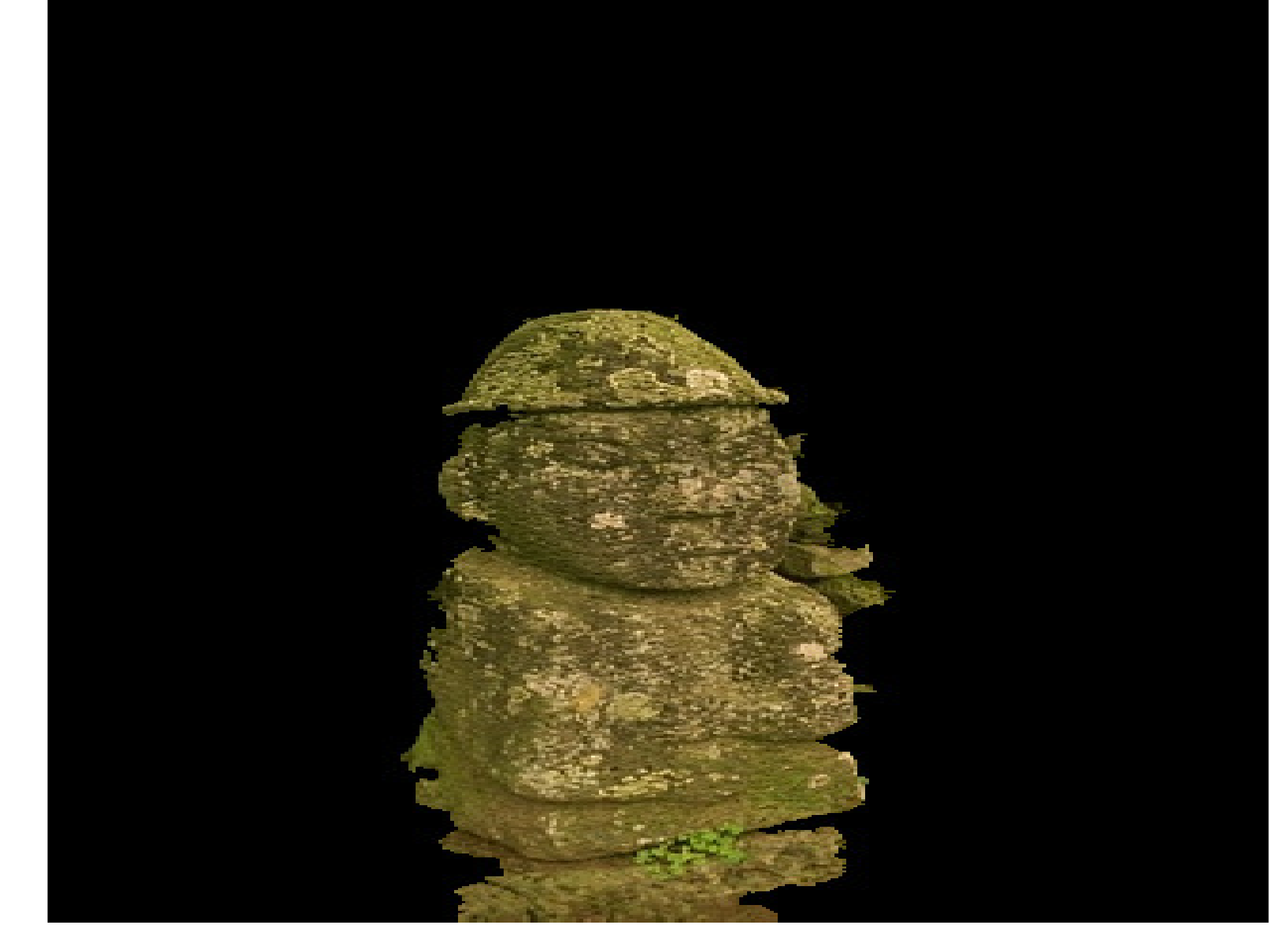}& 
     \includegraphics[width=0.16\linewidth]{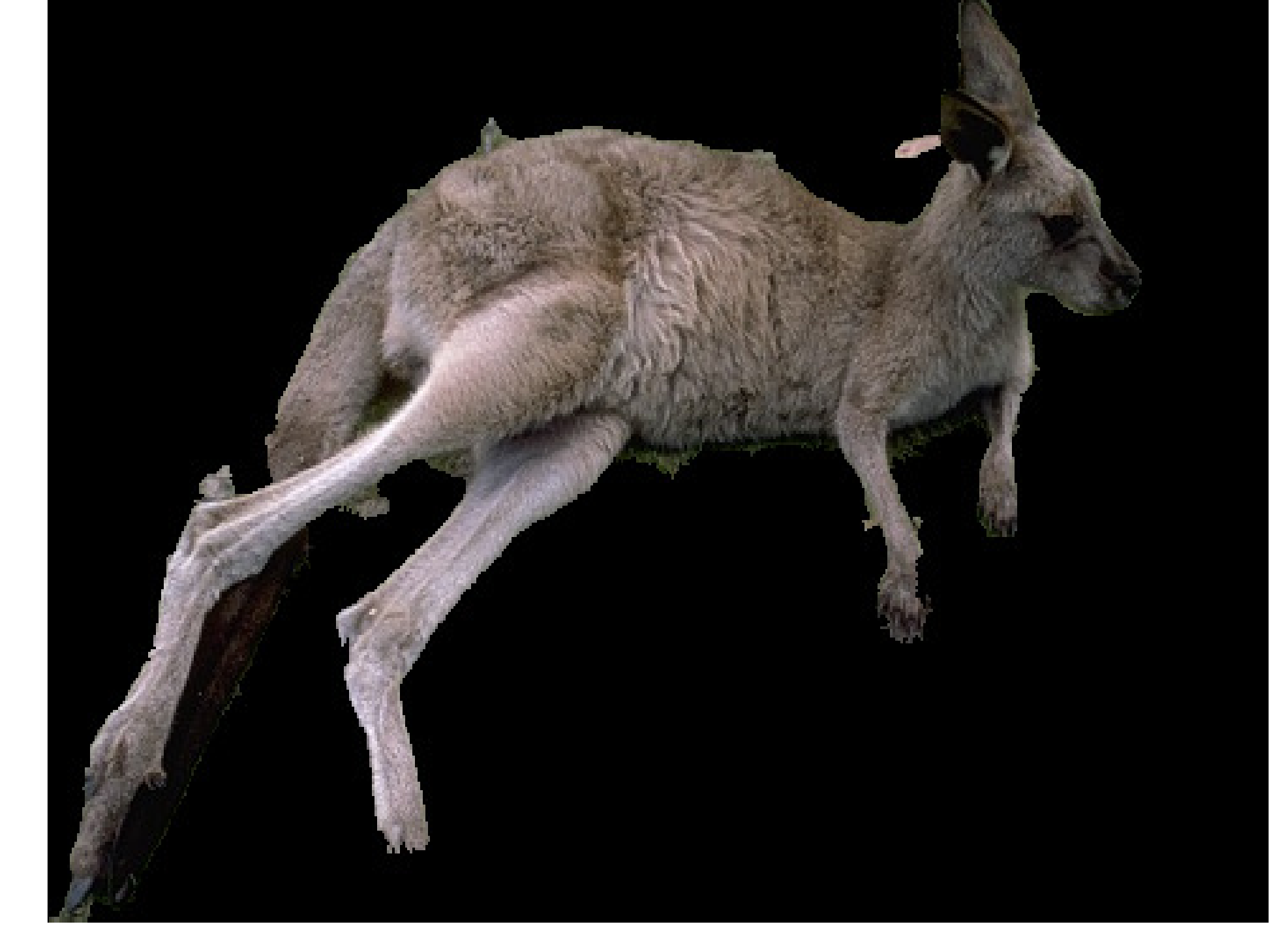}&
     \includegraphics[width=0.16\linewidth]{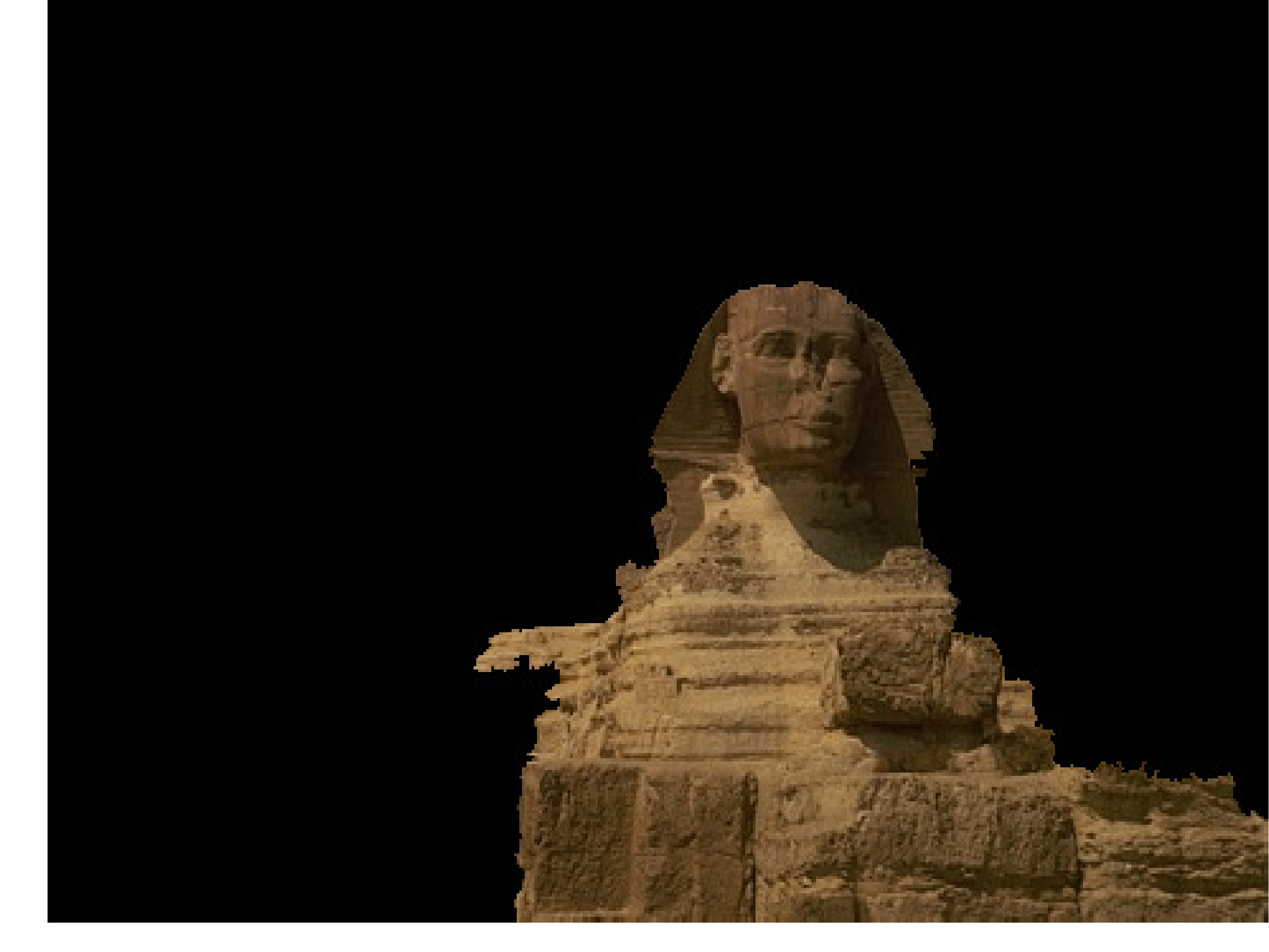}&
     \includegraphics[width=0.16\linewidth]{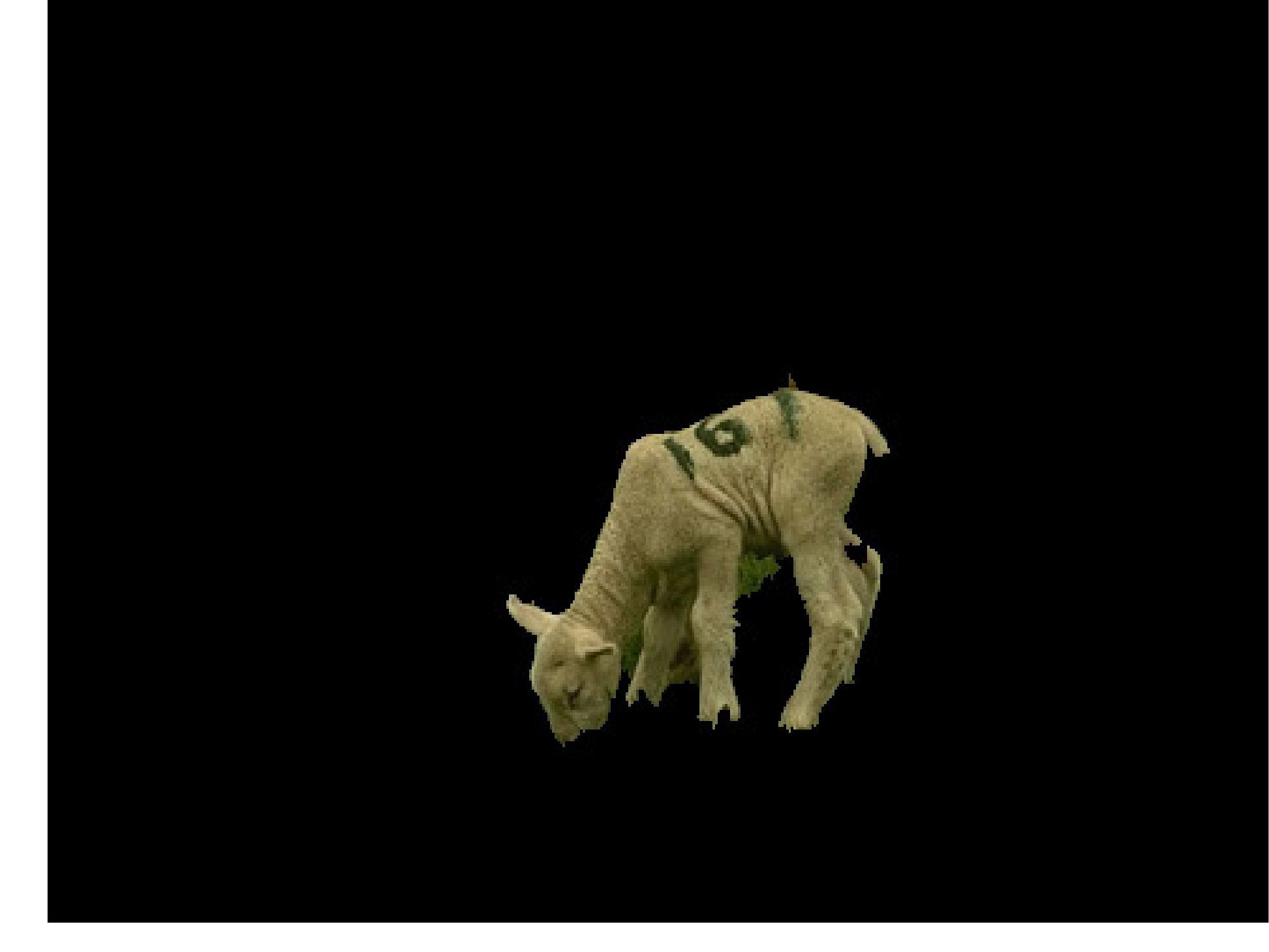}&
     \includegraphics[width=0.16\linewidth]{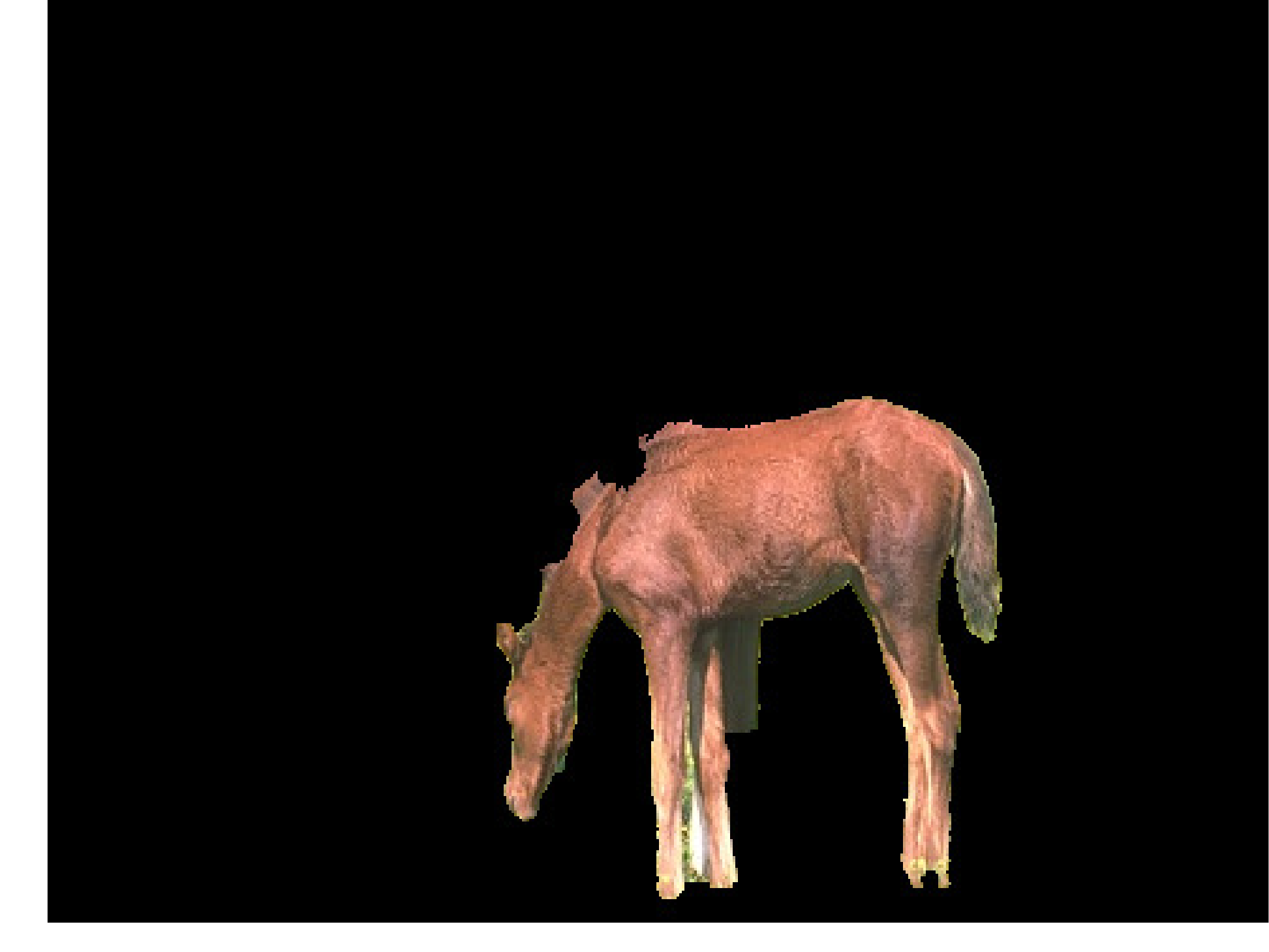} \\[-1ex]
   \end{tabular}
   \caption{\small Image segmentation results on the Berkeley dataset. The red and blue marker indicates the annotated foreground and background super-pixels, respectively.}
   \label{fig:seg1}
 \end{figure*}

We compare the runtime and quality of each algorithm. 
Figure~\ref{fig:seg1} shows the segmentation results while the quantitative results are displayed in Table \ref{tab:seg1}. For all the considered images, our approach gives superior foreground object segmentation compared to SDCut and BNCut. Moreover, as seen in Table~\ref{tab:seg1}, our solver is $35\times$ faster than SDCut and yields lower objective energy. Segmentation using BCR is achieved using only rank $2$ solutions whereas SDCut requires rank 7 solutions to obtain results of comparable accuracy.\footnote{The optimal solutions found by SDCut all had rank $7$ except for one solution of rank $5$.} Note that while BNCut with rank $1$ solutions is much faster than SDP based methods, the BNCut segmentation results are not on par with SDP approaches.

\subsection{Co-segmentation} 
We next consider image co-segmentation, in which segmentation of the same object is jointly computed on multiple images simultaneously.  Because co-segmentation involves multiple images, it provides a testbed for large problem instances. Co-segmentation balances a tradeoff between two criteria: (i) color and spatial consistency within a single image and (ii) discrimination between foreground and background pixels over multiple images. We closely follow the work of Joulin \textit{et al.} \cite{Joulin_2010}, whose formulation is given by
\begin{align}
\underset{\mathbf{x}\in\{\pm1\}^N}{\text{minimize}} \,\, \mathbf{x}^T \mathbf{A} \mathbf{x} \quad
\text{subject to} \enspace (\mathbf{x}^T\delta_i)^2 \leq \lambda^2, \quad \forall i = 1,\dots,M, \label{coseg:eq1} 
\end{align}
where $M$ is the number of images and $N = \sum_{i=1}^M N_i$ is the total number of pixels over all images. 
The matrix $\mathbf{A} = \mathbf{A}_b + \frac{\mu}{N} \mathbf{A}_w,$  where $\mathbf{A}_w$ is the intra-image affinity matrix and $\mathbf{A}_b$ is the inter-image discriminative clustering cost matrix computed using the $\chi^2$ distance between SIFT features in different images (see \cite{Joulin_2010} for a details).
%
%

 To solve this problem with BCR, we re-write~\eqref{coseg:eq1} in the form \eqref{no_q} to obtain
\begin{equation}
\begin{aligned}
& \underset{\mathbf{X} \in \mathbb{R}^{N\times r}}{\text{minimize}}
& & \trace(\mathbf{X}^T \mathbf{A}\mathbf{X}) \\
& \text{subject to:}
& & \trace(\mathbf{X}^T\mathbf{Z}_i\mathbf{X}) = 1, \quad \forall i=1,\ldots,N\\
& & & \trace(\mathbf{X}^T\mathbf{\Delta}_i\mathbf{X}) \leq \lambda^2, \,\, \forall i=1,\ldots,M,
\end{aligned}
\label{coseg:eq2}
\end{equation}
where $\mathbf{\Delta}_i = \mathbf{\delta_i} \mathbf{\delta_i}^T$ and $\mathbf{Z}_i=\mathbf{e}_i\mathbf{e}_i^T$. 
%
%
%
%
Finally, \eqref{coseg:eq2} is solved using BCR \eqref{biconvex}, following which one can recover the optimal score vector $\mathbf{x}_p^*$ as the leading eigenvector of $\mathbf{X^*}.$ The final binary solution is extracted by thresholding $\mathbf{x}_p^*$ to obtain integer-valued labels \cite{Journee}.

 \begin{table*}[tb]
\parbox{0.40\linewidth}{
\centering
\small
   \caption{\small Results on image segmentation. Numbers are the mean over the images in Fig. \ref{fig:seg1}. Lower numbers are better. The proposed algorithm and the best performance are highlighted. }
   \begin{tabular}{|c||c|c|c|} 
     \hline
     Method & BNCut & SDCut & {\bf BCR}  \\ [0.5ex] 
     \hline\hline
     Time (s) & {\bf 0.08} & 27.64 & 0.97  \\ 
     \hline
     Objective & 10.84 & 6.40 & {\bf 6.34} \\ 
     \hline
     Rank & 1 & 7 & 2 \\ 
     \hline
   \end{tabular}
\label{tab:seg1}
   }
\hfill
\parbox{0.55\linewidth}{
\centering
\small
  \caption{\small Co-segmentation results. The proposed algorithm and the best performance is highlighted.}
  \begin{tabular}{cc|c|c|c|c|l}
    \cline{3-6}
    & & \multicolumn{4}{ c| }{Test Cases} \\ \cline{1-6}
    \multicolumn{2}{|c|}{Dataset} & horse & face & car-back & car-front \\ \cline{1-6}
    \multicolumn{2}{|c|}{Number of images} & 10 & 10 & 6 & 6 \\ \cline{1-6}
    \multicolumn{2}{|c|}{Variables in BQPs} & 4587& 6684 & 4012 & 4017 \\ \cline{1-6}
    \multicolumn{1}{ |c  }{\multirow{3}{*}{Time (s)} } &
    \multicolumn{1}{ |c| }{LowRank} & 2647 & 1614 & 724 & 749 &     \\ \cline{2-6}
    \multicolumn{1}{ |c  }{}                        &
    \multicolumn{1}{ |c| }{SDCut} & 220 & 274 & 180 & 590 &     \\ \cline{2-6}
    \multicolumn{1}{ |c  }{}                        &
    \multicolumn{1}{ |c| }{\bf BCR} & {\bf 18.8} & {\bf 61.8} & {\bf 46.7} & {\bf 44.7} &     \\ \cline{1-6}
    \multicolumn{1}{ |c  }{\multirow{3}{*}{Objective} } &
    \multicolumn{1}{ |c| }{LowRank} & 4.84 & 4.48 & 5.00 & 4.17 &     \\ \cline{2-6}
    \multicolumn{1}{ |c  }{}                        &
    \multicolumn{1}{ |c| }{SDCut} & 5.24 & 4.94 & 4.53 & 4.27 &     \\ \cline{2-6}
    \multicolumn{1}{ |c  }{}                        &
    \multicolumn{1}{ |c| }{\bf BCR} & {\bf 4.64} & {\bf 3.29} & {\bf 4.36} & {\bf 3.94} &     \\ \cline{1-6}
    \multicolumn{1}{ |c  }{\multirow{3}{*}{Rank} } &
    \multicolumn{1}{ |c| }{LowRank} & 18 & 11 & 7 & 10 &     \\ \cline{2-6}
    \multicolumn{1}{ |c  }{}                        &
    \multicolumn{1}{ |c| }{SDCut} & 3 & 3 & 3 & 3 &     \\ \cline{2-6}
    \multicolumn{1}{ |c  }{}                        &
    \multicolumn{1}{ |c| }{\bf BCR} & {\bf  2} & {\bf  2} & {\bf  2} & {\bf  2} &     \\ \cline{1-6}
  \end{tabular}
  \label{table:coseg1}
    }
\end{table*}

{\bf Experiments:}
We compare BCR to two well-known co-segmentation methods, namely low-rank factorization~\cite{Journee} (denoted LR) and SDCut~\cite{Wang_2013}. We use publicly available code for LR and SDCut. We test on the Weizman horses\footnote{www.msri.org/people/members/eranb/} and MSRC\footnote{\mbox{www.research.microsoft.com/en-us/projects/objectclassrecognition/}} datasets with a total of four classes (horse, car-front, car-back, and face) containing $6\sim10$ images per class. Each image is over-segmented to $400\sim700$ SLIC superpixels using the VLFeat~\cite{vedaldi2012vlfeat} toolbox, giving a total of around $4000$ $\sim$ $7000$ super-pixels per class. Relative to image segmentation problems, this application requires $10\times$ more variables.
%

\begin{figure*}[!tp]
  \centering
  \includegraphics[width=0.115\linewidth]{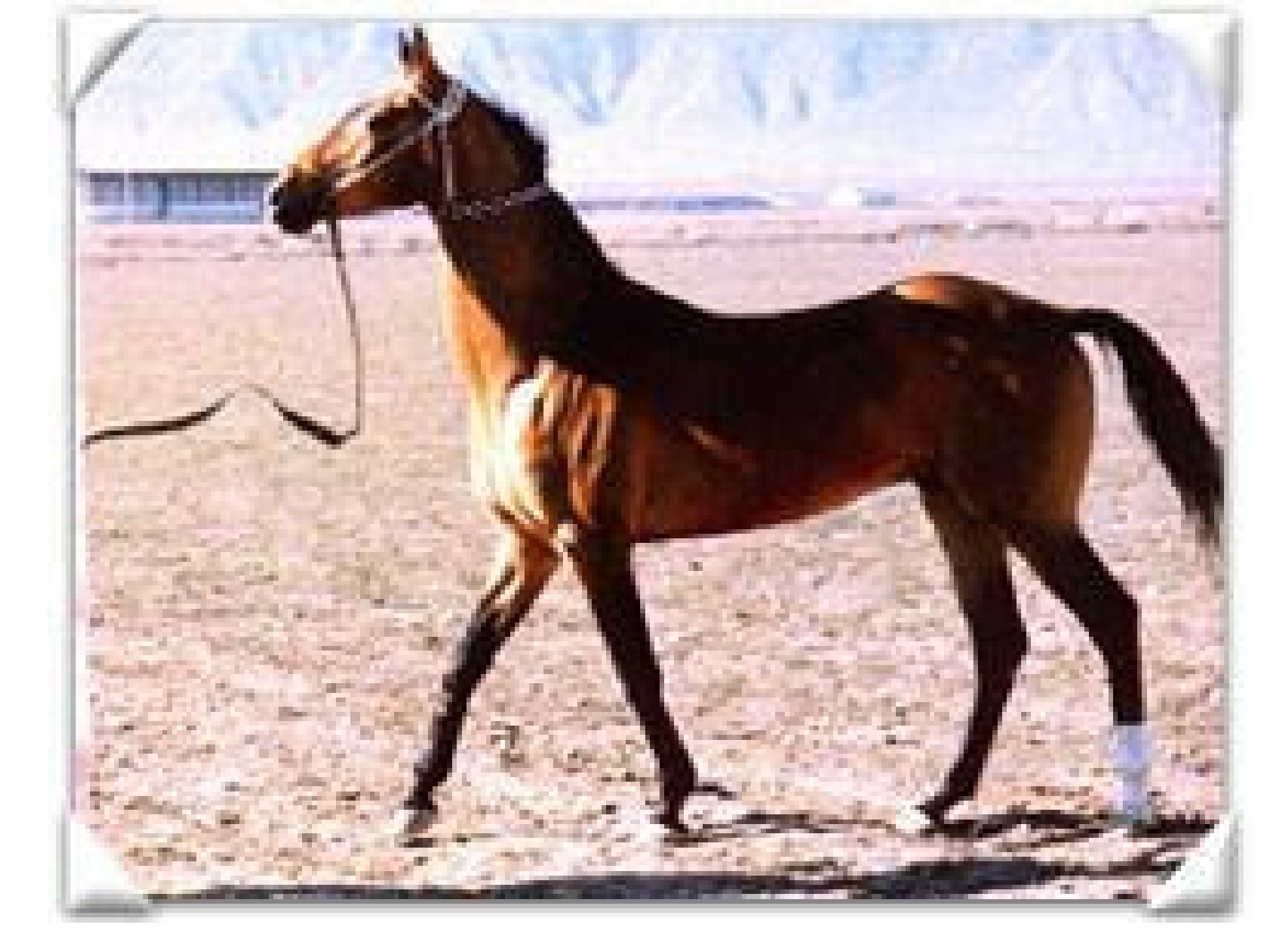}
  \includegraphics[width=0.115\linewidth]{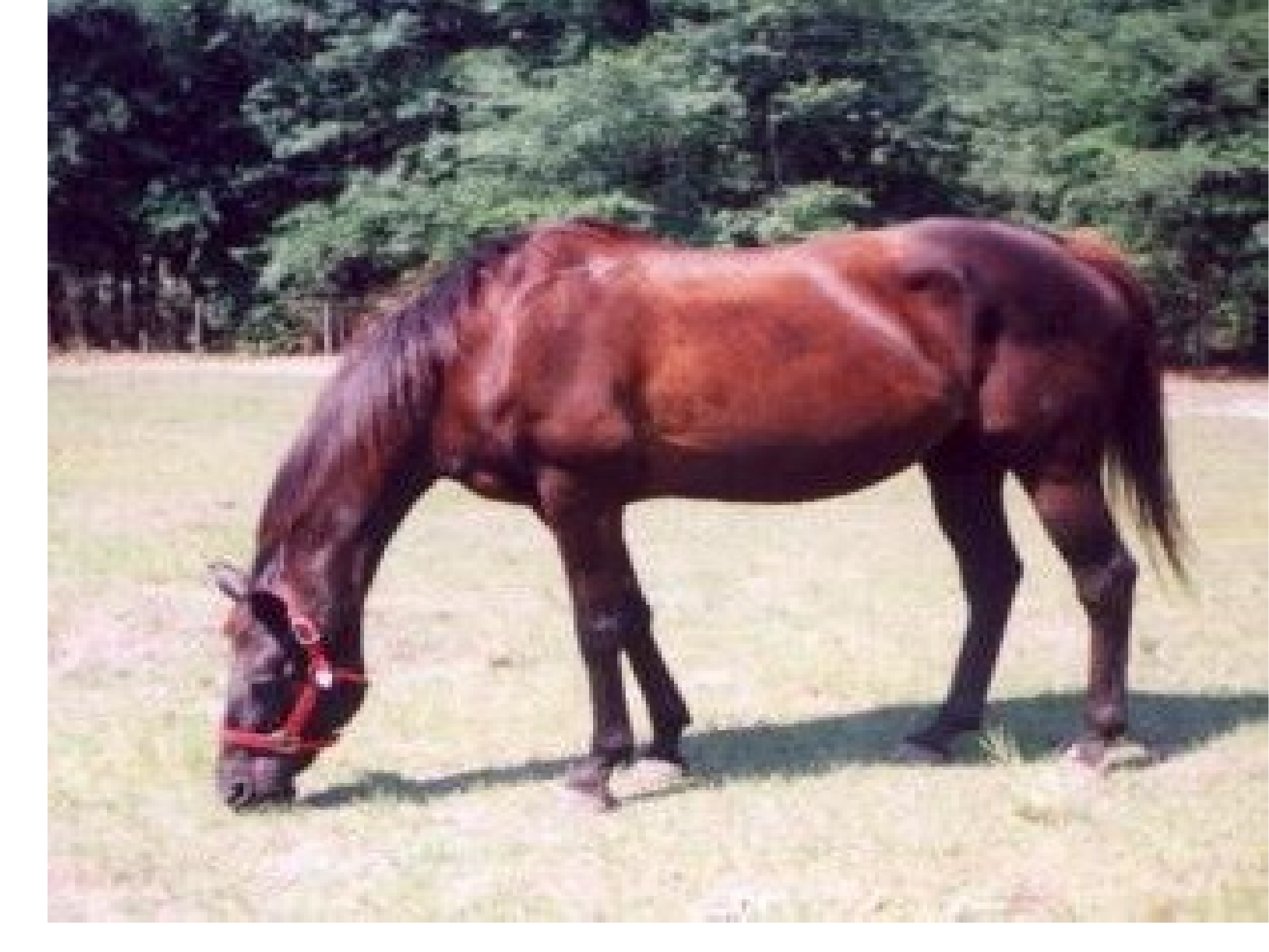}
  \includegraphics[width=0.115\linewidth]{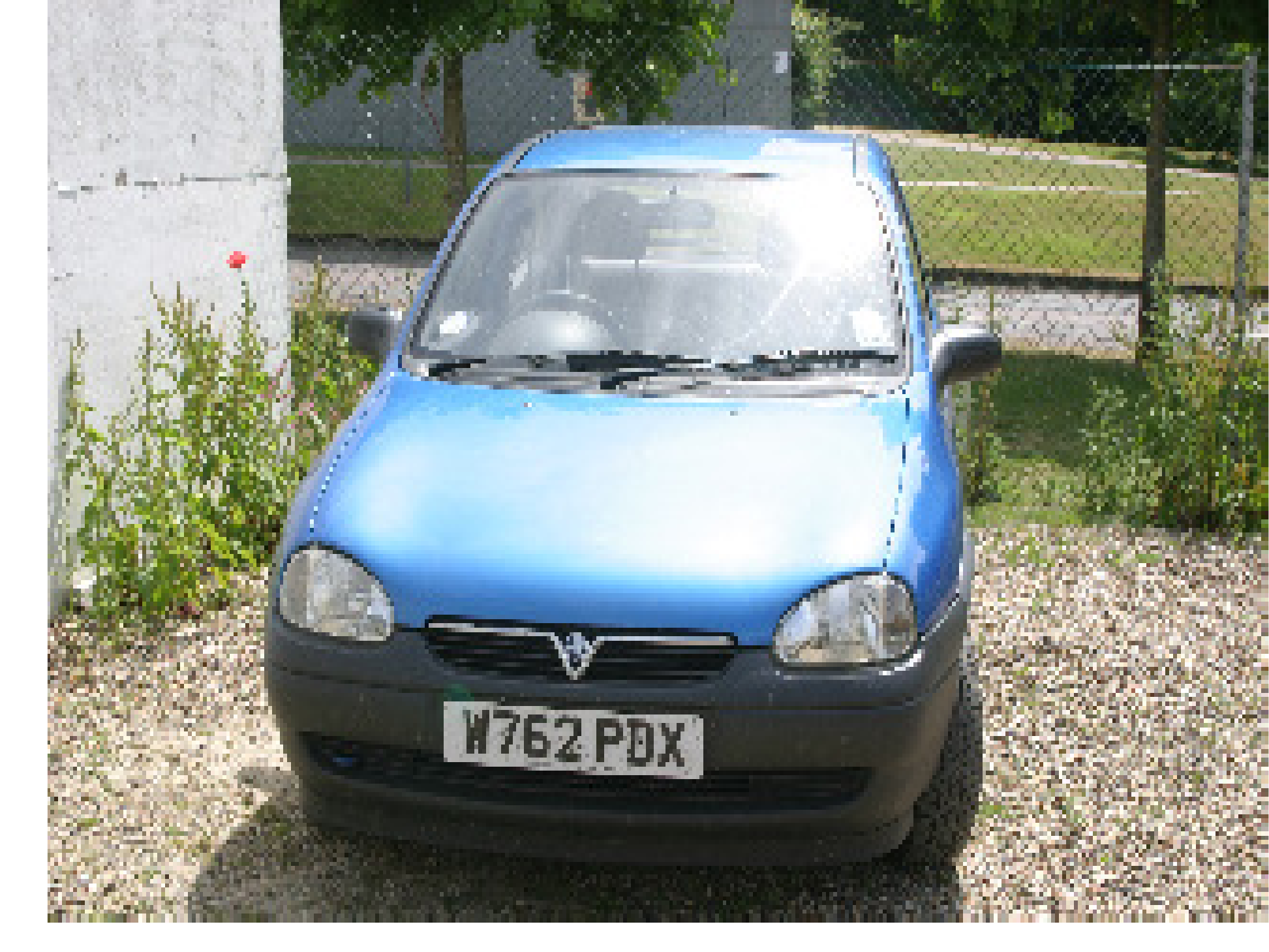}
  \includegraphics[width=0.115\linewidth]{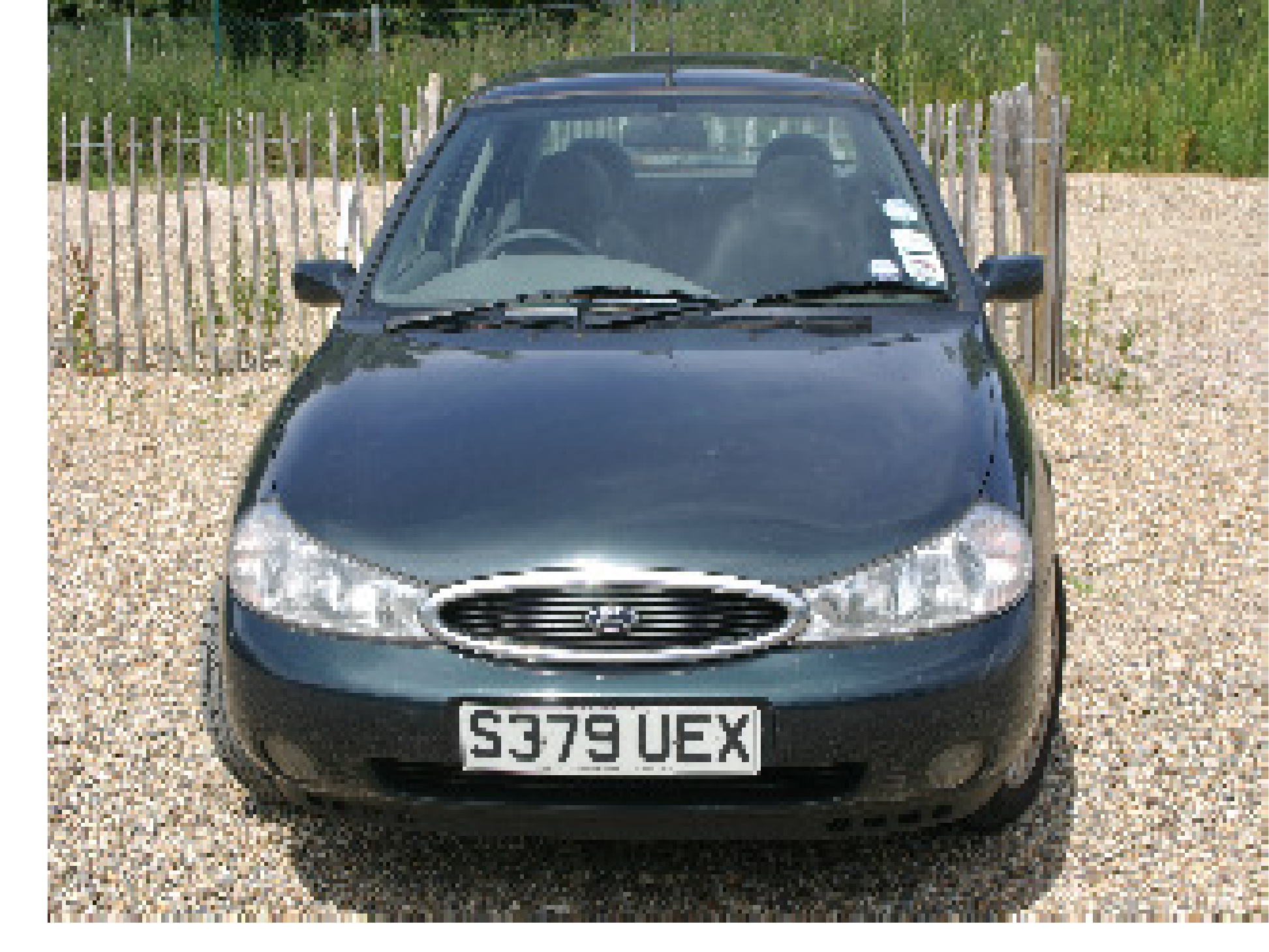}
  \includegraphics[width=0.115\linewidth]{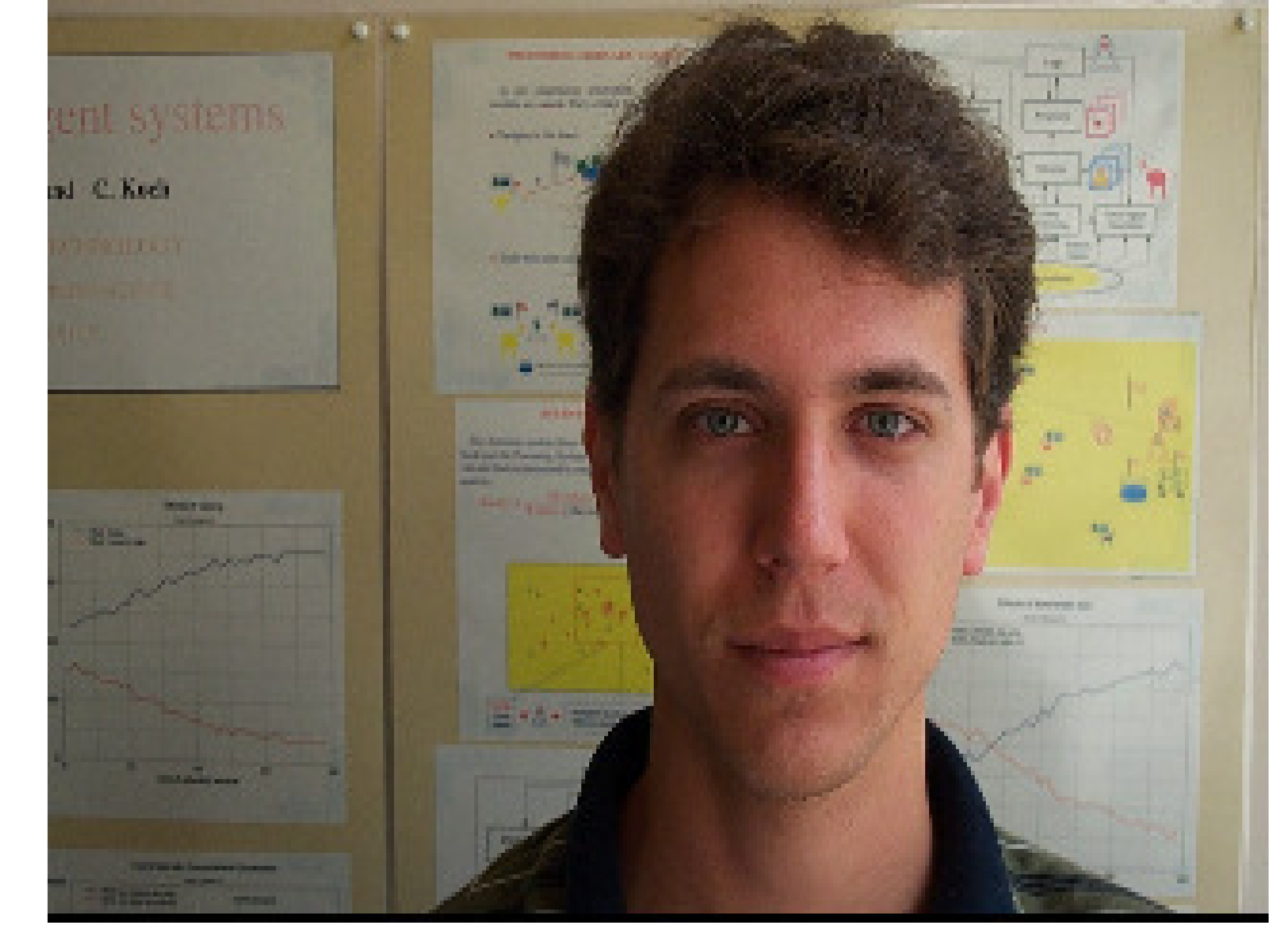}
  \includegraphics[width=0.115\linewidth]{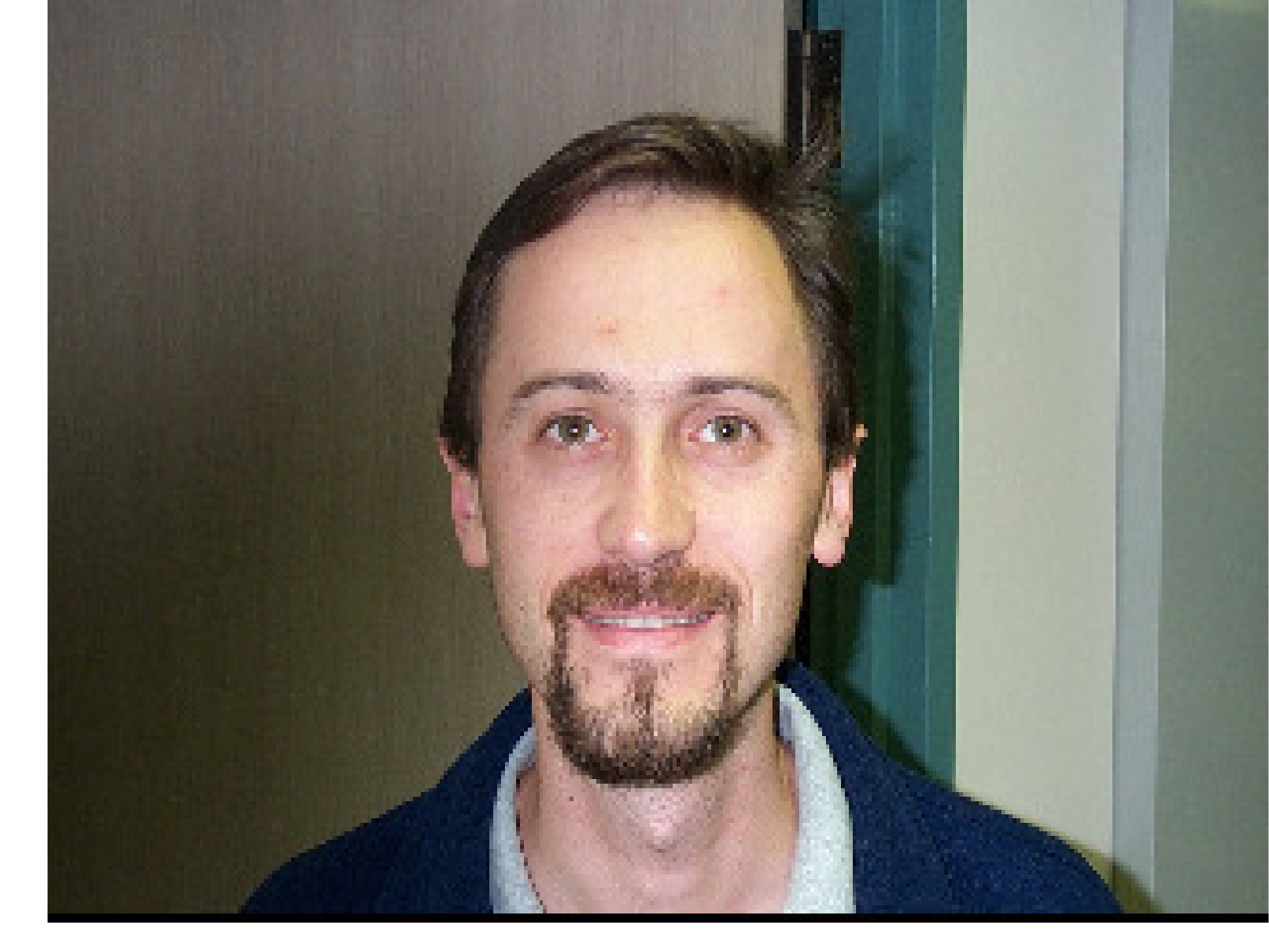}
  \includegraphics[width=0.115\linewidth]{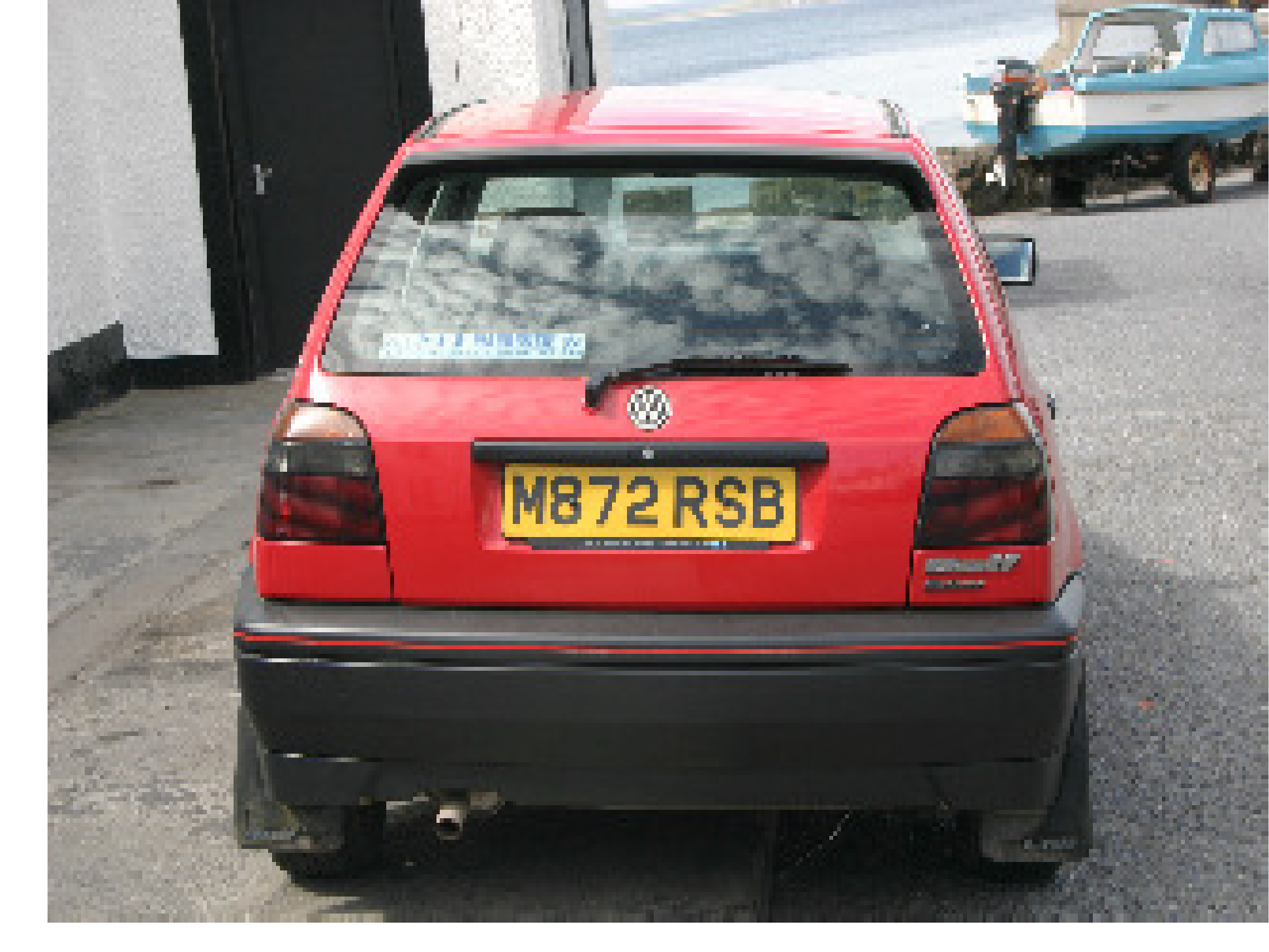}
  \includegraphics[width=0.115\linewidth]{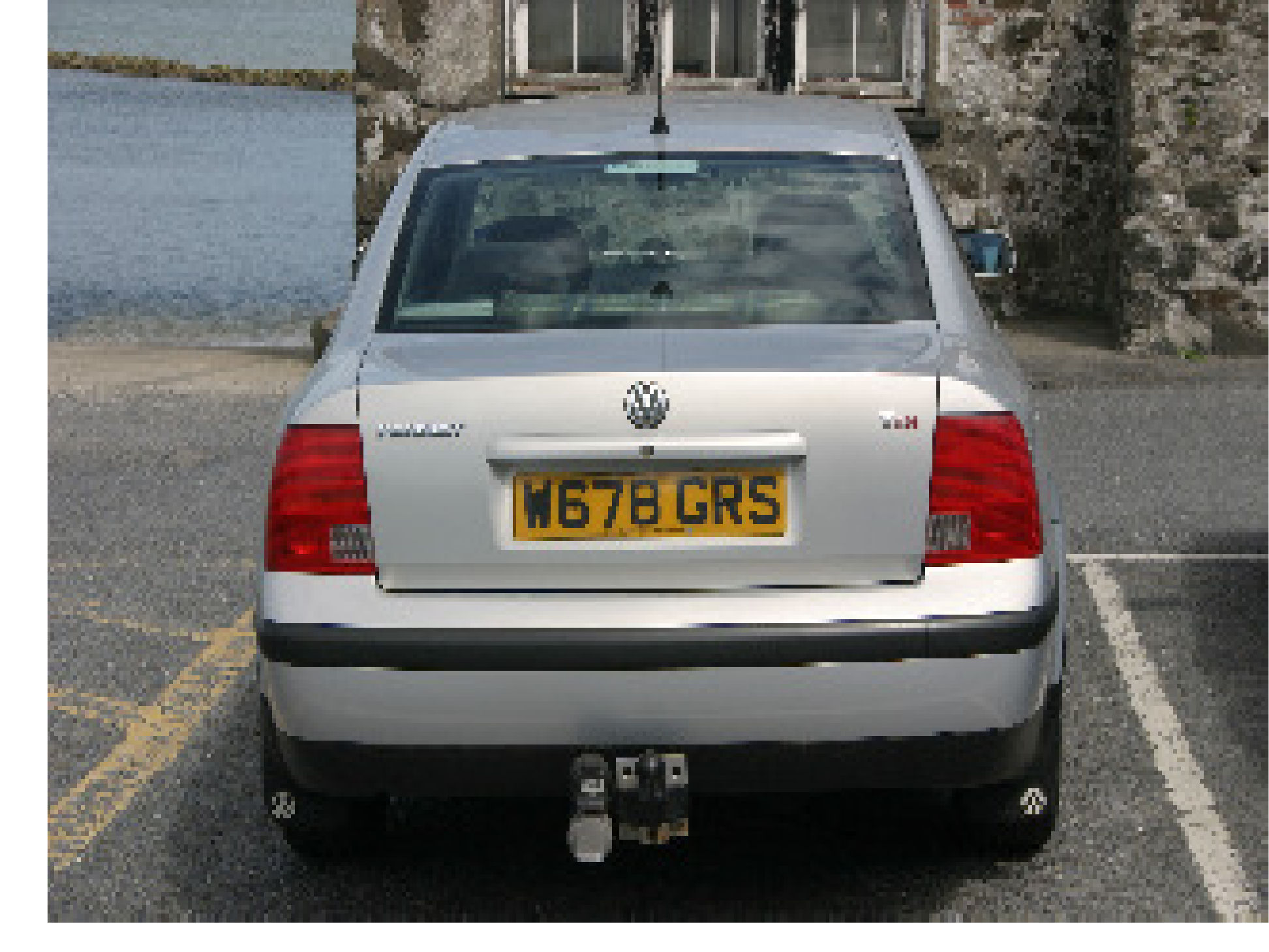}

  \includegraphics[width=0.115\linewidth]{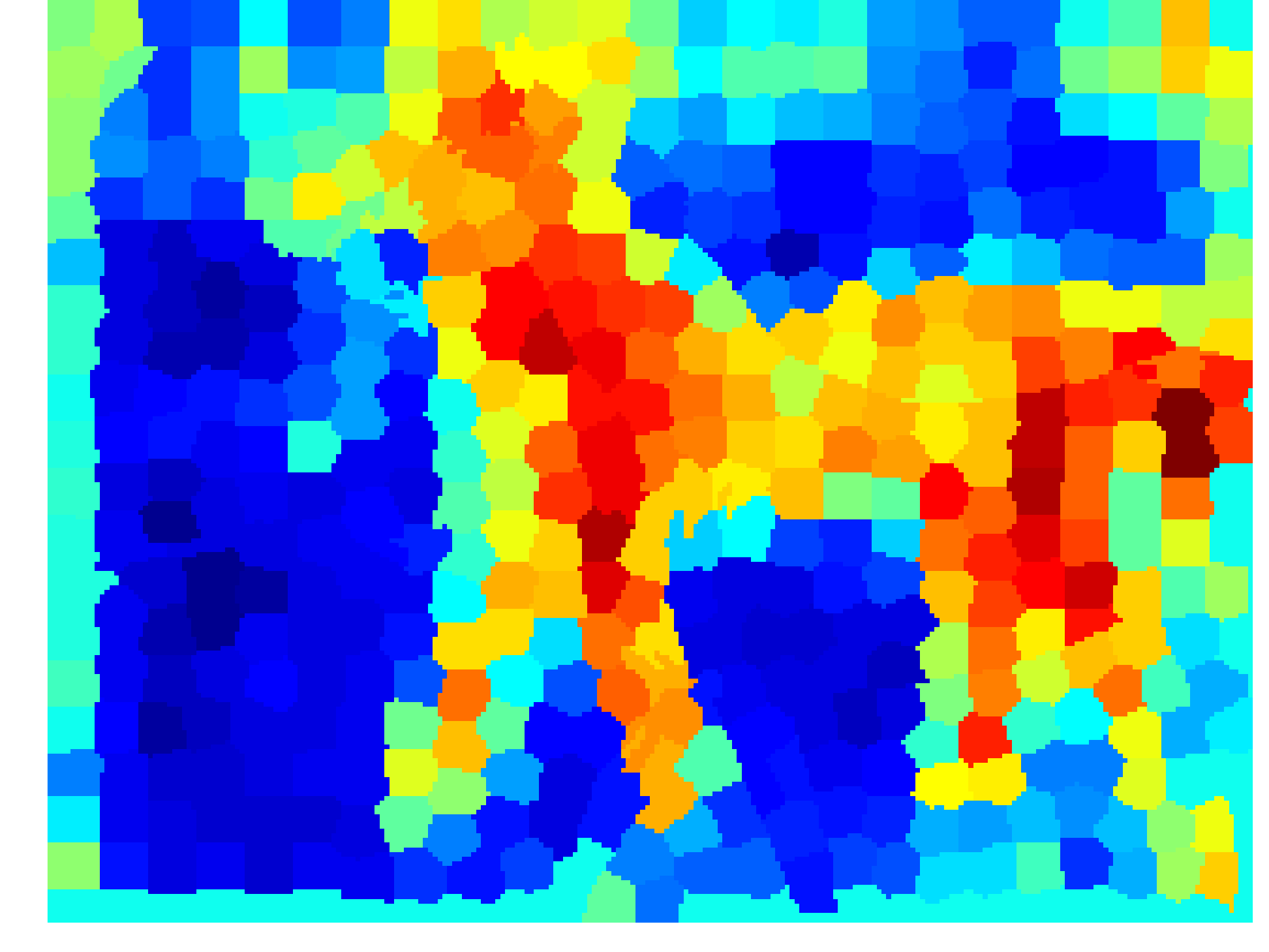}
  \includegraphics[width=0.115\linewidth]{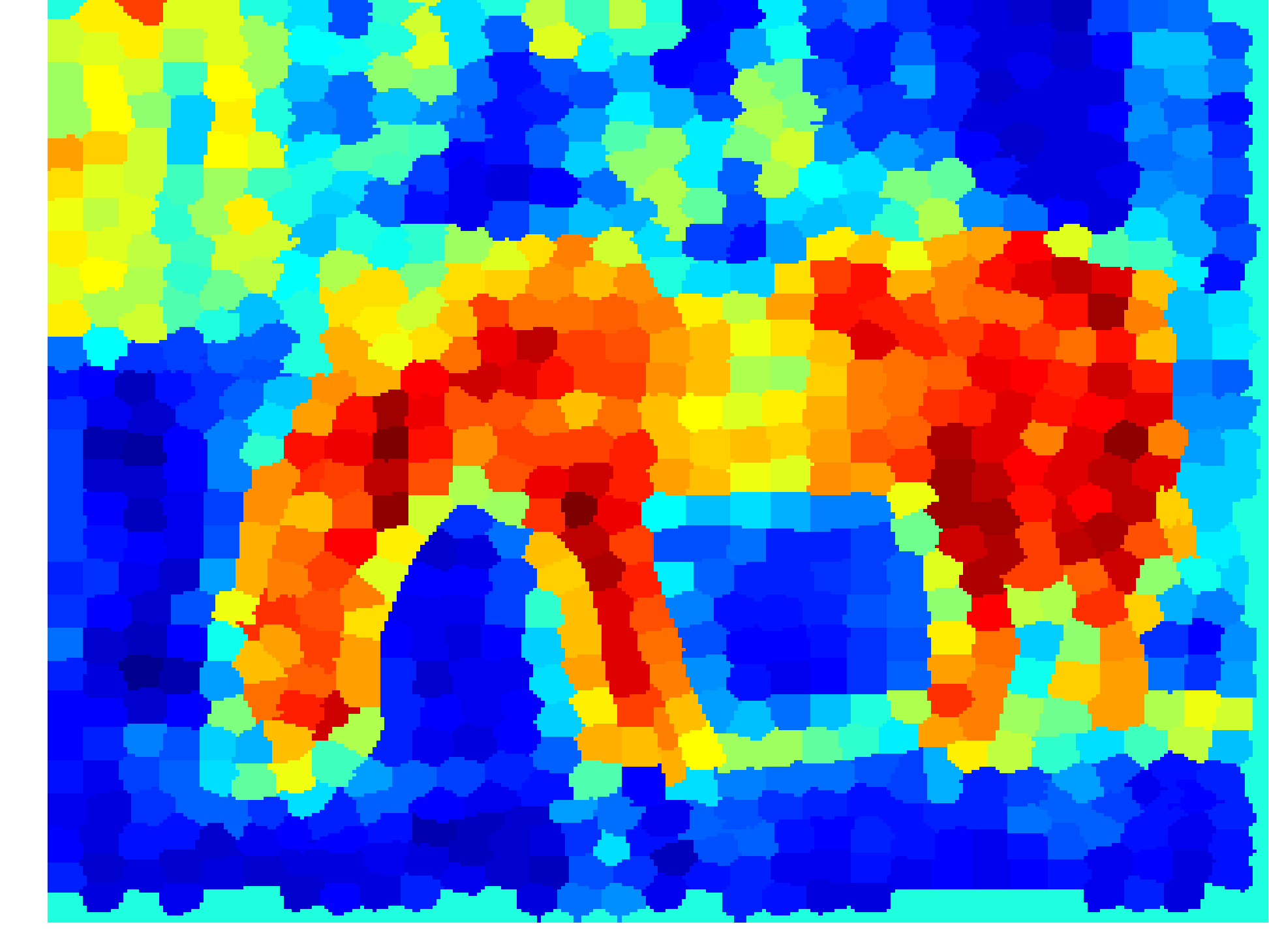}
  \includegraphics[width=0.115\linewidth]{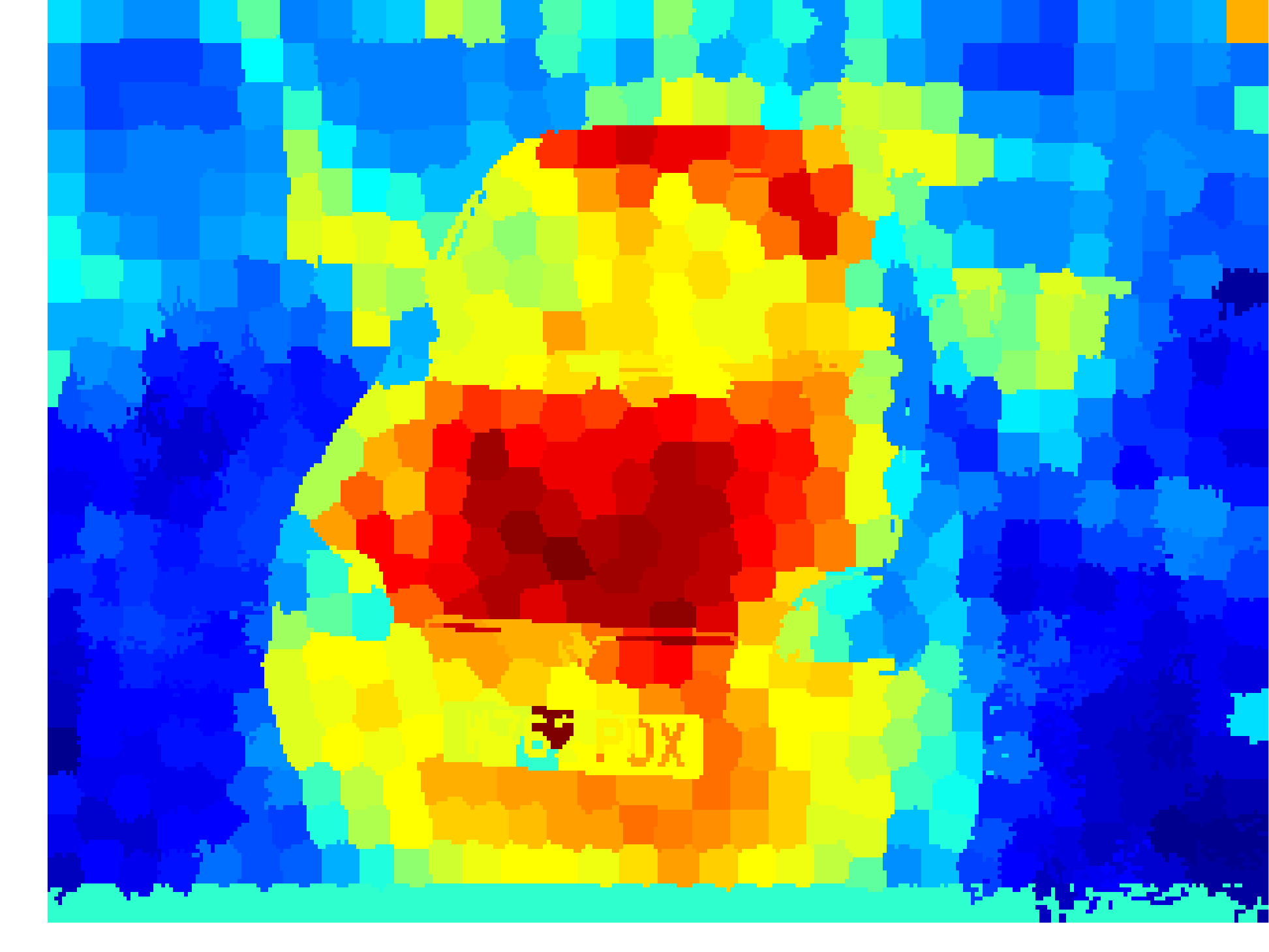}
  \includegraphics[width=0.115\linewidth]{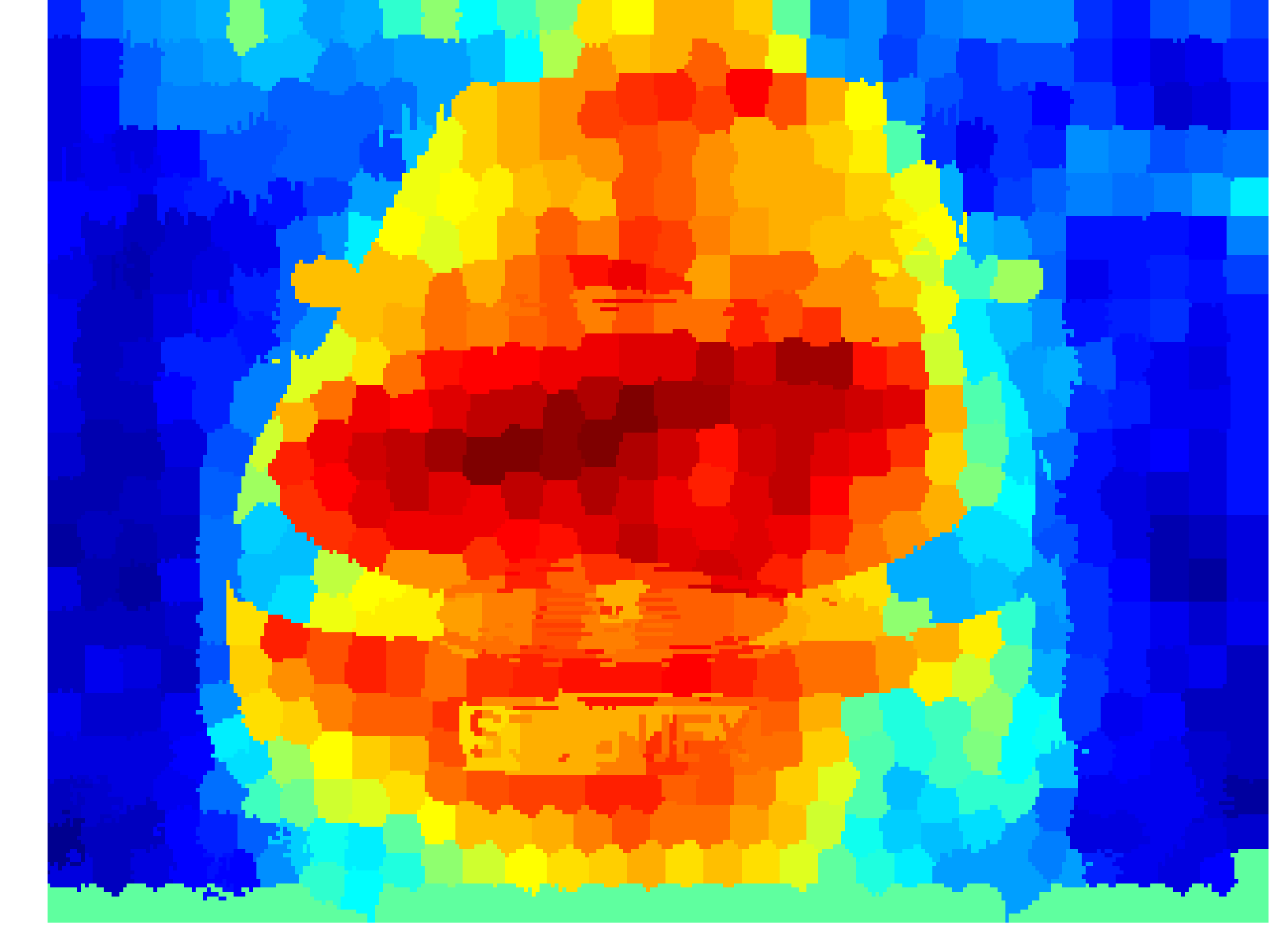}
  \includegraphics[width=0.115\linewidth]{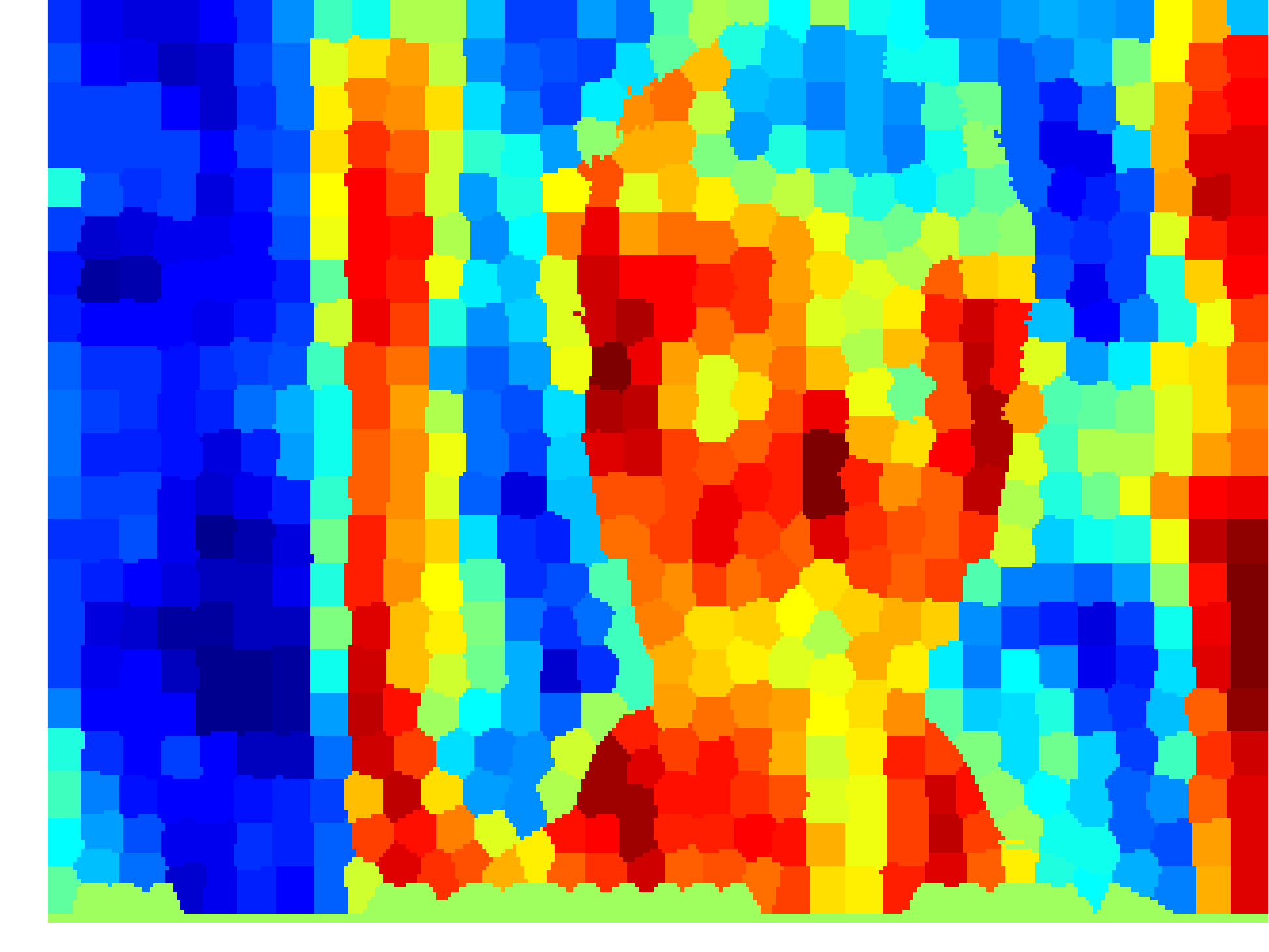}
  \includegraphics[width=0.115\linewidth]{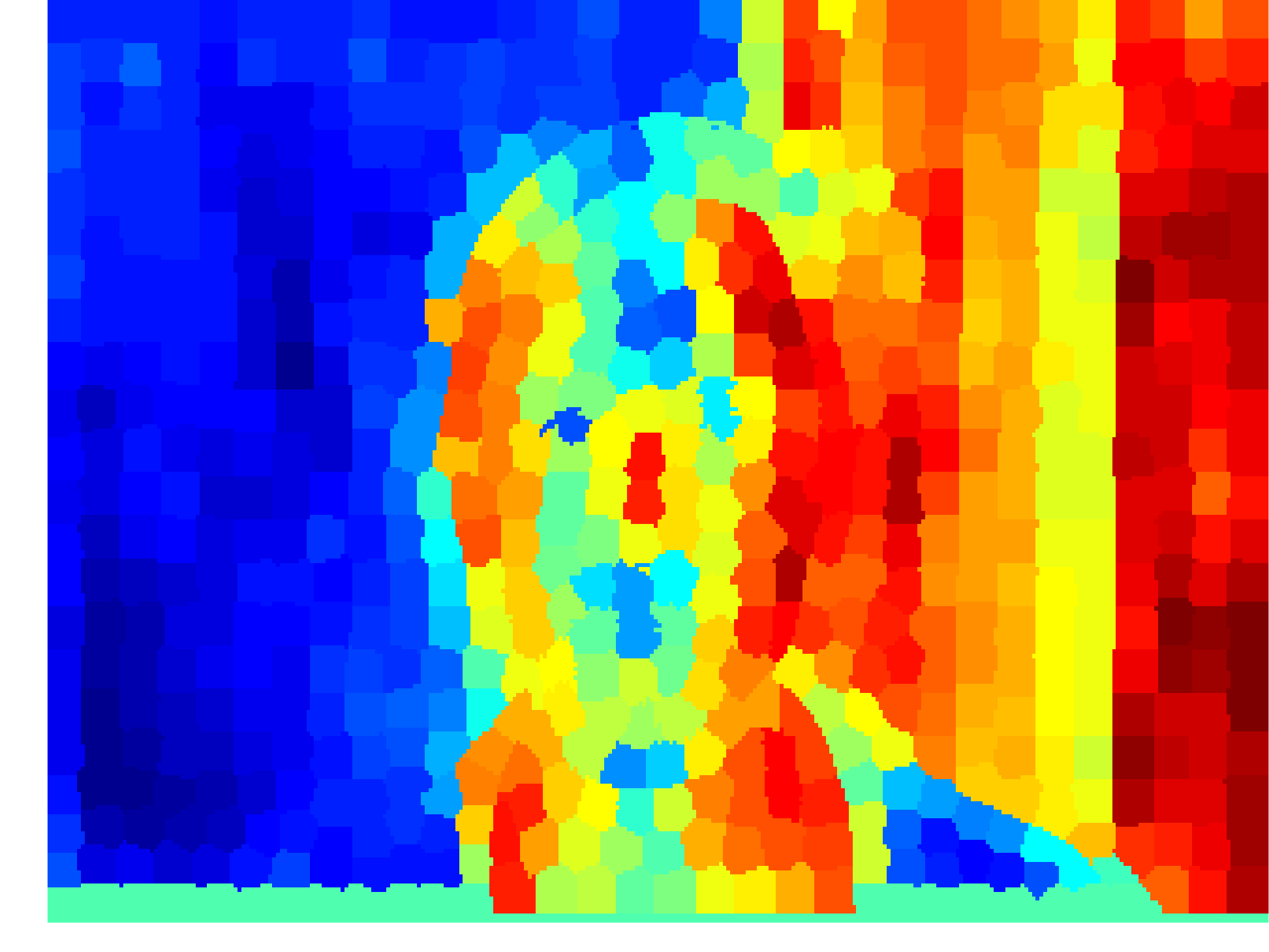}
  \includegraphics[width=0.115\linewidth]{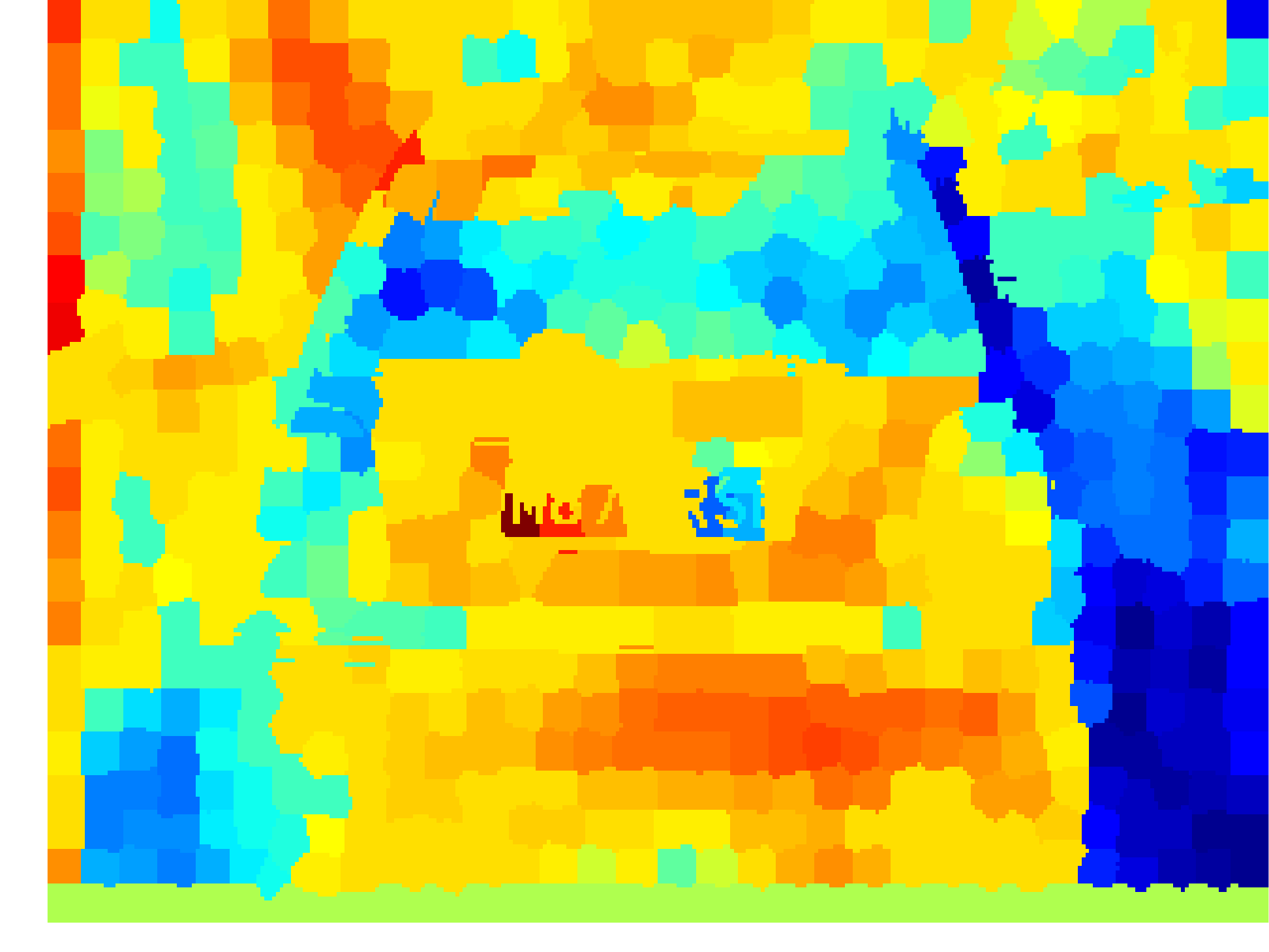}
  \includegraphics[width=0.115\linewidth]{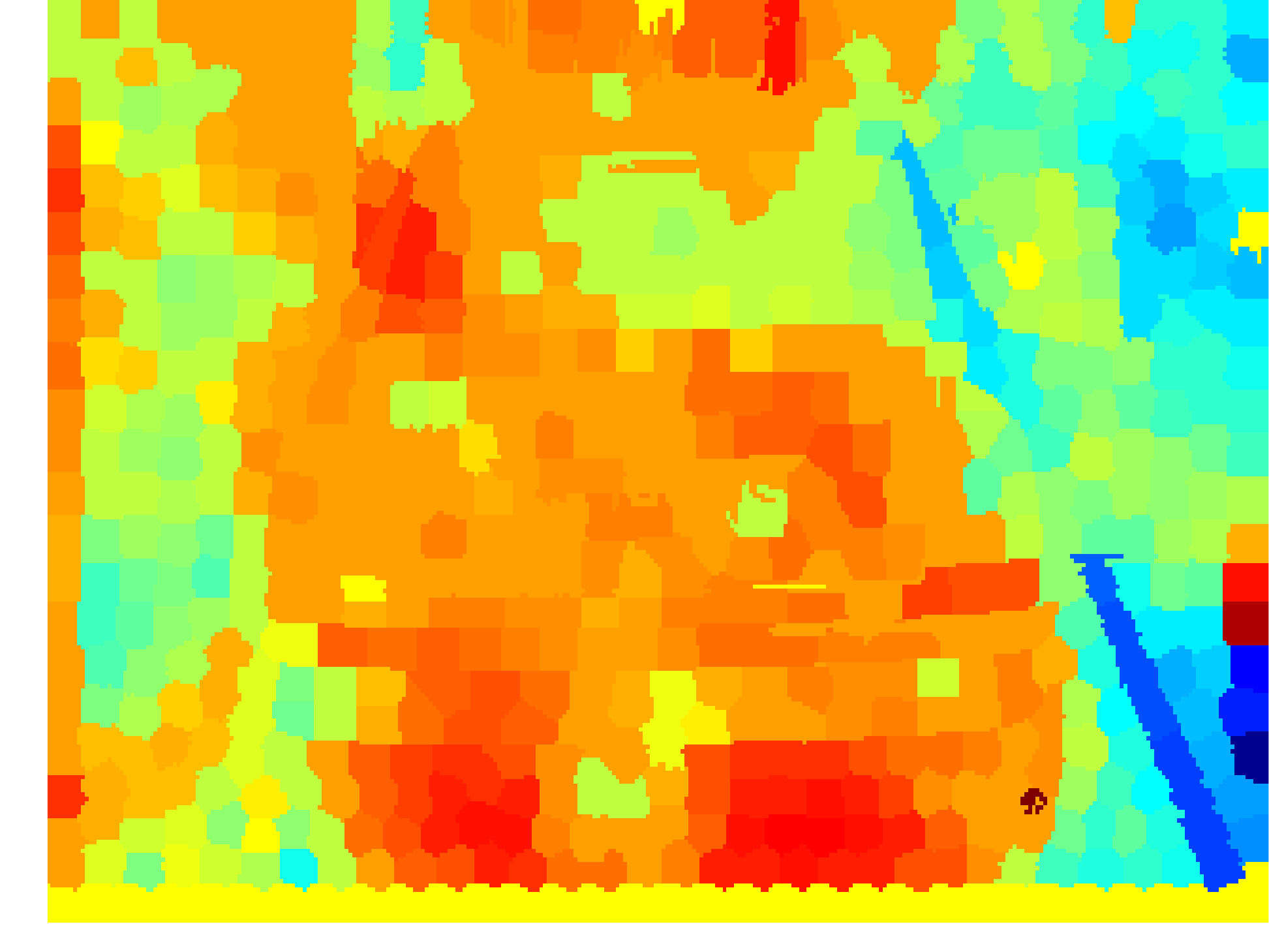}

  \includegraphics[width=0.115\linewidth]{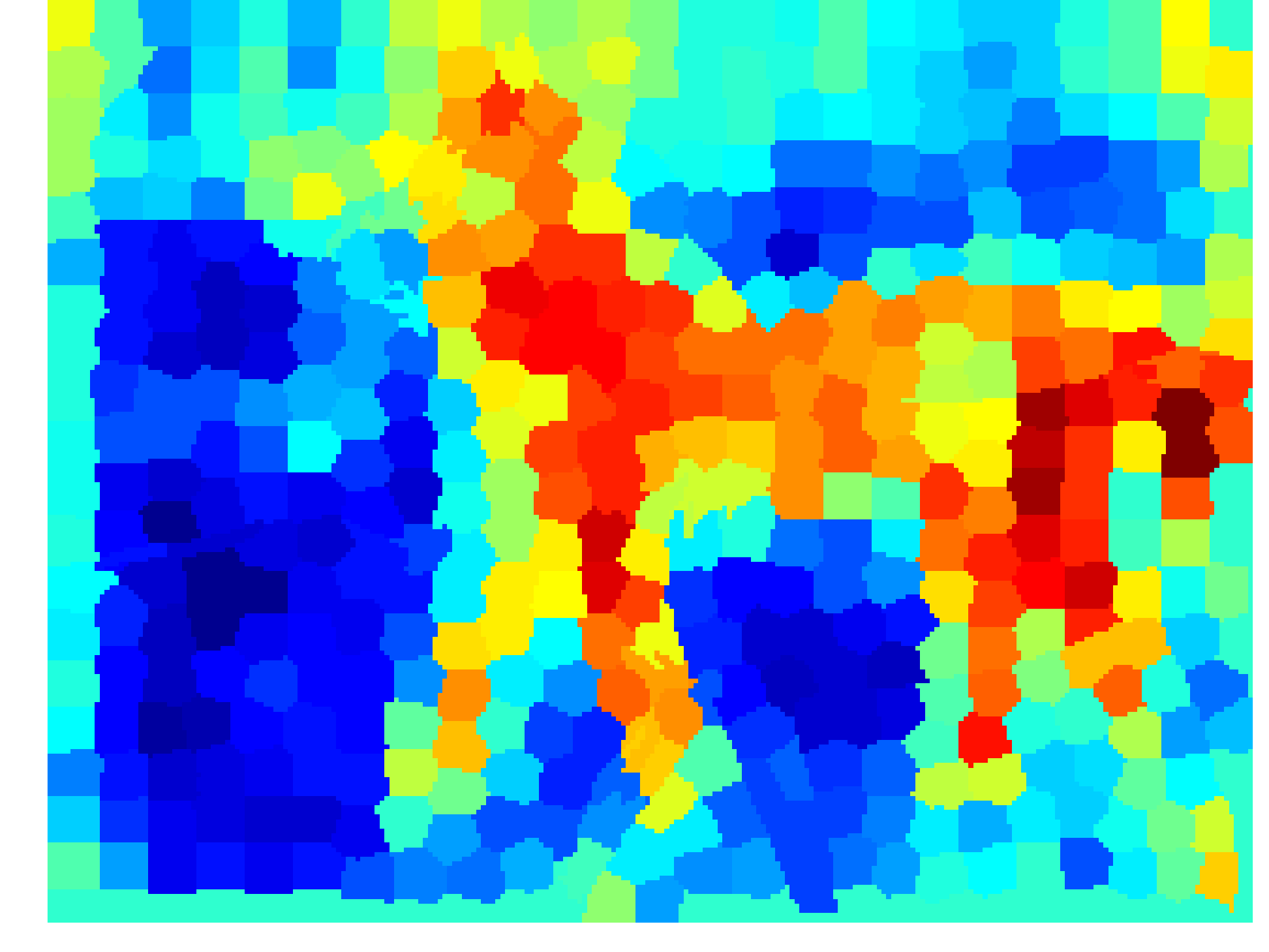}
  \includegraphics[width=0.115\linewidth]{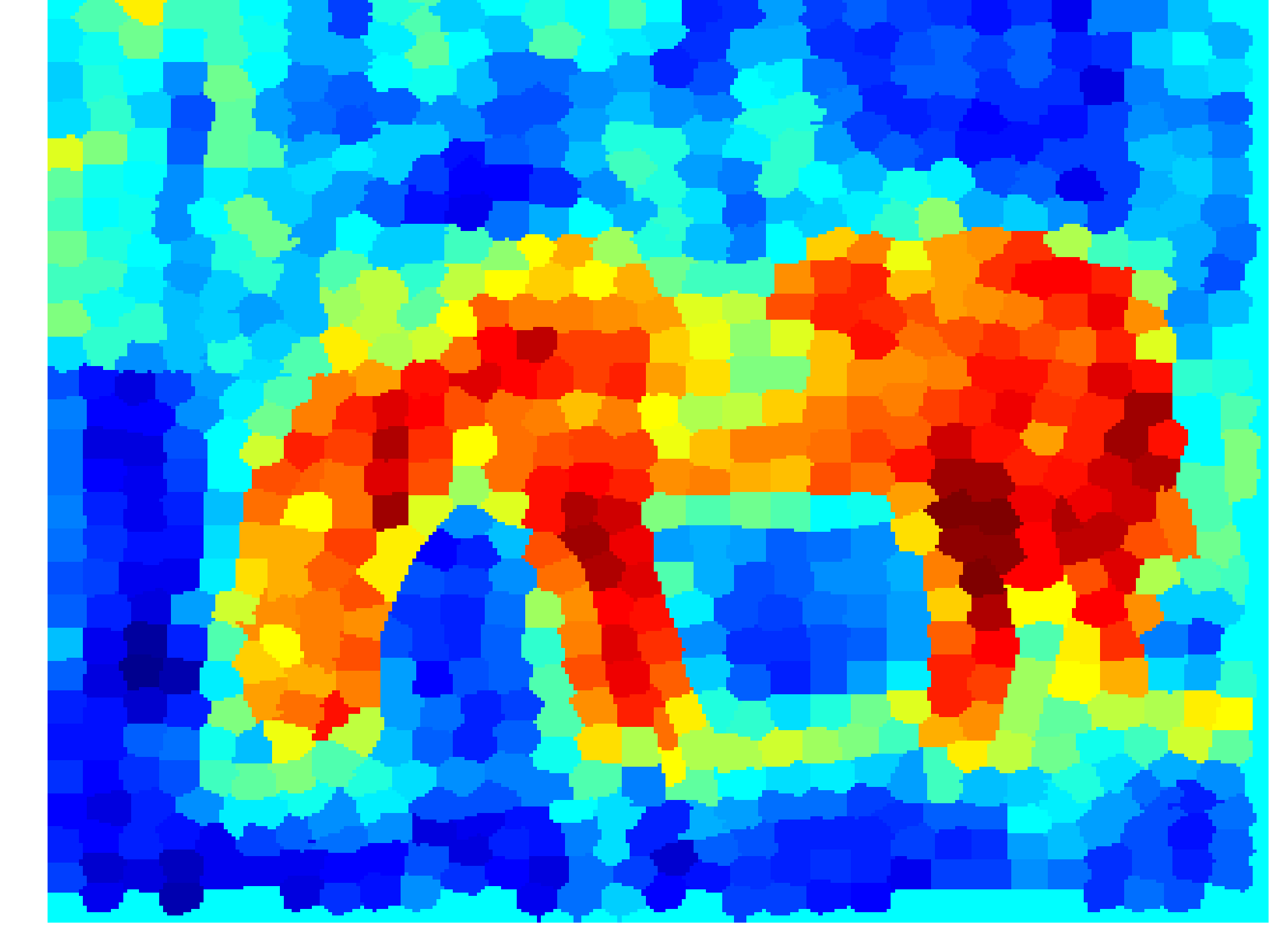}
  \includegraphics[width=0.115\linewidth]{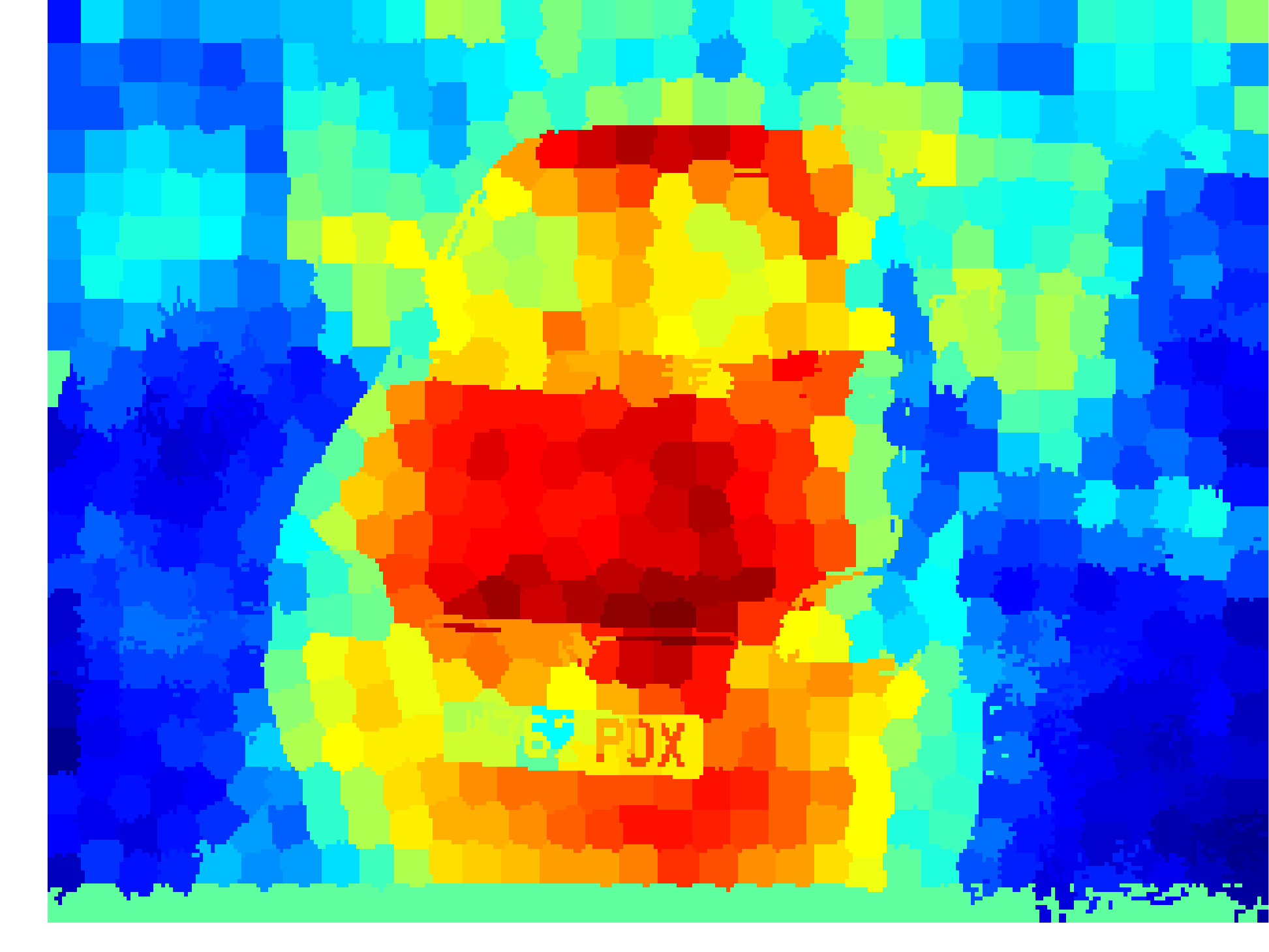}
  \includegraphics[width=0.115\linewidth]{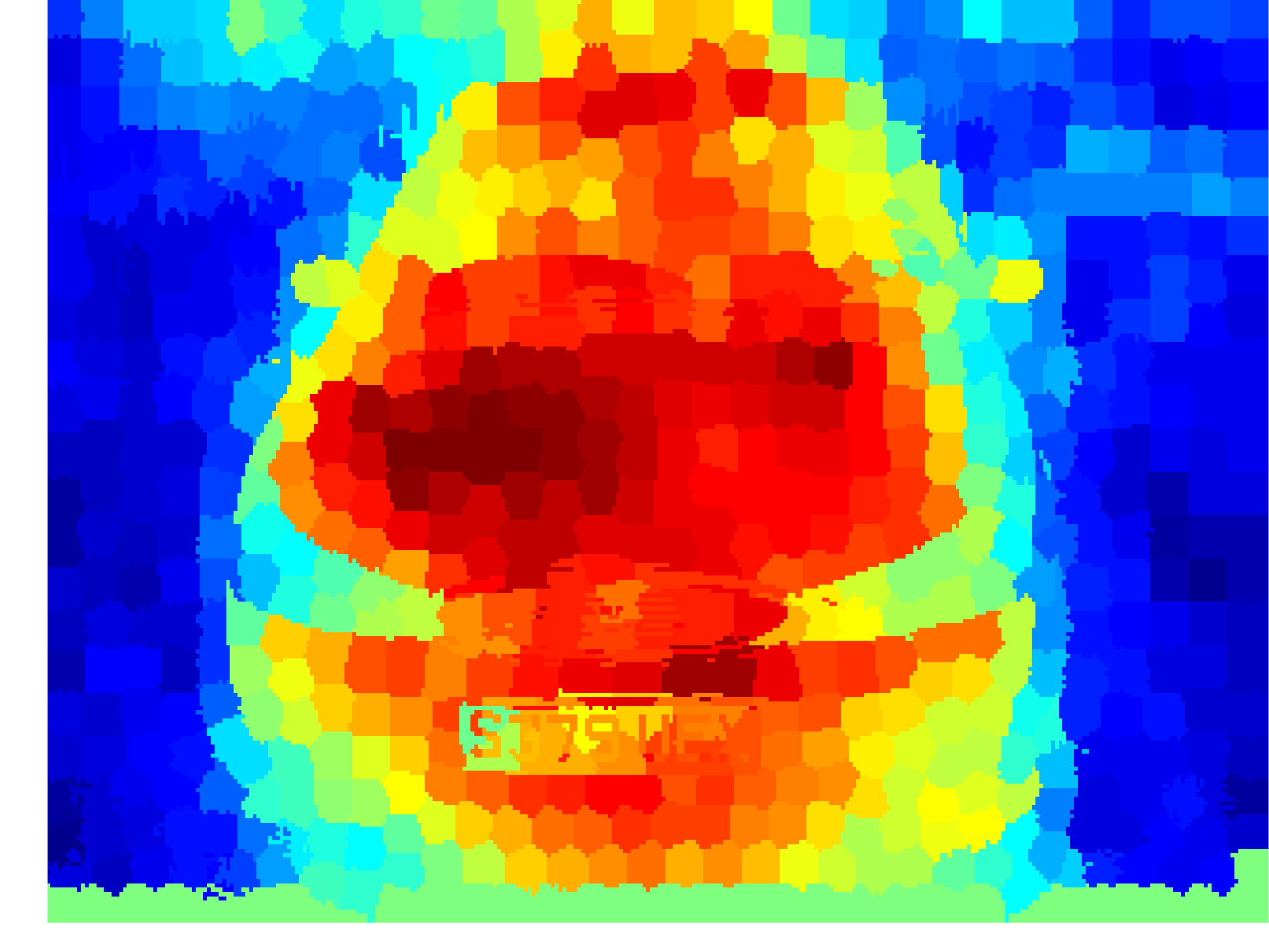}
  \includegraphics[width=0.115\linewidth]{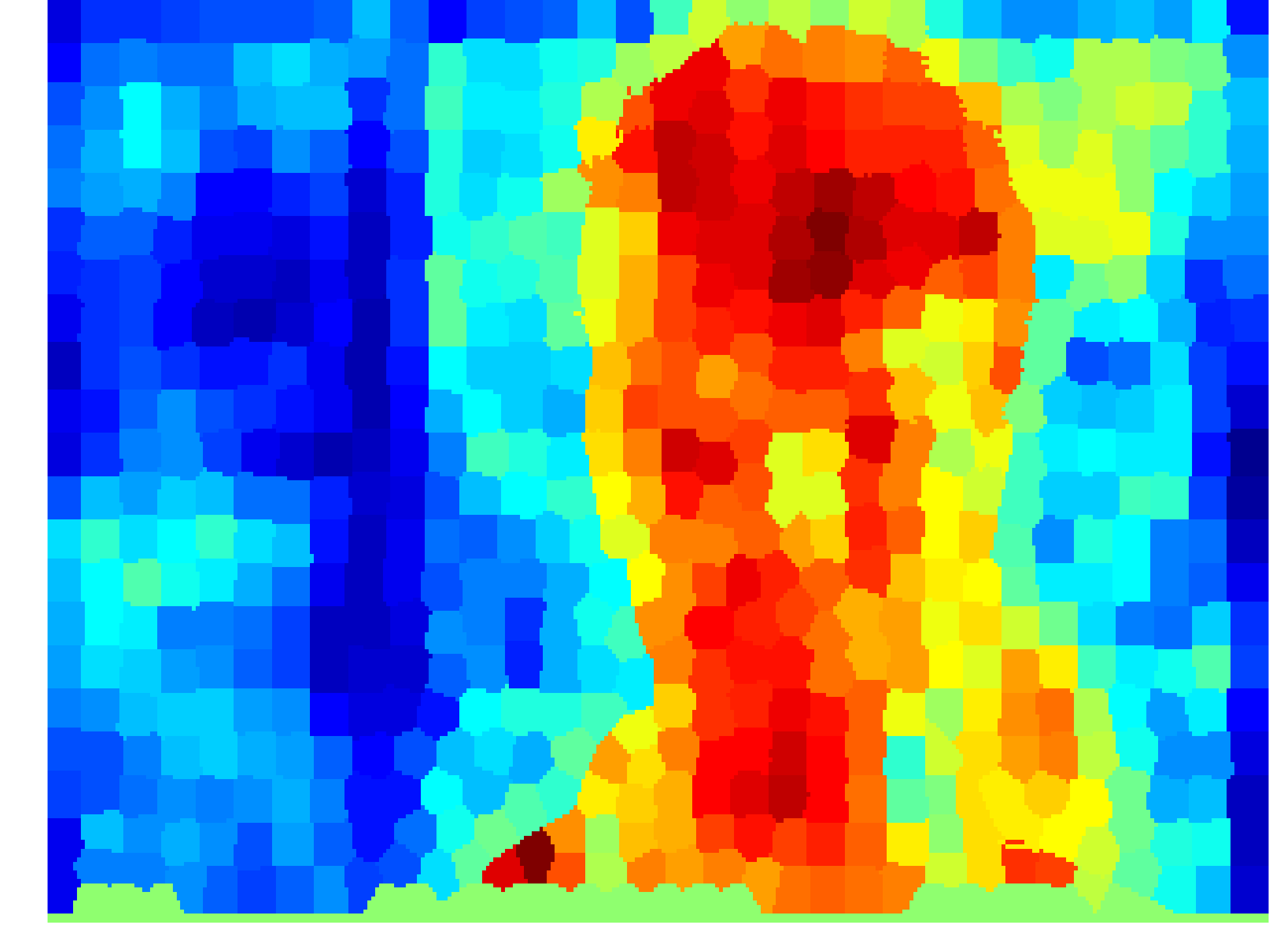}
  \includegraphics[width=0.115\linewidth]{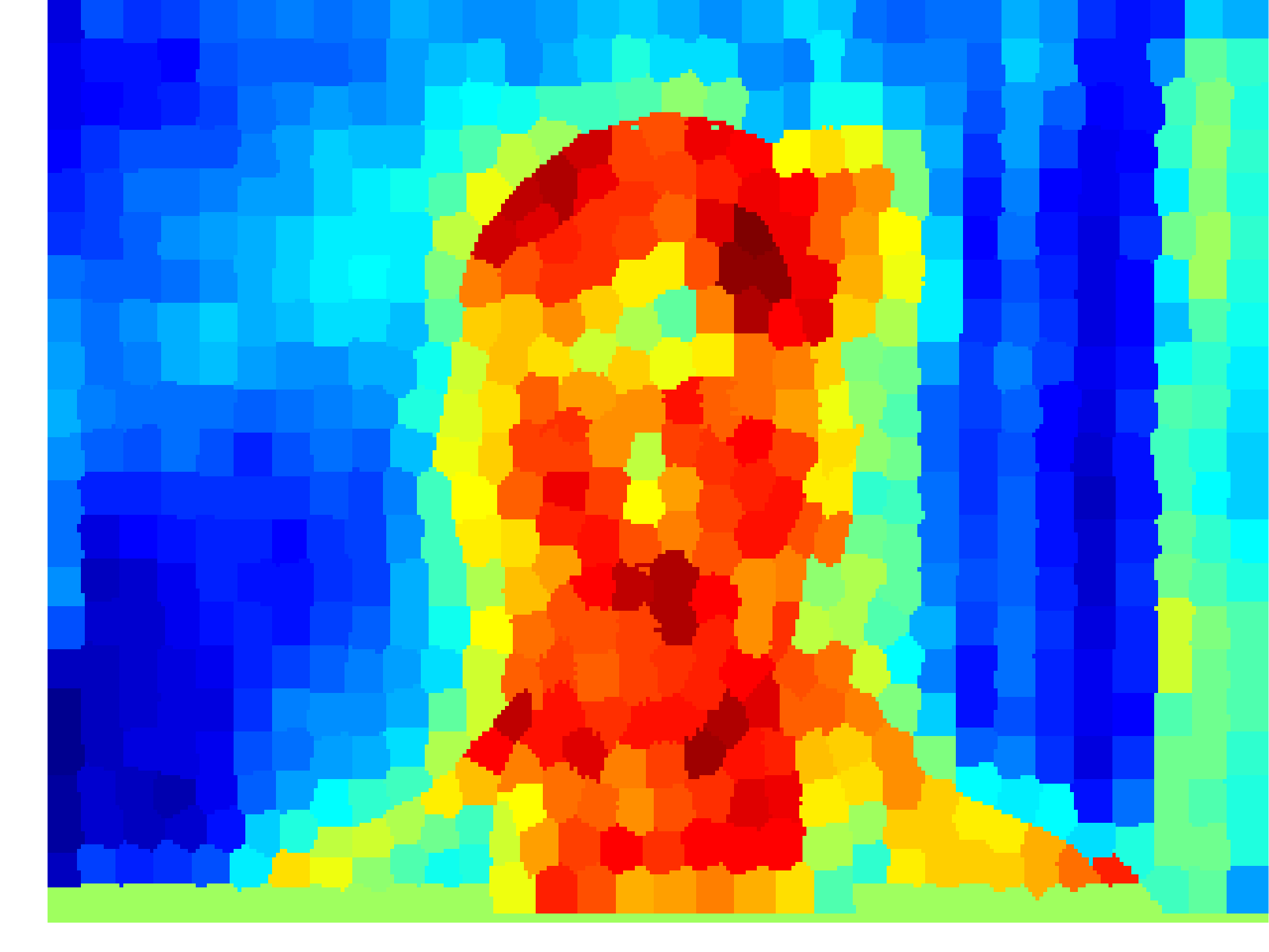}
  \includegraphics[width=0.115\linewidth]{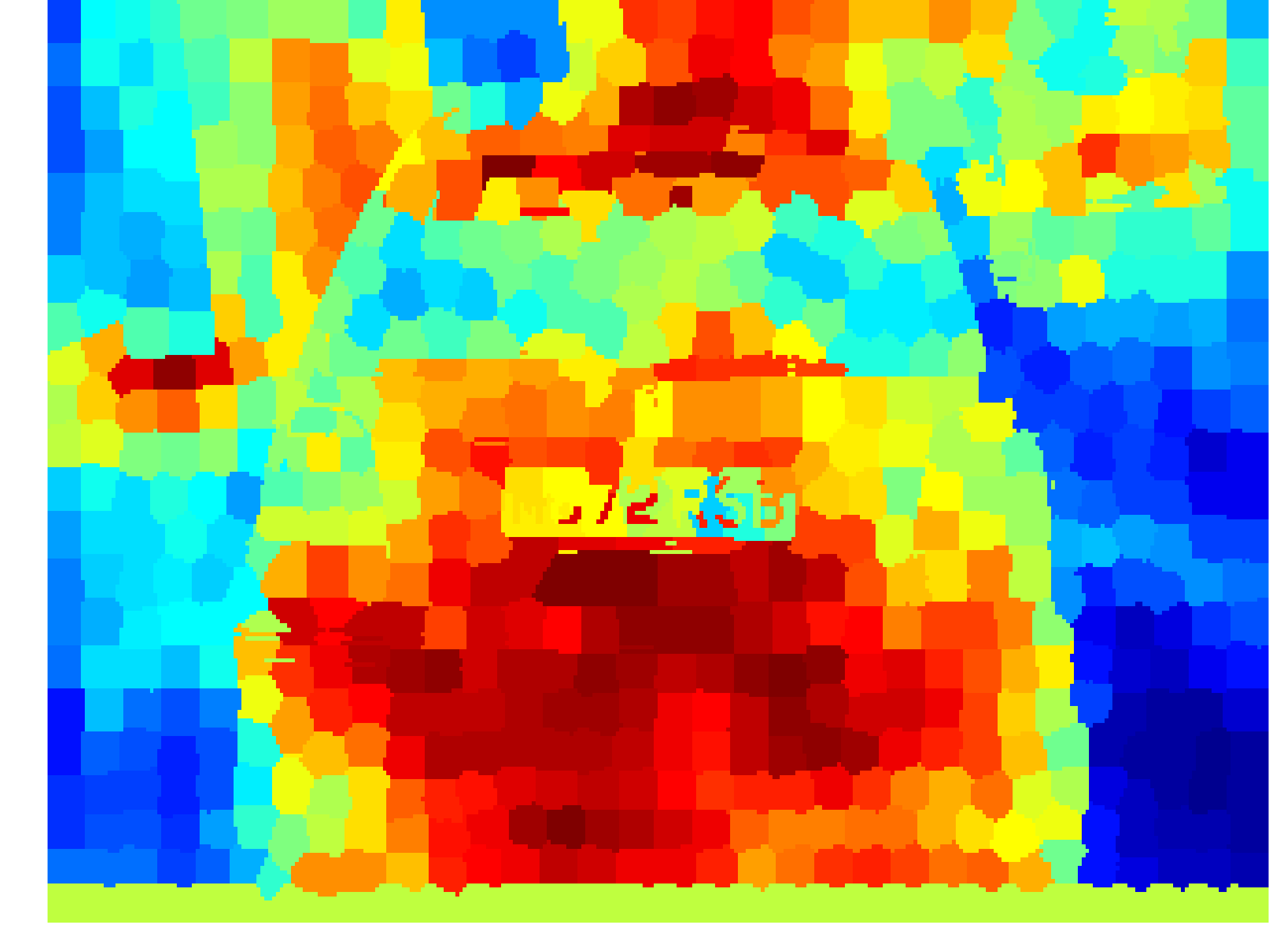}
  \includegraphics[width=0.115\linewidth]{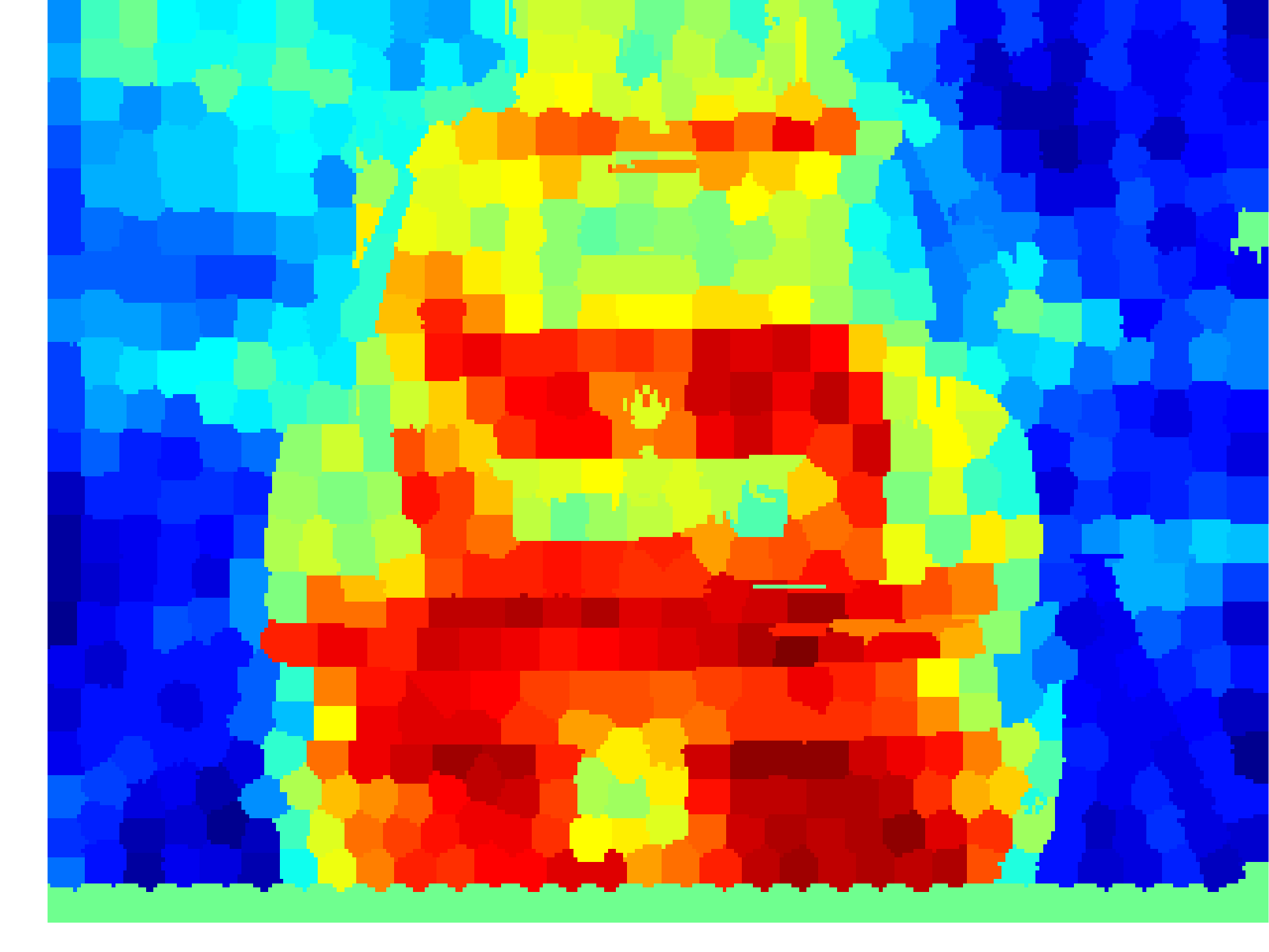}

  \includegraphics[width=0.115\linewidth]{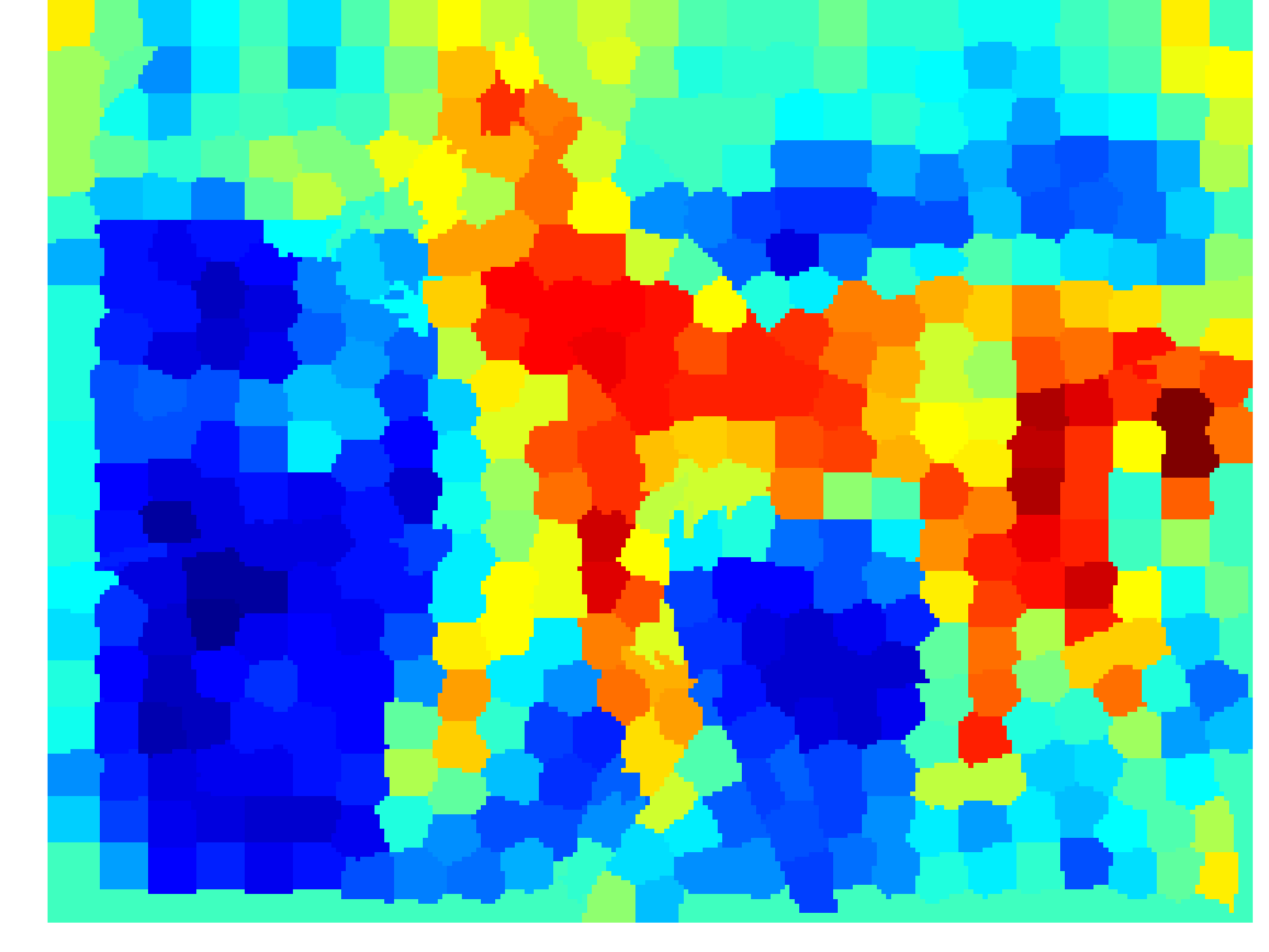}
  \includegraphics[width=0.115\linewidth]{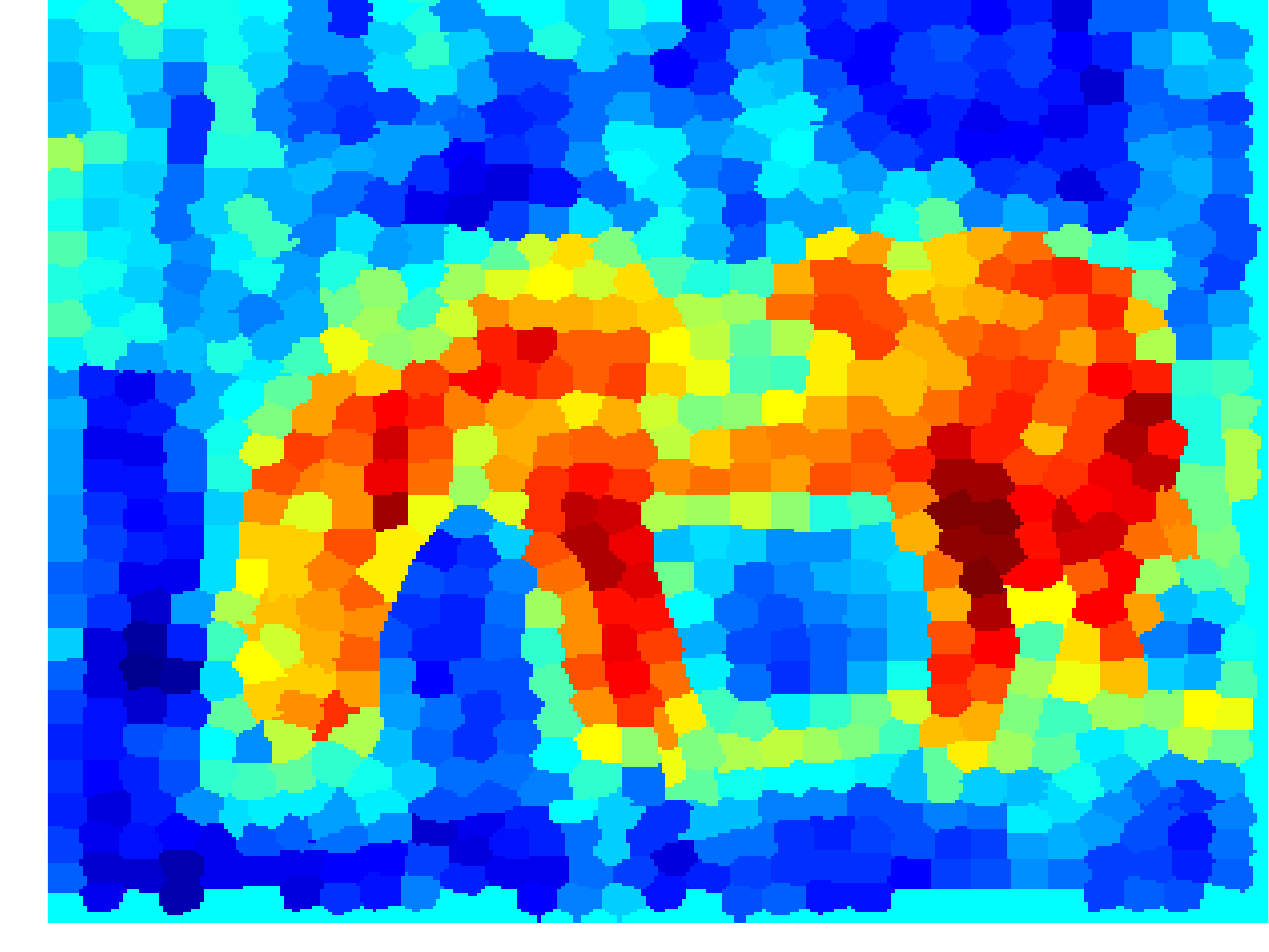}
  \includegraphics[width=0.115\linewidth]{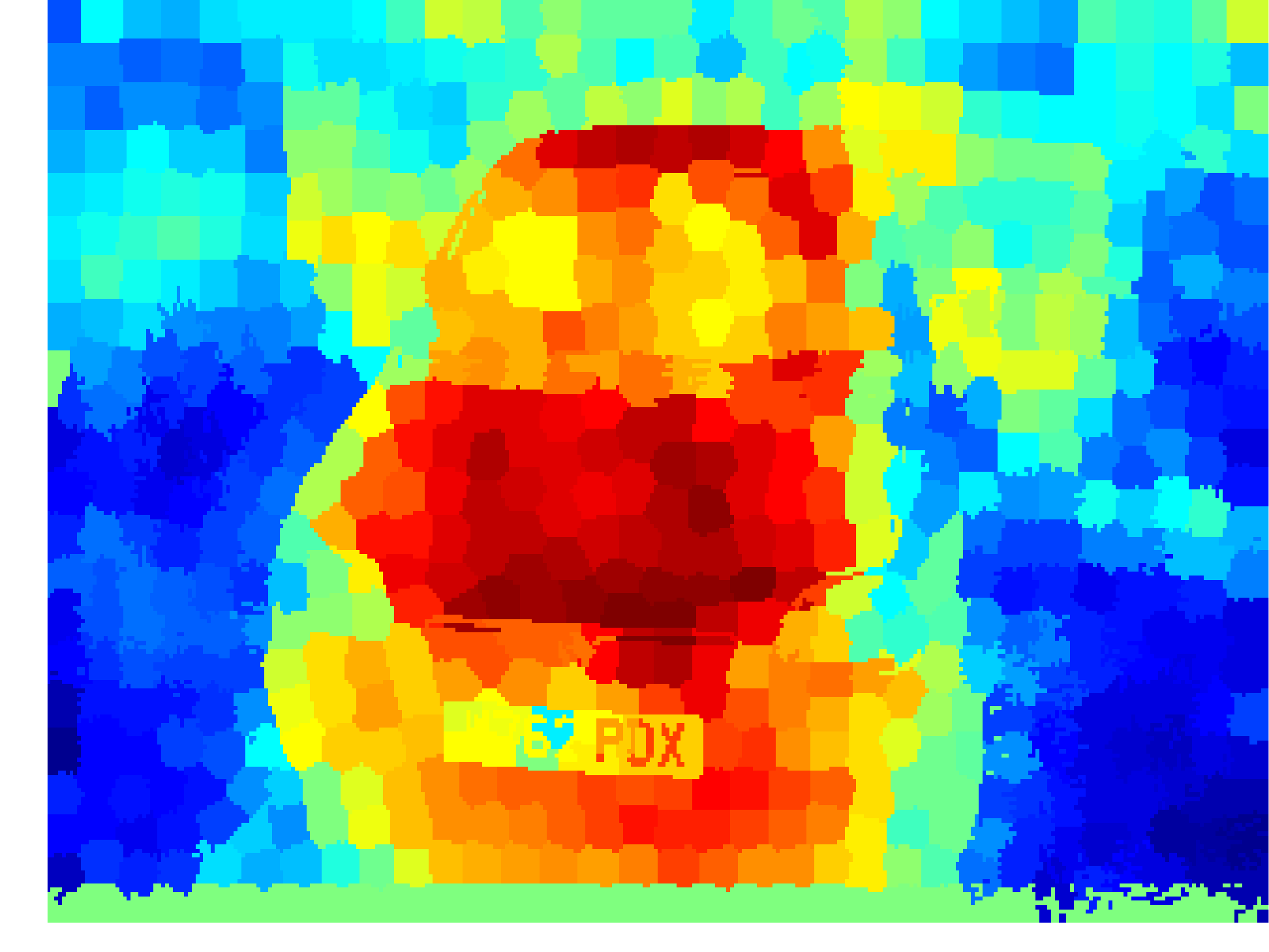}
  \includegraphics[width=0.115\linewidth]{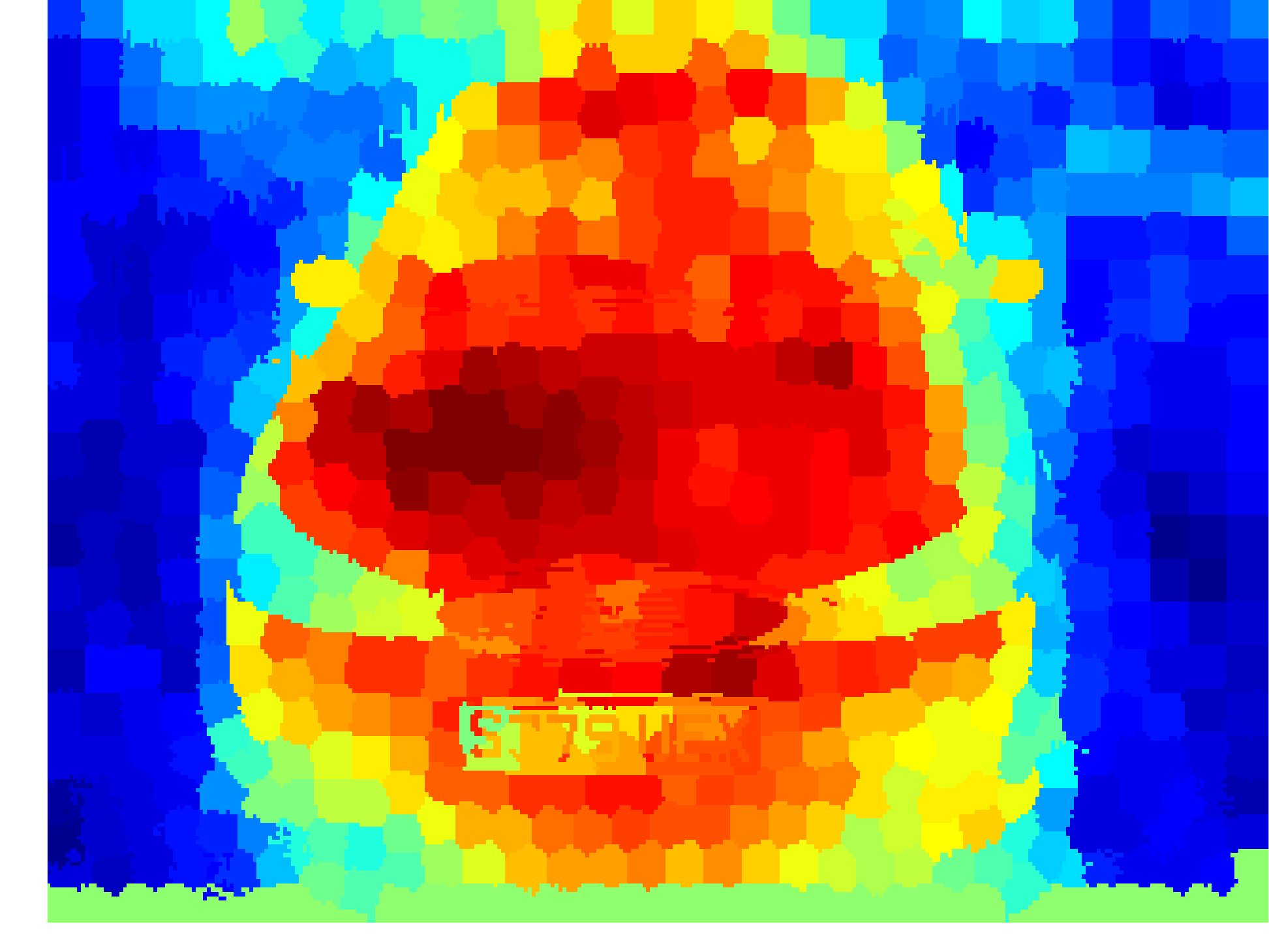}
  \includegraphics[width=0.115\linewidth]{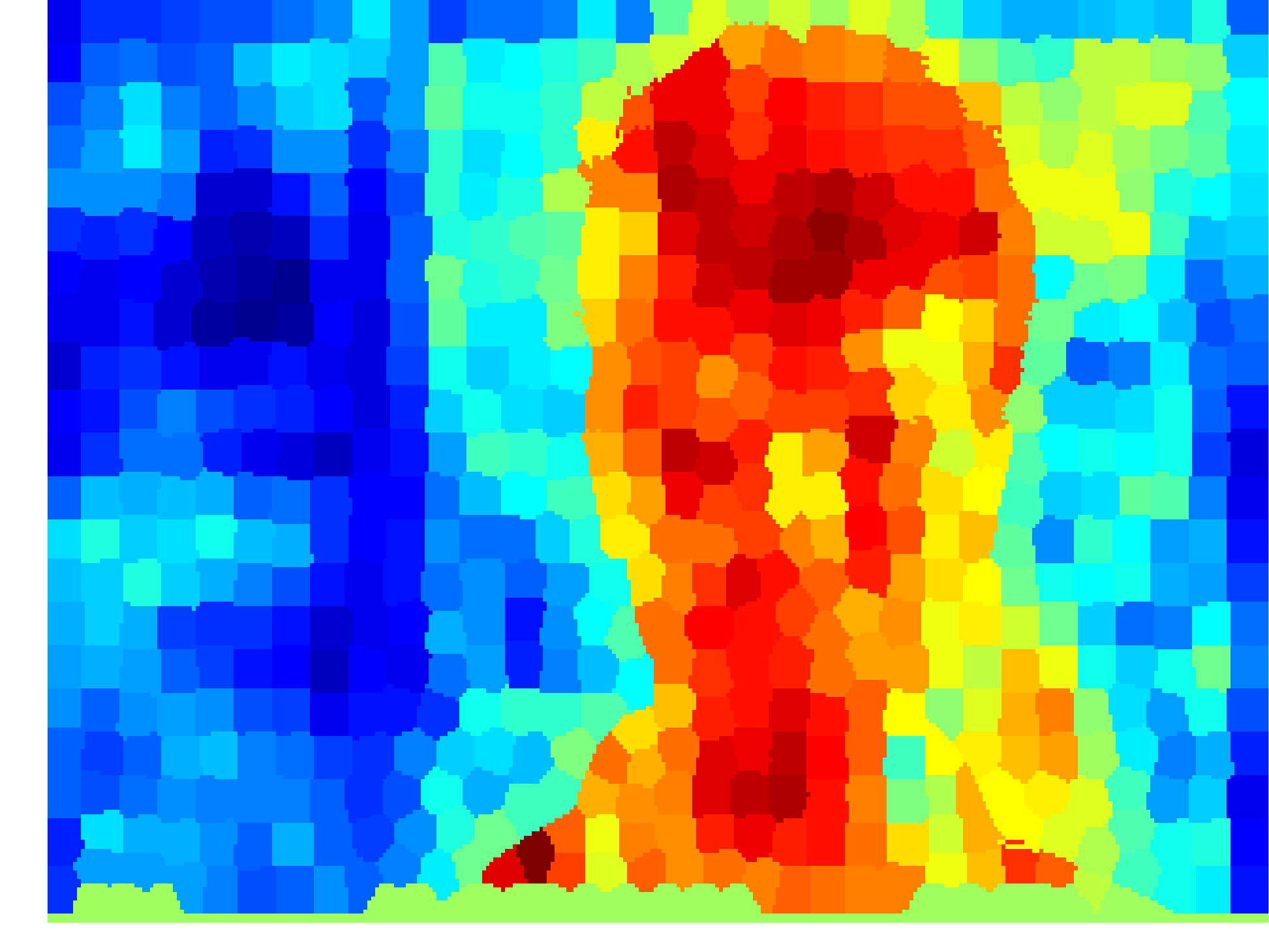}
  \includegraphics[width=0.115\linewidth]{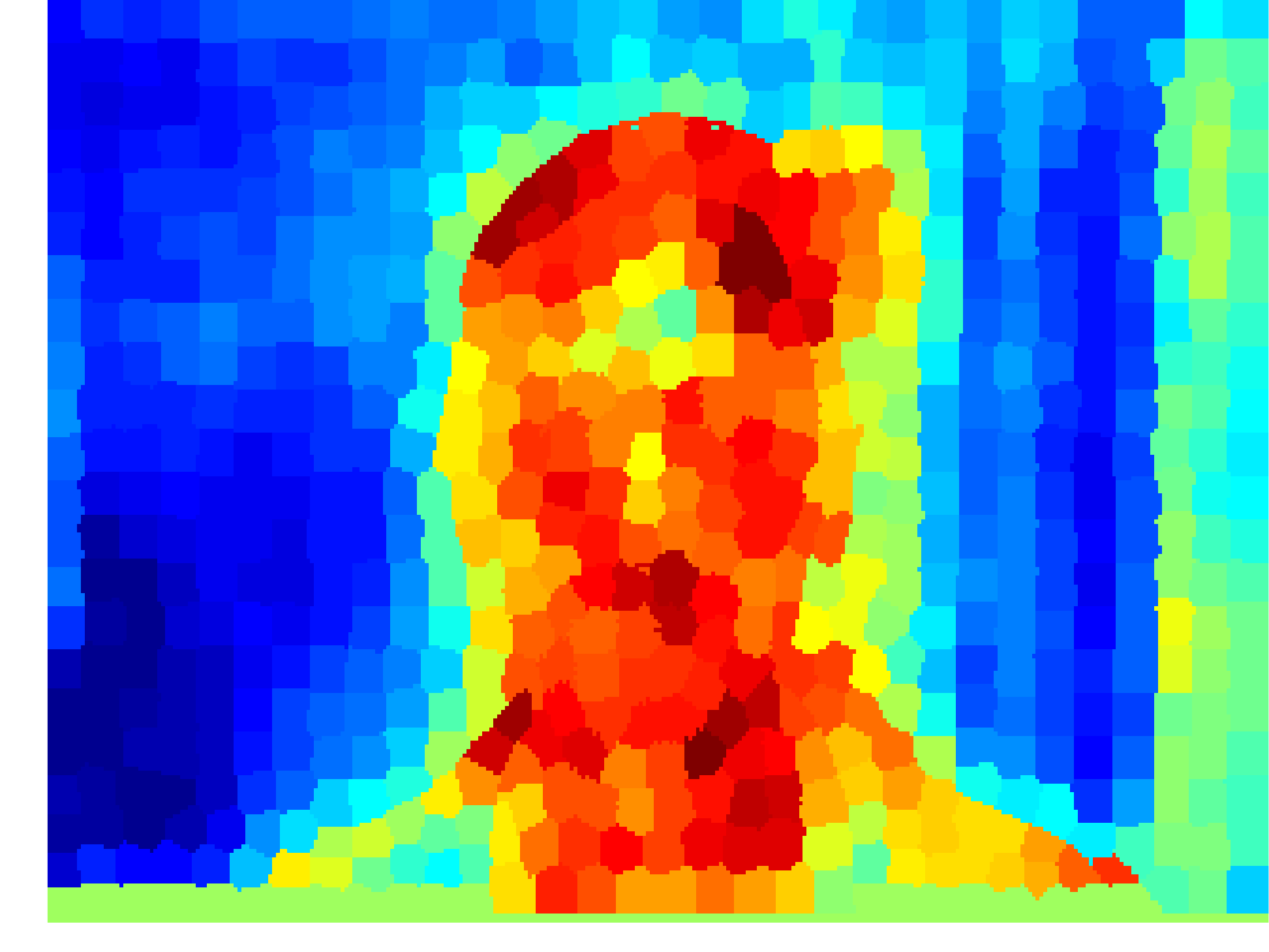}
  \includegraphics[width=0.115\linewidth]{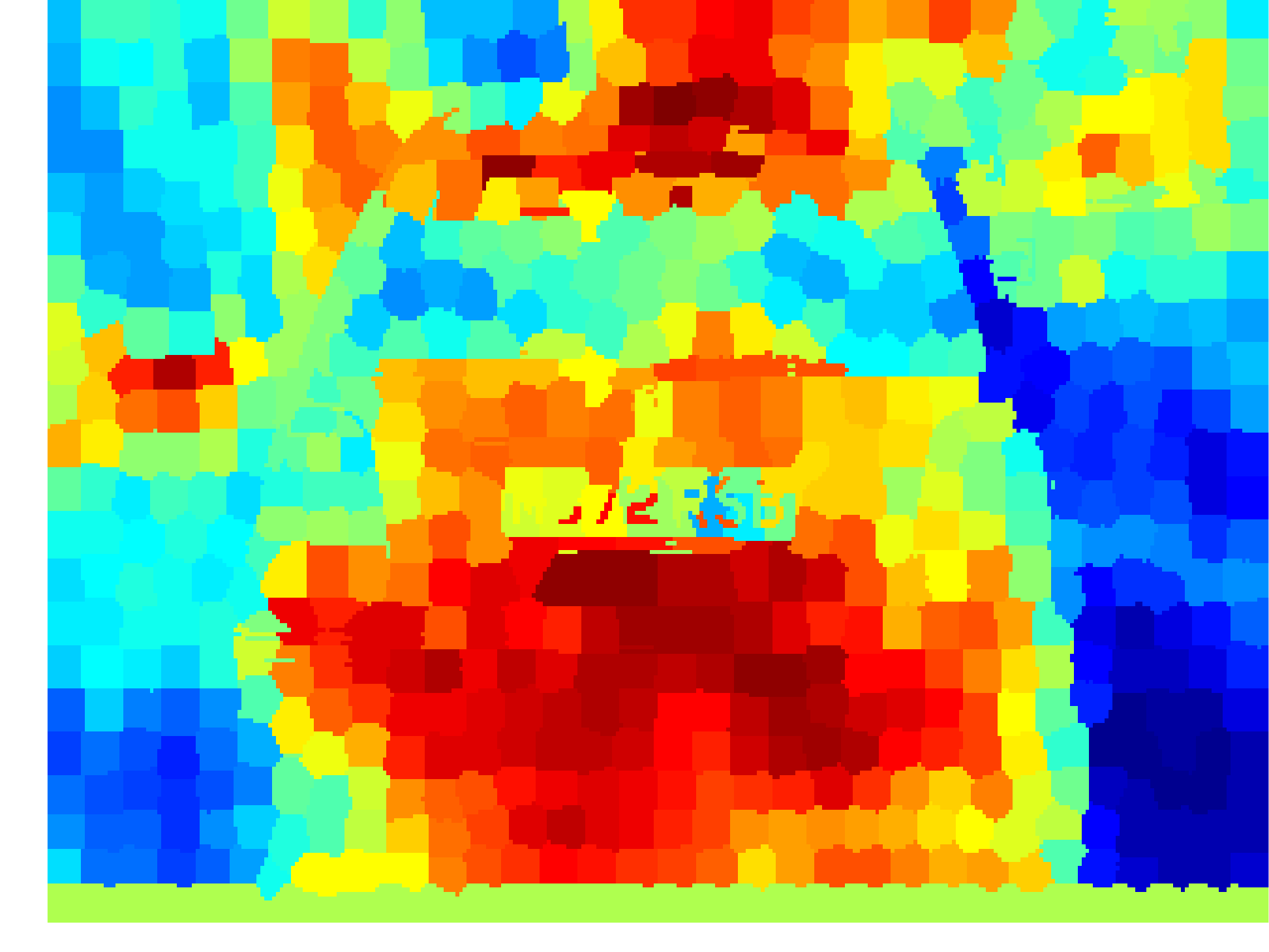}
  \includegraphics[width=0.115\linewidth]{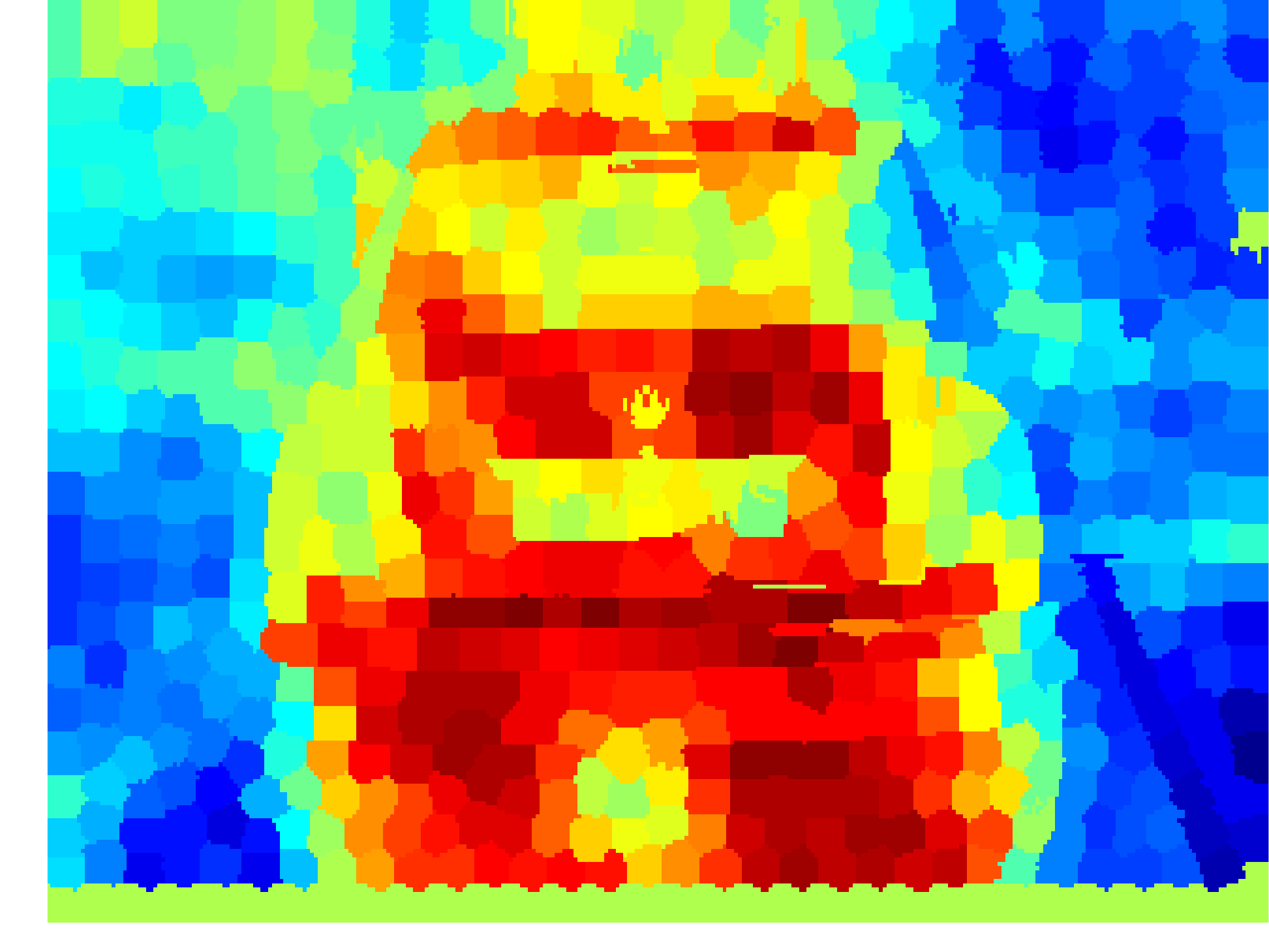}

  \caption{\small Co-segmentation results on the Weizman horses and MSRC
    datasets. From top to bottom: the original images, the results of LR, SDCut, and BCR, respectively.}
  \label{fig:coseg1}
\end{figure*}

Qualitative results are presented in Figure \ref{fig:coseg1} while Table \ref{table:coseg1} provides a quantitative comparison. From Table \ref{table:coseg1}, we observe that on average our method converges $\sim9.5\times$ faster than SDCut and $\sim60\times$ faster than LR. Moreover, the optimal objective value achieved by BCR is significantly lower than that achieved by both SDCut and LR methods. Figure~\ref{fig:coseg1} displays the visualization of the final score vector $\mathbf{x}_p^*$ for selected images, depicting that in general SDCut and BCR produce similar results.  Furthermore, the optimal BCR score vector $\mathbf{x}_p^*$ is extracted from a rank-$2$ solution, as compared to rank-$3$ and rank-$7$ solutions needed to get comparable results with SDCut and LR.

\subsection{Metric Learning on Manifolds}

Large SDPs play a central role in manifold methods for classification and dimensionality reduction on image sets and videos \cite{harandi2014manifold,wang2012covariance,huang2015log}.
 Manifold methods rely heavily on covariance matrices, which accurately characterize second-order statistics of variation between images. 
 Typical methods require computing distances between matrices along a Riemannian manifold---a task that is expensive for large matrices and limits the applicability of these techniques.  It is of interest to perform dimensionality reduction on SPD matrices, thus enabling the use of covariance methods on very large problems.

 In this section, we discuss dimensionality reduction on manifolds of SPD matrices using BCR, which is computationally
 much faster than the state-of-the-art while achieving comparable (and often better) performance.
\noindent
Consider a set of high-dimensional SPD matrices $\{\mathbf{S}_1, \dots, \mathbf{S}_n\}$ where $\mathbf{S}_i \in S^+_{N\times N}.$  We can project these onto a low-dimensional manifold of rank $K<N$ by solving
\begin{equation}
  \begin{aligned}
\underset{\mathbf{X}\in S^+_{N\times N}, \eta_{ij} \geq 0}{\text{minimize}} \quad &\trace(\mathbf{X}) + \textstyle \mu \sum_{i,j} \eta_{ij}\\ 
\text{subject to} \quad & \mathbb{D}_X(\mathbf{S}_i, \mathbf{S}_j) \leq u + \eta_{ij}, \quad \forall (i,j) \in \mathcal{C} \\
&  \mathbb{D}_X(\mathbf{S}_i, \mathbf{S}_j) \geq l - \eta_{ij}, \quad \forall (i,j) \in \mathcal{D}
\end{aligned} \label{ml:eq1}
\end{equation}
where $\mathbf{X}$ is a (low-dimensional) SPD matrix, $\mathbb{D}_X$ is Riemannian distance metric, and $\eta_{ij}$ are slack variables.  The sets $\mathcal{C}$ and $\mathcal{D}$ contain pairs of similar/dissimilar matrices labeled by the user, and the scalars $u$ and $l$ are given upper and lower bounds. For simplicity, we measure distance using the log-Euclidean metric (LEM)  defined by \cite{huang2015log}
\begin{align}
\mathbb{D}(\mathbf{S}_i,\mathbf{S}_j) &= \|\log(\mathbf{S}_i) - \log(\mathbf{S}_j)\|_F^2 
= \trace\!\left((\mathbf{R}_i - \mathbf{R}_j)^T(\mathbf{R}_i - \mathbf{R}_j)\right),
\end{align}
where $\mathbf{R}_i = \log(\mathbf{S}_i)$ is a matrix logarithm. When  $\mathbf{X}$ has rank $K,$ it is a transformation onto the space of rank $K$ covariance matrices, where the new distance is given by~\cite{huang2015log}
\begin{align}
\mathbb{D}_X(\mathbf{S}_i,\mathbf{S}_j) 
&= \trace\!\left(\mathbf{X}(\mathbf{R}_i - \mathbf{R}_j)^T(\mathbf{R}_i - \mathbf{R}_j)\right)\!. \label{ml:eq2}
\end{align}


We propose to solve the semi-definite program \eqref{ml:eq1} using the representation $\mathbf{X} = \mathbf{Y}\mathbf{Y}^T$ which puts our problem in the form \eqref{no_q} with  $\mathbf{A}_{ij} = (\mathbf{R}_i - \mathbf{R}_j)^T(\mathbf{R}_i - \mathbf{R}_j)$.   This problem is then solved using BCR, where the slack variables $\{\eta_{ij}\}$ are removed and instead a hinge loss penalty approximately enforces the inequality constraints in \eqref{biconvex}.
In our experiments we choose $u = \rho - \xi\tau$ and $l = \rho + \xi\tau$, where $\rho$ and $\tau$ are the mean and standard deviation of the pairwise distances between $\{S_i\}$ in the original space, respectively. The quantities $\xi$ and $\mu$ are treated as hyper-parameters.

{\bf Experiments:}
We analyze the performance of our approach (short BCRML) against state-of-the-art manifold metric learning algorithms
using three image set classification databases: ETH-80, YouTube Celebrities (YTC), and YouTube Faces (YTF) \cite{wolf2011face}. The ETH-80 database consists of a 10 image set for each of 8 object categories. YTC contains 1,910 video sequences for 47 subjects from YouTube. YTF is a face verification database containing 3,425 videos of 1,595 different people. Features were extracted from images as described in \cite{huang2015log}.
 Faces were cropped from each dataset using bounding boxes, and scaled to size $20\times 20$ for the ETH and YTC datasets.    For YTF we used a larger $30\times 30$ scaling, as larger images were needed to replicate the results reported in~\cite{huang2015log}.

We compare BCR to three state-of-the-art schemes: LEML \cite{huang2015log}  is based on a log-Euclidean metric, and minimizes the logdet divergence between matrices using Bregman projections. SPDML \cite{harandi2014manifold} optimizes a cost function on the Grassmannian manifold while making use of either the affine-invariant metric (AIM) or Stein metric. We use publicly available code for LEML and SPDML and follow the details in \cite{harandi2014manifold,huang2015log} to select algorithm specific hyper-parameters using cross-validation. For BCRML, we fix $\alpha$ to be $1 / \sqrt{|\mathcal{C} \cup \mathcal{D}|}$ and $\mu$ as $\alpha/2$. The $\xi$ is fixed to $0.5,$ which performed well under cross-validation. For SPDML, the dimensionality of the target manifold $K$ is fixed to 100. In LEML, the dimension cannot be reduced and thus the final dimension is the same as the original. Hence, for a fair comparison, we report the performance of BCRML using full target dimension (BCRML-full) as well as for $K = 100$ (BCRML-100). 

Table \ref{tab:ml} summarizes the classification performance on the above datasets. We observe that BCRML performs almost the same or better than other ML algorithms. One can apply other algorithms to gain a further performance boost after projecting onto the low-dimensional manifold. Hence, we also provide a performance evaluation for LEML and BCRML using the LEM based CDL-LDA recognition algorithm \cite{wang2012covariance}. The last three columns of Table \ref{tab:ml} display the runtime measured on the YTC dataset. We note that BCRML-100 trains roughly $2\times$ faster and overall runs  about $3.5 \times$ faster than the next fastest method. Moreover, on testing using CDL-LDA, the overall computation time is  approximately $5 \times$ faster in comparison to the next-best performing approach.
 \begin{table}[tb]
   \centering
   \caption{\small Image set classification results for state-of-the-art metric learning algorithms. The last three columns report computation time in seconds. The last 3 rows report performance using CDL-LDA after dimensionality reduction.  Methods using the proposed BCR are listed in bold. }   
   {\small
   \begin{tabular}{|c||c|c|c||c|c|c|} 
     \hline
     Method & ETH-80 & YTC & YTF & Train (s) & Test (s) & Total (s) \\ [0.5ex] 
     \hline\hline
     AIM & 89.25 $\pm$ 1.69 & 62.77 $\pm$ 2.89 & 59.82 $\pm$ 1.63 & - & 5.189 & 1463.3 \\ 
     \hline
     Stein & 89.00 $\pm$ 2.42 & 62.02 $\pm$ 2.71 & 57.56 $\pm$ 2.17 & - & 3.593 & 1013.3 \\ 
     \hline     
     LEM & 90.00 $\pm$ 2.64 & 62.06 $\pm$  3.04 & 59.78 $\pm$ 1.69 & - & 1.641 & 462 \\
     \hline \hline
     SPDML-AIM \cite{harandi2014manifold}  & 91.00 $\pm$ 3.39 & 65.32 $\pm$ 2.77  & 61.64 $\pm$ 1.46 & 3941 & 0.227 & 4005\\
     \hline	     
     SPDML-Stein \cite{harandi2014manifold} & 90.75 $\pm$ 3.34 & 66.10 $\pm$ 2.92	 & 61.66 $\pm$ 2.09 & 1447 & \textbf{0.024} & 1453.7\\
     \hline
     LEML \cite{huang2015log} & 92.00 $\pm$ 2.18 & 62.13 $\pm$ 3.13 &	60.92 $\pm$ 1.95  & 93 & 1.222 & 437.7\\
     \hline
     \textbf{BCRML-full} & 92.00$\pm$ 3.12 &	64.40 $\pm$ 2.92 & 	60.58 $\pm$ 1.75 & 189 & 1.222 & 669.7 \\
     \hline          
     \textbf{BCRML-100} & 92.25 $\pm$ 3.78 &	64.61 $\pm$ 2.65 & \textbf{62.42} $\pm$\textbf{ 2.14} & \textbf{45} & 0.291 & 127 \\
     \hline \hline
     CDL-LDA \cite{wang2012covariance} & \textbf{94.25} $\pm$ \textbf{3.36} & 72.94 $\pm$ 1.81 &	N/A & - & 1.073 & 302.7\\
     \hline 
     LEML+CDL-LDA \cite{huang2015log} & 94.00 $\pm$ 3.57 & 73.01 $\pm$ 1.67 &	N/A & 93 & 0.979 & 369\\
     \hline              
     \textbf{BCRML-100+CDL-LDA} & 93.75 $\pm$ 3.58  & \textbf{73.48} $\pm$ \textbf{1.83} & 	N/A & 45 & 0.045 & \textbf{57.7} \\
     \hline
   \end{tabular}
   }
\label{tab:ml}
 \end{table}
\section{Conclusion}
We have presented a novel biconvex relaxation framework (BCR) that enables the solution of general semidefinite programs (SDPs) at low complexity and with a small memory footprint. 
We have provided an alternating minimization (AM) procedure along with a new initialization method that, together, are guaranteed to converge, computationally efficient (even for large-scale problems), and able to handle a variety of SDPs.
Comparisons of BCR with state-of-the-art methods for specific computer 
vision problems, such as segmentation, co-segmentation, and metric learning, show that BCR provides similar or better solution quality with significantly lower runtime.
While this paper only shows applications for a select set of computer vision problems, determining the efficacy of BCR for other problems in signal processing, machine learning, control, etc. is left for future work.  


\noindent
{\bf Acknowledgements}: The work of S.~Shah and T.~Goldstein was supported in part by the US National Science Foundation (NSF) under grant CCF-1535902 and by the US Office of Naval Research under grant N00014-15-1-2676. The work of A.~Yadav and D.~Jacobs was supported by the US NSF under grants IIS-1526234 and IIS-1302338. The work of C.~Studer was supported in part by Xilinx Inc., and by the US NSF under grants ECCS-1408006 and CCF-1535897.

\bibliographystyle{splncs}
\bibliography{1259}

\begin{thebibliography}{10}

\bibitem{shi2000normalized}
Shi, J., Malik, J.:
\newblock Normalized cuts and image segmentation.
\newblock Pattern Analysis and Machine Intelligence, IEEE Transactions on
  \textbf{22}(8) (2000)  888--905

\bibitem{keuchel2003binary}
Keuchel, J., Schno, C., Schellewald, C., Cremers, D.:
\newblock Binary partitioning, perceptual grouping, and restoration with
  semidefinite programming.
\newblock IEEE Transactions on Pattern Analysis and Machine Intelligence
  \textbf{25}(11) (2003)  1364--1379

\bibitem{torr2003solving}
Torr, P.H.:
\newblock Solving markov random fields using semi definite programming.
\newblock In: Artificial Intelligence and Statistics. (2003)

\bibitem{goemans1995improved}
Goemans, M.X., Williamson, D.P.:
\newblock Improved approximation algorithms for maximum cut and satisfiability
  problems using semidefinite programming.
\newblock Journal of the ACM (JACM) \textbf{42}(6) (1995)  1115--1145

\bibitem{arie2012global}
Arie-Nachimson, M., Kovalsky, S.Z., Kemelmacher-Shlizerman, I., Singer, A.,
  Basri, R.:
\newblock Global motion estimation from point matches.
\newblock In: 3D Imaging, Modeling, Processing, Visualization and Transmission
  (3DIMPVT), 2012 Second International Conference on, IEEE (2012)  81--88

\bibitem{weinberger2006unsupervised}
Weinberger, K.Q., Saul, L.K.:
\newblock Unsupervised learning of image manifolds by semidefinite programming.
\newblock International Journal of Computer Vision \textbf{70}(1) (2006)
  77--90

\bibitem{mitra2010large}
Mitra, K., Sheorey, S., Chellappa, R.:
\newblock Large-scale matrix factorization with missing data under additional
  constraints.
\newblock In: Advances in Neural Information Processing Systems (NIPS). (2010)
  1651--1659

\bibitem{luo2010semidefinite}
Luo, Z.Q., Ma, W.k., So, A.M.C., Ye, Y., Zhang, S.:
\newblock Semidefinite relaxation of quadratic optimization problems.
\newblock IEEE Signal Processing Magazine \textbf{27}(3) (May 2010)  20--34

\bibitem{lasserre2001explicit}
Lasserre, J.B.:
\newblock An explicit exact sdp relaxation for nonlinear 0-1 programs.
\newblock In: Integer Programming and Combinatorial Optimization.
\newblock Springer (2001)  293--303

\bibitem{boyd1997semidefinite}
Boyd, S., Vandenberghe, L.:
\newblock Semidefinite programming relaxations of non-convex problems in
  control and combinatorial optimization.
\newblock In: Communications, Computation, Control, and Signal Processing.
\newblock Springer (1997)  279--287

\bibitem{ecker2008semidefinite}
Ecker, A., Jepson, A.D., Kutulakos, K.N.:
\newblock Semidefinite programming heuristics for surface reconstruction
  ambiguities.
\newblock In: Computer Vision--ECCV 2008.
\newblock Springer (2008)  127--140

\bibitem{shirdhonkar2005non}
Shirdhonkar, S., Jacobs, D.W.:
\newblock Non-negative lighting and specular object recognition.
\newblock In: Computer Vision, 2005. ICCV 2005. Tenth IEEE International
  Conference on. Volume~2., IEEE (2005)  1323--1330

\bibitem{vandenberghe1996semidefinite}
Vandenberghe, L., Boyd, S.:
\newblock Semidefinite programming.
\newblock SIAM review \textbf{38}(1) (July 1996)  49--95

\bibitem{heiler2005semidefinite}
Heiler, M., Keuchel, J., Schn{\"o}rr, C.:
\newblock Semidefinite clustering for image segmentation with a-priori
  knowledge.
\newblock In: Pattern Recognition.
\newblock Springer (2005)  309--317

\bibitem{shen2011scalable}
Shen, C., Kim, J., Wang, L.:
\newblock A scalable dual approach to semidefinite metric learning.
\newblock In: Computer Vision and Pattern Recognition (CVPR), 2011 IEEE
  Conference on, IEEE (2011)  2601--2608

\bibitem{netrapalli2013phase}
Netrapalli, P., Jain, P., Sanghavi, S.:
\newblock Phase retrieval using alternating minimization.
\newblock In: Advances in Neural Information Processing Systems (NIPS). (2013)
  2796--2804

\bibitem{candes2015phase}
Cand\`es, E.J., Li, X., Soltanolkotabi, M.:
\newblock Phase retrieval via wirtinger flow: Theory and algorithms.
\newblock Information Theory, IEEE Transactions on \textbf{61}(4) (2015)
  1985--2007

\bibitem{lee2010practical}
Lee, J.D., Recht, B., Srebro, N., Tropp, J., Salakhutdinov, R.R.:
\newblock Practical large-scale optimization for max-norm regularization.
\newblock In: Advances in Neural Information Processing Systems. (2010)
  1297--1305

\bibitem{Wang_2013}
Wang, P., Shen, C., van~den Hengel, A.:
\newblock A fast semidefinite approach to solving binary quadratic problems.
\newblock In: The IEEE Conference on Computer Vision and Pattern Recognition
  (CVPR). (June 2013)

\bibitem{biasedncut}
Maji, S., Vishnoi, N.K., Malik, J.:
\newblock Biased normalized cuts.
\newblock In: IEEE Conference on Computer Vision and Pattern Recognition
  (CVPR), IEEE (2011)  2057--2064

\bibitem{Journee}
Journ{\'e}e, M., Bach, F., Absil, P.A., Sepulchre, R.:
\newblock Low-rank optimization on the cone of positive semidefinite matrices.
\newblock SIAM Journal on Optimization \textbf{20}(5) (2010)  2327--2351

\bibitem{huang2015log}
Huang, Z., Wang, R., Shan, S., Li, X., Chen, X.:
\newblock Log-euclidean metric learning on symmetric positive definite manifold
  with application to image set classification.
\newblock In: Proceedings of the 32nd International Conference on Machine
  Learning (ICML-15). (2015)  720--729

\bibitem{harandi2014manifold}
Harandi, M.T., Salzmann, M., Hartley, R.:
\newblock From manifold to manifold: Geometry-aware dimensionality reduction
  for spd matrices.
\newblock In: European Conference on Computer Vision, Springer (2014)  17--32

\bibitem{bai2004graph}
Bai, X., Yu, H., Hancock, E.R.:
\newblock Graph matching using spectral embedding and alignment.
\newblock In: International Conference on Pattern Recognition (ICPR).
  Volume~3., IEEE (2004)  398--401

\bibitem{wang2013markov}
Wang, C., Komodakis, N., Paragios, N.:
\newblock Markov random field modeling, inference \& learning in computer
  vision \& image understanding: A survey.
\newblock Computer Vision and Image Understanding \textbf{117}(11) (2013)
  1610--1627

\bibitem{Joulin_2010}
Joulin, A., Bach, F., Ponce, J.:
\newblock Discriminative clustering for image co-segmentation.
\newblock In: Proceedings of the Conference on Computer Vision and Pattern
  Recognition (CVPR). (2010)

\bibitem{wang2015efficient}
Wang, P., Shen, C., van~den Hengel, A.:
\newblock Efficient {SDP} inference for fully-connected {CRF}s based on
  low-rank decomposition.
\newblock In: The IEEE Conference on Computer Vision and Pattern Recognition
  (CVPR). (June 2015)

\bibitem{schellewald2005probabilistic}
Schellewald, C., Schn{\"o}rr, C.:
\newblock Probabilistic subgraph matching based on convex relaxation.
\newblock In: Energy minimization methods in computer vision and pattern
  recognition, Springer (2005)  171--186

\bibitem{olsson2007solving}
Olsson, C., Eriksson, A.P., Kahl, F.:
\newblock Solving large scale binary quadratic problems: Spectral methods vs.
  semidefinite programming.
\newblock In: Computer Vision and Pattern Recognition, 2007. CVPR'07. IEEE
  Conference on, IEEE (2007)  1--8

\bibitem{lang2005fixing}
Lang, K.:
\newblock Fixing two weaknesses of the spectral method.
\newblock In: Advances in Neural Information Processing Systems (NIPS). (2005)
  715--722

\bibitem{Sturm98usingsedumi}
Sturm, J.F.:
\newblock Using sedumi 1.02, a matlab toolbox for optimization over symmetric
  cones.
\newblock Optimization methods and software \textbf{11}(1-4) (1999)  625--653

\bibitem{Toh98sdpt3}
Toh, K.C., Todd, M., Tutuncu, R.:
\newblock Sdpt3 - a matlab software package for semidefinite programming.
\newblock Optimization Methods and Software \textbf{11} (1998)  545--581

\bibitem{okatani2007wiberg}
Okatani, T., Deguchi, K.:
\newblock On the wiberg algorithm for matrix factorization in the presence of
  missing components.
\newblock International Journal of Computer Vision \textbf{72}(3) (2007)
  329--337

\bibitem{burer2003nonlinear}
Burer, S., Monteiro, R.D.:
\newblock A nonlinear programming algorithm for solving semidefinite programs
  via low-rank factorization.
\newblock Mathematical Programming \textbf{95}(2) (2003)  329--357

\bibitem{duchisinger}
Duchi, J.C., Singer, Y.:
\newblock Efficient online and batch learning using forward backward splitting.
\newblock Journal of Machine Learning Research \textbf{10} (2009)  2899--2934

\bibitem{douglasgunn}
Douglas, J., Gunn, J.E.:
\newblock A general formulation of alternating direction methods.
\newblock Numerische Mathematik \textbf{6}(1) (1964)  428--453

\bibitem{xu2013block}
Xu, Y., Yin, W.:
\newblock A block coordinate descent method for regularized multiconvex
  optimization with applications to nonnegative tensor factorization and
  completion.
\newblock SIAM Journal on imaging sciences \textbf{6}(3) (2013)  1758--1789

\bibitem{NIPS2015_5830}
Zheng, Q., Lafferty, J.:
\newblock A convergent gradient descent algorithm for rank minimization and
  semidefinite programming from random linear measurements.
\newblock In: Neural Information Processing Systems (NIPS).
\newblock (2015)

\bibitem{GoldsteinStuderBaraniuk:2014}
Goldstein, T., Studer, C., Baraniuk, R.:
\newblock A field guide to forward-backward splitting with a {FASTA}
  implementation.
\newblock arXiv eprint \textbf{abs/1411.3406} (2014)

\bibitem{ratiocut}
Wang, S., Siskind, J.M.:
\newblock Image segmentation with ratio cut.
\newblock IEEE Transactions on Pattern Analysis and Machine Intelligence
  \textbf{25} (2003)  675--690

\bibitem{yu2004segmentation}
Yu, S.X., Shi, J.:
\newblock Segmentation given partial grouping constraints.
\newblock Pattern Analysis and Machine Intelligence, IEEE Transactions on
  \textbf{26}(2) (2004)  173--183

\bibitem{MartinFTM01}
Martin, D., Fowlkes, C., Tal, D., Malik, J.:
\newblock A database of human segmented natural images and its application to
  evaluating segmentation algorithms and measuring ecological statistics.
\newblock In: Proc. 8th Int'l Conf. Computer Vision. Volume~2. (July 2001)
  416--423

\bibitem{vedaldi2012vlfeat}
Vedaldi, A., Fulkerson, B.:
\newblock Vlfeat: An open and portable library of computer vision algorithms
  (2008) (2012)

\bibitem{wang2012covariance}
Wang, R., Guo, H., Davis, L.S., Dai, Q.:
\newblock Covariance discriminative learning: A natural and efficient approach
  to image set classification.
\newblock In: Computer Vision and Pattern Recognition (CVPR), 2012 IEEE
  Conference on, IEEE (2012)  2496--2503

\bibitem{wolf2011face}
Wolf, L., Hassner, T., Maoz, I.:
\newblock Face recognition in unconstrained videos with matched background
  similarity.
\newblock In: Computer Vision and Pattern Recognition (CVPR), 2011 IEEE
  Conference on, IEEE (2011)  529--534

\end{thebibliography}
\end{document}